\documentclass{article}

\usepackage{microtype}
\usepackage{graphicx}
\usepackage{subcaption}
\usepackage{booktabs} % for professional tables
\usepackage{longtable}

\usepackage{hyperref}

\usepackage[preprint]{icml2026}

\usepackage{amsmath}
\usepackage{amssymb}
\usepackage{mathtools}
\usepackage{amsthm}
\usepackage{bm,bbm}
\usepackage{physics}
\usepackage{xcolor}
\usepackage{enumitem}

\usepackage[capitalize,noabbrev]{cleveref}

\theoremstyle{plain}
\newtheorem{theorem}{Theorem}[section]

\newtheorem{lemma}[theorem]{Lemma}

\theoremstyle{definition}

\theoremstyle{remark}

\newcommand\numberthis{\addtocounter{equation}{1}\tag{\theequation}}
\newcommand{\argmin}{\mathop{\arg\min}}
\newcommand{\argmax}{\mathop{\arg\max}}
\newcommand{\m}{\middle|}
\newcommand{\R}{\mathcal{R}}
\newcommand{\nec}{\mathcal{A}}

\newcommand{\sC}{\mathcal{C}}
\newcommand{\sI}{\mathcal{I}}

\newcommand{\sM}{\mathcal{M}}
\newcommand{\sP}{\mathbb{P}}
\newcommand{\E}{\mathbb{E}}
\newcommand{\dis}{\mathcal{D}_\alpha}
\newcommand{\Fdis}{\mathcal{F}_\alpha}
\newcommand{\Pdis}{\mathcal{P}_\alpha}

\newcommand{\ind}{\mathbbm{1}}
\newcommand{\pn}[1]{^{(#1)}}
\newcommand{\ul}{\underline{\lambda}}

\newcommand{\ma}{m_\alpha}
\newcommand{\da}{d_\alpha}

\newcommand{\nn}{\nonumber\\}
\newcommand{\n}{\nonumber}

\newcommand{\lams}{\lambda^*}

\newcommand{\wlow}{w^*(\dis)}
\newcommand{\supp}[1]{\operatorname{supp}\left(#1\right)}

\newcommand{\necc}[1]{\nec(#1)}
\newcommand{\necone}[1]{\nec_{#1}}
\newcommand{\necr}[1]{\nec|#1}
\newcommand{\neccr}[2]{\nec(#1)|#2}
\newcommand{\base}{\mathcal{A}}
\newcommand{\basec}[1]{\sI(#1)}

\newcommand{\basecr}[2]{\sI(#1)|#2}

\newcommand{\Bex}[1]{\armset_{#1}^{\textrm{ex}}}
\newcommand{\Bpair}{(\armset^{\cap}_{\prec i},\armset^{\cap}_{\succ i})}
\newcommand{\gap}[1]{\Delta_{#1,\min}}
\newcommand{\bw}{{\bf w}}
\newcommand{\easyp}[2]{\mathbf{Q}_{#1}\qty(#2)}
\newcommand{\easypstar}[2]{\mathbb{Q}^*_{#1}\qty(#2)}
\newcommand{\sele}{A}
\newcommand{\armset}{E}
\newcommand{\blk}[2]{\kappa_{#1}(#2)}
\newcommand{\blkomit}[1]{\kappa_{#1}}
\newcommand{\indset}{{\bm x}}

\usepackage[textsize=tiny]{todonotes}

\icmltitlerunning{A Polynomial-Time BOBW Algorithm for Matroid Semi-Bandits}

\begin{document}

\twocolumn[
  \icmltitle{A Polynomial-Time Best-of-Both-Worlds Algorithm\\
  for Matroid Semi-Bandit Problems}



  \icmlsetsymbol{equal}{*}

  \begin{icmlauthorlist}
    \icmlauthor{Botao Chen}{KyotoU}
    \icmlauthor{Junya Honda}{KyotoU,RIKEN}
    % \icmlauthor{Firstname3 Lastname3}{comp}
  \end{icmlauthorlist}

  \icmlaffiliation{KyotoU}{Kyoto University, Kyoto, Japan}
  \icmlaffiliation{RIKEN}{RIKEN AIP, Tokyo, Japan}

  \icmlcorrespondingauthor{Botao Chen}{chen.botao.63r@st.kyoto-u.ac.jp}

  \icmlkeywords{Machine Learning, ICML}

  \vskip 0.3in
]



\printAffiliationsAndNotice{}  % no special notice (required even if empty)

\begin{abstract}
Follow-the-Perturbed-Leader (FTPL) offers a computationally attractive framework for online learning, with recent advances establishing Best-of-Both-Worlds (BOBW) guarantees for various bandit settings.
For matroid semi-bandits, existing methods offering such guarantees primarily rely on Follow-the-Regularized-Leader (FTRL), which requires solving an optimization problem at each round.
We establish BOBW guarantees for FTPL with Fr\'{e}chet and Pareto distributions with a certain parameter with geometric resampling in the matroid semi-bandit problem, showing that FTPL can simultaneously achieve the minimax-optimal $O(\sqrt{mdT})$ regret in adversarial setting and a logarithmic regret in stochastic setting.
To the best of our knowledge, this is the first polynomial-time algorithm
for matroid semi-bandits that achieves BOBW guarantees.
\end{abstract}

\section{Introduction}

Many sequential decision-making problems involve combinatorial structures and limited feedback, and the combinatorial semi-bandit problem \citep{cesa2012combinatorial} provides a fundamental framework for studying such problems.
This framework is an extension of the multi-armed bandit (MAB) problem, and is applied to many real-world scenarios, such as online advertising \citep{nuara2022online}, recommendation systems \citep{wang2017efficient}, and adaptive routing \citep{gai2012combinatorial}. 
In this problem, at each round $t\in[T]=\{1,2,\ldots,T\}$, the learner selects an action $a_t$ from an action set $\nec\subset\{0,1\}^d$.
The loss vector $\ell_t\in[0,1]^d$ is determined by the environment, and the learner can only observe $\{\ell_{t,i}:a_{t,i}=1\}$, i.e., the individual losses of the selected base-arms.
The matroid semi-bandit problem \citep{kveton2014matroid} is an important special case of the combinatorial semi-bandit problem, in which the feasible actions correspond to the bases of a matroid.
This structure captures a variety of applications.
Uniform matroids capture cardinality constraints in online advertising
% and news selection 
\citep{kale2010non}.
Partition matroids model slot-based constraints in ad placement
\citep{streeter2009online} and category-based diversity constraints
in recommendation \citep{abbassi2013diversity,katariya2019conservative}.
Graphic matroids capture spanning-tree constraints in network routing
\citep{kveton2014matroid}.
Transversal matroids represent matching-based feasibility constraints in task assignment \citep{chen2016pure}.

The combinatorial semi-bandit problem has been extensively investigated in the stochastic and adversarial settings, which represent two fundamental regimes for how the environment generates the loss sequence.
In the stochastic setting, the losses of each base-arm $i$ are independent and identically distributed (i.i.d.) over time, following a fixed but unknown distribution with mean $\mu_i$.
Combinatorial Upper Confidence Bound (CUCB) algorithms have been
extensively studied for stochastic combinatorial semi-bandits
\citep{chen2013combinatorial,kveton2015tight}.
Furthermore, \citet{kveton2014matroid} exploited
the matroid structure to establish sharper regret bounds.
\citet{wang2018thompson} later proposed Combinatorial Thompson Sampling (CTS) for combinatorial semi-bandits, and derived a regret bound for matroid semi-bandits.
All of these algorithms select actions using the greedy method (see Algorithm~\ref{alg:greedy}) in matroid semi-bandits.

In the adversarial setting, the loss vectors $\ell_t\in[0,1]^d$ may be chosen arbitrarily by an adversary, without any assumption on the distribution of the loss vectors \citep{audibert2014regret,neu2015first}.
In this setting, several algorithms achieve the minimax regret bound $O(\sqrt{mdT})$, including OSMD \citep{audibert2014regret} and FTRL with a hybrid regularizer \citep{zimmert2019beating}. This matches the lower bound $\Omega(\sqrt{mdT})$ established by \citet{audibert2014regret}.

Although minimax regret guarantees are essential for robustness against adversarial losses, they may fail to exploit the favorable structure of stochastic environments, where logarithmic regret can be achieved. Best-of-Both-Worlds (BOBW, \citealp{bubeck2012best}) algorithms aim to bridge these two settings by simultaneously achieving optimal regret guarantees in both settings, without prior knowledge of the environment.
BOBW guarantees have also been extensively studied for combinatorial semi-bandits \citep{wei2018more,zimmert2019beating,ito2021hybrid,tsuchiya2023further}.

However, most existing BOBW algorithms rely on the Follow-the-Regularized-Leader (FTRL) framework, which requires solving an optimization problem at each round.
This computational issue becomes more pronounced for matroid semi-bandits.
In particular, no polynomial-time implementation of existing FTRL-based algorithms is currently known for matroid semi-bandits and more general settings, which limits their practical applicability despite their strong regret guarantees.
In contrast, Follow-the-Perturbed-Leader (FTPL) greedily selects an action by minimizing a randomly perturbed cumulative loss, which gains wide attention due to its optimization-free nature and computational efficiency.
Recent work showed that FTPL achieves BOBW guarantees for MAB~\citep{pmlr-v201-honda23a,pmlr-v247-lee24a,lee2025revisiting}, with its computational complexity subsequently reduced to nearly linear time~\citep{chen2025cgr}.
These guarantees were further extended to $m$-set semi-bandits~\citep{chen2026further}, which is a special case of matroid semi-bandits.
In addition, the action selection of FTPL can be efficiently implemented by the greedy method under matroid constraints.
These results naturally raise the question of whether FTPL can achieve BOBW guarantees for matroid semi-bandits, which cover a broader class of combinatorial constraints and present substantially greater analytical challenges.

\paragraph{Contribution of the Paper}
We establish FTPL as a computationally efficient BOBW algorithm
for general matroid semi-bandits.
To be specific, FTPL with either Fr\'{e}chet or Pareto perturbations
of shape parameter $\alpha=2$ with geometric resampling
simultaneously achieves minimax-optimal adversarial regret
and gap-dependent logarithmic stochastic regret.
To the best of our knowledge, this is 
not only the first algorithm
with greedy action selection to provide BOBW guarantees,
but, more fundamentally, the first such algorithm to
achieve minimax optimality in this setting.

The main technical challenge is that existing analyses of FTPL for MAB~\citep{pmlr-v201-honda23a,pmlr-v247-lee24a} and
$m$-set semi-bandits~\citep{chen2026further}, which exploit
the Top-$m$ ($m=1$ for MAB) selection rule, do not carry over directly
to general matroid constraints.
For matroid semi-bandits, whether a base-arm is selected depends on the identities, rather than only the number, of the preceding base-arms, which limits the applicability of the symmetry and counting arguments in the previous works.
We overcome this challenge by providing more flexible and general bounds, which leverage the structural properties of matroids without relying on counting arguments or symmetry among base-arms. 
These results support the generalization of regret analysis of FTPL for general matroid semi-bandits, and may also provide a methodological foundation for future work.

\subsection{Related Work}

\paragraph{Follow-the-Perturbed-Leader.}
FTPL dates back to \citet{hannan1957approximation} for repeated games and was later developed into an efficient framework for adversarial online
linear optimization by \citet{kalai2005efficient}.
For adversarial MAB, \citet{abernethy2015fighting} used convex smoothing to establish near-optimal FTPL regret with bounded-hazard-rate perturbations.
FTPL has also been studied for linear bandits \citep{mcmahan2004online,wei2026follow}, combinatorial semi-bandits \citep{neu2015first,JMLR:v17:15-091}, and MDP bandits \citep{dai2022follow}.
Recent work established BOBW guarantees for FTPL in MAB with Fr\'{e}chet-type perturbations \citep{pmlr-v201-honda23a,pmlr-v247-lee24a,lee2025revisiting}.
Further developments include improvements in computational efficiency~\citep{chen2025cgr}, bandits with expert advice~\citep{lee2026adaptive}, and $m$-set semi-bandits~\citep{chen2026further}.
\paragraph{Combinatorial semi-bandits.}
Standard approaches to stochastic combinatorial semi-bandits include CUCB \citep{chen2013combinatorial,kveton2015tight} and CTS \citep{wang2018thompson}.
Under matroid constraints, \citet{kveton2014matroid} proposed OMM, which combines upper confidence bounds with greedy basis selection to achieve logarithmic regret.
\citet{talebi2016optimal} developed KL-OSM, which is asymptotically optimal for independent Bernoulli rewards.
Focusing on computational efficiency, \citet{tzeng2024matroid} introduced FasterCUCB, which dynamically maintains approximate maximum-weight bases to accelerate learning for several matroid classes.

In the adversarial setting, online stochastic mirror descent
achieves $O(\sqrt{mdT})$ regret \citep{audibert2014regret}.
FTRL with a hybrid regularizer achieves the same adversarial regret bound together with logarithmic stochastic regret, meaning the BOBW guarantees \citep{zimmert2019beating}.
Subsequent work provides hybrid data-dependent bounds and
sharper stochastic bounds for matroid semi-bandits
\citep{ito2021hybrid}, as well as variance-dependent stochastic
bounds \citep{tsuchiya2023further}.
These FTRL-based methods apply to matroid semi-bandits but
require regularized optimization over the convex hull of the
action set and sampling feasible actions with prescribed marginals.

\section{Preliminaries}

In this part, we first describe the problem setup and then introduce the framework of FTPL with geometric resampling.
In the following, we write $\supp{v}$ for the support of a vector $v$ and
$\supp{\mathcal V}\coloneqq\{\supp{v} : v\in\mathcal V\}$ for the family
of supports of $\mathcal V\subset\{0,1\}^d$.
We use $e_i$ and ${\bf 0}$ to denote the $i$-th unit vector and the zero vector in $\mathbb{R}^d$, respectively. For $E\subset[d]$, we write $\bm{1}_E=\sum_{i\in E}e_i$ for its indicator vector.
Appendix~\ref{app:notation} provides a summary of the notation used in this paper.

\subsection{Matroids \& Matroid Semi-Bandits}

\paragraph{Matroids.} 
A matroid $\sM\triangleq([d],\sI)$ consists of a ground set $[d]$ and a family of independent sets $\sI\subset \{0,1\}^d$ satisfying the following properties: 
(1) ${\bf 0}\in\sI$; 
(2) \textit{hereditary property}: if $\indset_A\in\sI$ and $\supp{\indset_B}\subset\supp{\indset_A}$, then $\indset_B\in\sI$; 
(3) \textit{augmentation property}: If $\indset_A,\indset_B\in\sI$ and $|\supp{\indset_A}|<|\supp{\indset_B}|$, then there exists an element $j\in \supp{\indset_B}\setminus \supp{\indset_A}$ such that $\indset_A+e_j\in\sI$.

Let $r_{\mathcal M}=\max_{\indset\in\sI}\left\lVert \indset\right\rVert_1$ denote the rank of the matroid $\sM$.
A vector $a\in\sI$ is called a basis of $\sM$, if it is maximal, i.e., $\left\lVert a\right\rVert_1= r_{\mathcal M}$.
We denote the family of bases of $\sM$ by $\base$.
In addition, for any $\armset_1,\armset_2\subset [d]$ with $\bm{1}_{E_1}\in\sI$ and $\armset_1\cap\armset_2=\varnothing$, denote
\begin{equation*}
    \basecr{\armset_1}{\armset_2}\coloneqq
    \qty{\indset: \indset+\bm{1}_{E_1}\in\sI,\,\supp{\indset}\subset\armset_2}.
\end{equation*}
In matroid terminology, $(\armset_2,\basecr{\armset_1}{\armset_2})$ is obtained from
$([d],\sI)$ by contracting $\armset_1$ and then restricting to $\armset_2$, which is again a matroid~\citep{oxley2011matroid}.
The contraction of $\armset_1$ treats its elements as already selected,
so that a set disjoint from $\armset_1$ is independent after the
contraction if and only if its union with $\armset_1$ is independent in
the original matroid. The restriction to $\armset_2$ retains only the
independent sets contained in $\armset_2$.
Similarly, we denote by $\neccr{\armset_1}{\armset_2}$ the family of bases of the matroid $(\armset_2,\basecr{\armset_1}{\armset_2})$.
When only contraction or restriction is involved, we write
$\necc{\armset_1}\coloneqq\neccr{\armset_1}{[d]\setminus\armset_1}$ and
$\necr{\armset_2}\coloneqq\neccr{\varnothing}{\armset_2}$, respectively.
For a single base-arm $j\in[d]$, we further write
$\sI_j\coloneqq\basecr{\{j\}}{[d]\setminus\{j\}}$ and
$\nec_j\coloneqq\necc{\{j\}}$.

For any weight vector $\bw\in\mathbb{R}^d$ and a basis family $\base$ of a matroid $\sM=(\armset,\sI)$, the greedy method can find 
\begin{equation*}
    \sele(\bw;\base)\in\argmax\nolimits_{a\in\nec}\left\langle a,\bw\right\rangle,
\end{equation*}
which is a maximum-weight basis of $\sM$.
We denote by $\sele_i(\bw;\base)$ the $i$-th component of this basis.
The procedure of the greedy method is described in Algorithm~\ref{alg:greedy}.

\begin{algorithm}[t]
    \caption{The greedy method for finding a maximum-weight basis of a matroid \citep{edmonds1971matroids}.}
    \label{alg:greedy}
    \begin{algorithmic}[1]
    \STATE {\bfseries Input:} Family of bases $\base$ of a matroid $\sM=(E,\sI)$, 
    weight $\bw\in\mathbb{R}^d$
    \STATE Set $\indset\coloneqq{\bf 0}$
    \STATE Let $\pi(1), \pi(2), \dots, \pi(|\armset|)$ be a permutation of $\armset$ such that $\bw_{\pi(1)}\geq \bw_{\pi(2)}\geq \dots\geq \bw_{\pi(|\armset|)}$
    \FOR{$i=1,2,\dots,|\armset|$}
        \IF{$\indset+e_{\pi(i)}\in \sI$}
            \STATE Set $\indset\coloneqq \indset+e_{\pi(i)}$
        \ENDIF
    \ENDFOR
    \STATE Set $\sele(\bw;\base)\coloneqq\indset$
    \end{algorithmic}
\end{algorithm}

\paragraph{Matroid Semi-Bandits.} We consider a $d$-dimensional action set $\nec\subset\{a\in\{0,1\}^d:\left\lVert a\right\rVert_1=m\}$, where $m$ means the size of each action $a_t\in\nec$, i.e., the number of base-arms selected at each round.
In the matroid semi-bandit problem, there exists a matroid
$\sM=([d],\sI)$ of rank $r_{\mathcal{M}}=m$ such that the action set $\nec$ is exactly the family of bases of $\sM$, i.e.,
\begin{equation*}
    \nec=\{a\in\sI : \left\lVert a\right\rVert_1=m\}.
\end{equation*}
Therefore, by the basis exchange property of matroids, for any $a,a'\in\nec$ and any
$i\in\supp{a}\setminus\supp{a'}$, there exists
$i'\in\supp{a'}\setminus\supp{a}$ such that
$a-e_i+e_{i'}\in\nec$, where $e_i$ denotes the $i$-th unit vector in
$\mathbb{R}^d$.

Given this action set, the interaction between the learner and the environment proceeds over $T$ rounds.
At each round $t\in[T]$, the loss vector $\ell_t\in[0,1]^d$ is generated by the environment, and the learner selects an action $a_t\in\nec$.
The learner then incurs the loss $\langle a_t,\ell_t\rangle$ and observes $\ell_{t,i}$ for each $i\in\mathrm{supp}(a_t)$, which is referred to as semi-bandit feedback.

The stochastic and adversarial settings characterize two fundamentally different ways in which the environment generates the loss vectors.
In the stochastic setting, the loss vectors $\{\ell_t\}_{t=1}^T$ are i.i.d. over time from a fixed but unknown distribution with mean $\mu\in[0,1]^d$. 
Then, the suboptimality gap of base-arm $i$ is defined as $\gap{i}=\min_{a\in\nec\setminus\{a^*\},a_i=1}(a-a^*)^\top \mu$, where $a^*\in\argmin_{a\in\nec}a^\top \mu$ denotes the unique optimal action.
In the adversarial setting, the environment may choose the loss vectors $\{\ell_t\}_{t=1}^T$ arbitrarily, possibly depending on the past history ${(\ell_s,a_s)}_{s=1}^{t-1}$.

The performance of the learner is measured by the pseudo-regret, which is defined as
\begin{equation*}
    \mathcal{R}(T)=\E\qty[\sum_{t=1}^{T}\left\langle \ell_t,a_t-a^*\right\rangle],
\end{equation*}
where $a^*\in\argmin_{a\in\nec}\E\qty[\sum_{t=1}^{T}\left\langle \ell_t,a\right\rangle]$ denotes the optimal action in hindsight.
We also write $A^*\coloneqq\supp{a^*}$.

\subsection{Follow-the-Perturbed-Leader}
\begin{algorithm}[t]
    \caption{Follow-the-Perturbed-Leader}
    \label{alg:FTPL}
    \begin{algorithmic}[1]
    \STATE {\bfseries Input:} Action set $\nec$, learning rate $\eta_t$, distribution $\mathcal{D}$
    \STATE Set $\hat{L}_1 \coloneqq \mathbf{0} \in \mathbb{R}^d$
    \FOR{$t = 1, 2, \dots, T$}
        \STATE Sample $r_t = (r_{t,1}, r_{t,2}, \dots, r_{t,d})$ i.i.d. from $\mathcal{D}$\label{line:perturbation}
        \STATE Choose action $a_t = {\argmin}_{a \in \nec} \langle a,\eta_t\hat L_t-r_t\rangle$ and observe $\{\ell_{t,i}:a_{t,i}=1\}$.\label{line:selection}
        \STATE Compute an estimator $\widehat{w_{t,i}^{-1}}$ for $w_{t,i}^{-1}$ by geometric resampling for all $i$ with $a_{t,i}=1$.
        \STATE Set $\hat{\ell}_t\coloneqq\sum_{i:a_{t,i}=1}\ell_{t,i}\widehat{w_{t,i}^{-1}}e_i$ and $\hat{L}_{t+1}\coloneqq\hat{L}_t+\hat{\ell}_t$.
    \ENDFOR
    \end{algorithmic}
\end{algorithm}

The procedure of FTPL is described in Algorithm~\ref{alg:FTPL}.
Since the learner can only observe the losses of the selected base-arms $\{\ell_{t,i}:a_{t,i}=1\}$ in the semi-bandit setting, FTPL and FTRL always require an estimator $\hat{\ell}_t$ for the loss vector $\ell_t$, which is specified in Section~\ref{subsec:GR}.
At each round $t$, FTPL maintains a cumulative estimated loss vector $\hat{L}_t\in\mathbb{R}^d$, and draws a perturbation vector $r_t=(r_{t,1},\ldots,r_{t,d})\in\mathbb{R}^d$ whose entries are i.i.d. from some distribution.
Then, FTPL selects an action
\begin{equation*}
    a_t=\argmin_{a\in\nec}\langle a,\eta_t\hat L_t-r_t\rangle,
\end{equation*}
where $\eta_t\in\mathbb{R}^+$ is the learning rate.
In this paper, we consider two perturbation distributions, both of which belong to a class of heavy-tailed distributions called Fr\'{e}chet-type distributions.
The first is the Fr\'{e}chet distribution $\Fdis$ with parameter $\alpha>1$, whose probability density function $f(x)$ and cumulative distribution function $F(x)$ are given by
\begin{equation}\label{eq:frechet_def}
    f(x)=\alpha x^{-(\alpha+1)}e^{-1/x^{\alpha}},\;
    F(x)=e^{-1/x^{\alpha}},\; x\geq 0.
\end{equation}
The second is the Pareto distribution $\Pdis$ with parameter $\alpha>1$, whose density and cumulative distribution function are
\begin{equation}\label{eq:pareto_def}
    f(x)=\alpha x^{-(\alpha+1)},\;
    F(x)=1-x^{-\alpha},\; x\geq 1.
\end{equation}

Now we consider the computational complexity of action decision in FTPL.
At each round, FTPL first generates a $d$-dimensional perturbation $r_t$, which requires $O(d)$ time, and then applies the greedy algorithm to the perturbed cumulative losses, i.e., finding
$\sele(r_t-\eta_t\hat{L}_t;\nec)$ by Algorithm~\ref{alg:greedy}.
Therefore, if an independence test takes $\mathcal{T}_{\mathrm{ind}}$ time, the complexity of each action selection lies between $\Omega(d)$ and $O\qty(d(\log d+\mathcal{T}_{\mathrm{ind}}))$.
For $\armset\subset[d]\setminus\{i\}$, we write $\armset\rightsquigarrow_{\nec}i$
if the greedy algorithm on $\nec$ selects $i$ when exactly the base-arms in $\armset$ precede it, and $\armset\not\rightsquigarrow_{\nec}i$ otherwise.
By Lemma~\ref{lem:greedy_equiv}, the ordering within $\armset$ does not affect whether $i$ is selected, and thus this notation is well-defined.
We establish this and other structural properties necessary for our analysis in Appendix~\ref{sec:structural_properties}.

\subsection{Geometric Resampling}\label{subsec:GR}
\begin{algorithm}[t]
    \caption{Geometric Resampling}
    \label{alg:GR}
    \begin{algorithmic}[1]
        \STATE \textbf{Input:}
        Chosen action $a_t$, action set $\nec$, learning rate $\eta_t$,\\ cumulative estimated loss $\hat{L}_t$, distribution $\mathcal{D}$
        \STATE Set $M_t \coloneqq \bm{0}\in\mathbb{R}^d$; $s\coloneqq a_t$
        \REPEAT
            \STATE $M_t \coloneqq M_t + s$\;
            \STATE Sample $r_t^{\prime} = (r_{t,1}^{\prime}, r_{t,2}^{\prime}, \dots, r_{t,d}^{\prime})$ i.i.d. from $\mathcal{D}$\;
            \STATE $a'_t = \argmin_{a \in \nec} \langle a,\eta_t\hat L_t-r_t'\rangle$\;
            \STATE $s\coloneqq s\circ\qty(\bm{1}_{[d]}-a'_t)$\;
        \UNTIL{$s=\bm{0}$}
        \STATE Set $\widehat{w_{t,i}^{-1}} \coloneqq M_{t,i}$ for all $i$ with $a_{t,i}=1$
    \end{algorithmic}
\end{algorithm}

Under semi-bandit feedback, only the losses of the selected base-arms are
revealed.
Therefore, most policies replace the unobserved loss vector $\ell_t$ by
an unbiased (or low-bias) estimator $\hat\ell_t$ and make decisions based
on the cumulative estimate $\hat L_t\coloneqq\sum_{s=1}^{t-1}\hat\ell_s$.
A standard choice is the importance-weighted (IW) estimator
$\hat{\ell}_t=\sum_{i:a_i=1}\qty(\ell_{t,i}/w_{t,i})e_{i}$,
where $w_{t,i}$ denotes the probability that base-arm $i$ is selected at
round $t$. This estimator is readily implementable for policies that
compute $w_t$ explicitly, such as FTRL~\citep{zimmert2019beating}.
For FTPL, however, $w_{t,i}$ is not explicitly available.
To address this issue, \citet{JMLR:v17:15-091} proposed geometric
resampling (GR), which estimates $w_{t,i}^{-1}$ directly instead of
computing $w_{t,i}$. Specifically, GR repeatedly draws a fresh perturbation
$r_t'\sim\mathcal{D}$ until
$\argmin_{a\in\nec}\langle a,\eta_t\hat L_t-r_t'\rangle$ 
again includes base-arm $i$.
Since each resampling selects $i$ independently with probability
$w_{t,i}$, the number of resamplings follows a geometric distribution with mean
$w_{t,i}^{-1}$, and thus serves as an unbiased estimator
$\widehat{w_{t,i}^{-1}}$ of $w_{t,i}^{-1}$.
For matroid semi-bandits, each resampling amounts to a single execution of the FTPL decision rule, which involves
sampling a fresh perturbation and then running
the greedy algorithm.
Therefore, each resampling also requires at least $\Omega(d)$ and at most $O(d(\log d+\mathcal{T}_{\mathrm{ind}}))$ time.
Furthermore, the number of resamplings $S_t=\max_{i:a_{t,i}=1}M_{t,i}$ under GR satisfies $\mathbb{E}[S_t|\hat{L}_t]\leq d$ \citep[Proposition~3]{JMLR:v17:15-091}.
Consequently, the average complexity of GR at each round can be bounded as
\begin{multline*}
    C_{\text{GR}}=\mathbb{E}[S_t|\hat{L}_t]\cdot O(d(\log d+\mathcal{T}_{\mathrm{ind}}))\\
    \leq O(d^2(\log d+\mathcal{T}_{\mathrm{ind}})),
\end{multline*}
which is independent of $\hat{L}_t$.
The procedure is summarized in Algorithm~\ref{alg:GR}, where 
$\bm{1}_{[d]}$ denotes the $d$-dimensional all-ones vector, and 
$a\circ b$ denotes the element-wise product of vectors $a$ and $b$, i.e., $(a\circ b)_i=a_ib_i$.

\section{Main Results}

We first show that FTPL with either Fr\'{e}chet or Pareto perturbations and geometric resampling achieves minimax-optimal regret in the adversarial setting.
The proof is provided in Appendix~\ref{sec:proof_adv_bound}.

\begin{theorem}\label{thm:adv_bound}
    In the adversarial setting, for learning rate $\eta_t=\frac{c}{\sqrt{t}}m^{\frac{1}{2}-\frac{1}{\alpha}}d^{\frac{1}{\alpha}-\frac{1}{2}}$,
    FTPL with geometric resampling and perturbations following $\Fdis$ or $\Pdis$ with $\alpha>1$ achieves
    \begin{equation*}
        \R(T)\leq O\qty(\sqrt{mdT}).
    \end{equation*}
\end{theorem}

In the stochastic setting, choosing $\alpha=2$ yields the following gap-dependent logarithmic regret bound for both Fr\'{e}chet and Pareto perturbations.

\begin{theorem}\label{thm:sto_bound} 
    Assume that $a^*=\argmin_{a\in\mathcal{A}}a^\top \mu$ is unique.
    Let $\Delta_{i,\min}=\min_{a\in\nec\setminus\{a^*\},\,a_i=1}\left\langle \mu,a-a^*\right\rangle $ and $\Delta_{\min}=\min_{i:a^*_i=0}\Delta_{i,\min}$.
    Then, FTPL with perturbations following $\mathcal{F}_2$ or $\mathcal{P}_2$ and learning rate $\eta_t=\frac{c}{\sqrt{t}}$ for some constant $c>0$ satisfies
    \begin{equation*}
        \R(T)\leq O\qty(\sum_{i:a^*_i=0}\frac{\log T}{\Delta_{i,\min}})+O\qty(\frac{m^3d}{\Delta_{\min}}).
    \end{equation*}
\end{theorem}
Together with Theorem~\ref{thm:adv_bound}, this establishes BOBW guarantees for FTPL with both Fr\'{e}chet and Pareto perturbations.
For other shape parameters $\alpha\in(1,2)\cup(2,\infty)$, the following theorem quantifies the stochastic regret guarantees.

\begin{theorem}\label{thm:sto_bound_alpha_not2}
    Assume that $a^*=\argmin_{a\in\mathcal{A}}a^\top \mu$ is unique.
    Let $\Delta_{i,\min}=\min_{a\in\nec\setminus\{a^*\},\,a_i=1}\left\langle \mu,a-a^*\right\rangle $ and $\Delta_{\min}=\min_{i:a^*_i=0}\Delta_{i,\min}$.
    Then, FTPL with perturbations following $\mathcal{F}_{\alpha}$ or $\mathcal{P}_{\alpha}$ and learning rate $\eta_t=\frac{c}{\sqrt{t}}m^{\frac{1}{2}-\frac{1}{\alpha}}d^{\frac{1}{\alpha}-\frac{1}{2}}$ for $c>0$ and $\alpha>2$ satisfies
    \begin{equation*}
        \R(T)\leq O\qty(\sum_{i:a^*_i=0}\frac{1}{\alpha-2}\frac{T^{\frac{\alpha-2}{2(\alpha-1)}}}{\Delta_{i,\min}^{\frac{1}{\alpha-1}}m^{\frac{\alpha-2}{2(\alpha-1)}}d^{\frac{2-\alpha}{2(\alpha-1)}}}).
    \end{equation*}
    If $\alpha\in(1,2)$, then
    \begin{equation*}
        \R(T)\leq O\qty(\sum_{i:a^*_i=0}\frac{1}{2-\alpha}\frac{T^{1-\frac{\alpha}{2}}}{\Delta_{i,\min}^{\alpha-1} m^{\frac{\alpha}{2}-1} d^{1-\frac{\alpha}{2}}}).
    \end{equation*}
\end{theorem}

For every $\alpha\in(1,2)\cup(2,\infty)$, these stochastic regret bounds have a better dependence on the horizon $T$ than the $O(\sqrt{T})$ adversarial rate, since the exponents of $T$ in both cases are strictly below $1/2$.
We outline the stochastic regret analysis in Section~\ref{subsec:outline} and provide the complete proofs of Theorems~\ref{thm:sto_bound} and~\ref{thm:sto_bound_alpha_not2} in Appendix~\ref{subsec:proofs_sto_bounds}.

\section{Regret Analysis}

We begin by introducing some key tools for the regret analysis.
For $\bm{u}\in\mathbb{R}^d$, $\armset\subset[d]$, we define $\sigma_i(\bm{u},\armset)$
as the position of $i$ when the entries $\{u_k\}_{k\in\armset}$ are sorted in descending order (ties are broken arbitrarily).
We omit $\bm{u}$ (resp.~$\armset$) when $\bm{u}=-\eta_t\hat L_t$ (resp.~$\armset=[d]$).
For example, the current best base-arm $\hat i_t^*$, i.e., the base-arm
with the smallest component of the cumulative estimated loss, satisfies
$\sigma_{\hat i_t^*}=1$.

The following definition yields a necessary condition for selecting base-arm $i$.
For any $\lambda\in\mathbb{R}^d$ and $i\in[d]$, we define
\begin{equation}\label{eq:def_blk}
    \blk{i}{\lambda}
    \coloneqq
    \argmin_{
        j\in[d]\setminus\{i\}:
        (\armset_{\preceq j}(\lambda)\setminus\{i\})
        \not\rightsquigarrow_{\nec}i
    }
    \sigma_j(-\lambda),
\end{equation}
where $\armset_{\preceq j}(\lambda)
    \coloneqq \qty{j'\in[d]:\sigma_{j'}(-\lambda)\leq\sigma_j(-\lambda)}$
is the set of base-arms up to and including $j$ in the order $\sigma(-\lambda)$.
Then, after removing $i$ from this order, $\blk{i}{\lambda}$ is the first
base-arm $j$
such that inserting $i$ immediately after $j$ would cause
the greedy algorithm to reject $i$\footnote{If $i\in\supp{a}$ for every $a\in\nec$, no such $j$ exists.
Therefore, it contributes zero to regret and we need not define
$\blk{i}{\lambda}$ for it.}.
We simply write $\blkomit{t,i}=\blk{i}{\eta_t\hat{L}_t}$.
With this definition, consider the extreme case where the perturbations of all base-arms other than $i$ take their minimum value $\nu$.
Then, $i$ is selected if
\begin{equation}\label{eq:r_start}
    r_{t,i}
    >
    \nu+\eta_t\hat{L}_{t,i}-\eta_t\hat{L}_{t,\blkomit{t,i}},
\end{equation}
and is not selected if the reverse strict inequality holds, while the case of equality
depends on the tie-breaking rule.
For example, in $m$-set semi-bandits, $\blkomit{t,i}$ is the base-arm with the $m$-th smallest component of the cumulative estimated loss $\hat{L}_{t}$ whenever $\sigma_i\geq m$.
We then write $\ul^{(i)}=\lambda-\bm{1}_{[d]} \lambda_{\blk{i}{\lambda}}$, and simply write $\ul_i=\ul^{(i)}_i$.

The probability vector of base-arm selection given $\hat{L}_t$ and $\nec$ is denoted by $w_t=\phi(\eta_t\hat{L}_t;\dis,\nec)=(\phi_1(\eta_t\hat{L}_t;\dis,\nec),\allowbreak\cdots,\allowbreak\phi_d(\eta_t\hat{L}_t;\dis,\nec))$, where for $\lambda\in\mathbb{R}^d$ we define
\begin{align*}
        \lefteqn{\phi_i(\lambda;\dis,\nec)}\\
        &=\!\!\int_{\nu-\lambda_{\blk{i}{\lambda}}}^\infty
        \sP_{\substack{r_{-i} \sim \dis\\r_i=z+\lambda_i}}
        \qty[\sele_i(r-\lambda;\nec)=1]
    \dd F(z+\lambda_i)
    \numberthis\label{eq:definition_phi_extension_no_underline}\\
    &=\!\!\int_{\nu}^\infty
        \sP_{\substack{r_{-i} \sim \dis\\r_i=z+\ul_i}}
        \qty[\sele_i(r-\ul\pn{i};\nec)=1]
    \dd F(z+\ul_i),
\end{align*}
where we write $r_{-i}\coloneqq\{r_k\}_{k\in[d]\setminus\{i\}}$ for brevity.
When referring to the original action set of the problem, we omit $\nec$ and simply write $\phi_i(\lambda;\dis)$.
Deriving an explicit expression for $\phi(\cdot)$ is more involved under matroid constraints than in MAB~\citep{pmlr-v201-honda23a} and $m$-set semi-bandits~\citep{chen2026further}.
We provide its derivation in Appendix~\ref{app:expression} as it may be of independent interest, although it is not used in our regret analysis.

\subsection{Regret Decomposition}

We begin with the following regret decomposition, which is identical to that used for $m$-set semi-bandits~\citep{chen2026further}. Since its derivation does not rely on the specific structure of the $m$-set selection, the same result applies directly to the matroid semi-bandit setting.

\begin{lemma}[regret decomposition]\label{lem:regret_decomposition}
    For any $\alpha>1$ we have
    \begin{align*}
        &\R(T) \leq \sum_{t=1}^{T} \mathbb{E} \qty[ \left\langle \hat{\ell}_t, \phi(\eta_t \hat{L}_t) - \phi(\eta_t (\hat{L}_t + \hat{\ell}_t)) \right\rangle ]  \\
        &+ \sum_{t=1}^T \qty(\frac{1}{\eta_{t+1}}-\frac{1}{\eta_t}) \E_{r_{t+1}\sim\dis}\qty[\left\langle r_{t+1}, a_{t+1} - a^*\right\rangle]\\
        &+ \frac{1}{\eta_1}\qty(\frac{\alpha}{\alpha-1}(m-1)^{1-\frac{1}{\alpha}}+\Gamma(1-\frac{1}{\alpha}))(d+1)^{\frac{1}{\alpha}},\numberthis\label{eq:regret_decomposition}
    \end{align*}
    where $\Gamma(\cdot)$ is the gamma function.
\end{lemma}

Consistent with the standard terminology in the BOBW literature \citep{zimmert2019beating,zimmert2021tsallis,pmlr-v201-honda23a,pmlr-v247-lee24a,chen2026further}, we call the first and second terms in \eqref{eq:regret_decomposition} the \textit{stability term} and \textit{penalty term}, respectively.

We provide the upper bound for the penalty term in the following lemma.
The proof for the penalty term follows that for $m$-set semi-bandits \citep{chen2026further}, except that a new construction of $\underline{\hat{L}}_{t,i}=\hat{L}_{t,i}-\hat{L}_{t,\blkomit{t,i}}$ is required for the matroid semi-bandit setting.
Appendix~\ref{sec:penalty_term} provides a version with explicit constants and its proof.

\begin{lemma}[penalty term]\label{lem:penalty_adv_bound}
    For any $\hat{L}_t\in\mathbb{R}^d$ and $\alpha>1$ we have
    \begin{equation*}
        \E_{r_t\sim\dis}\qty[\left\langle r_t, a_t - a^*\right\rangle\m\hat{L}_t]
        \leq O\qty(m^{1-\frac{1}{\alpha}}d^{\frac{1}{\alpha}}).
    \end{equation*}
    In addition, if $\underline{\hat{L}}_{t,i}>0$ for any $i$ with $a^*_i=0$, we also have
    \begin{equation*}
        \E_{r_t\sim\dis}\qty[\left\langle r_t, a_t - a^*\right\rangle\m\hat{L}_t]
        \leq
        \sum_{i:a^*_i=0}\frac{\alpha/(\alpha-1)}{(\eta_t\underline{\hat{L}}_{t,i})^{\alpha-1}}.
    \end{equation*} 
\end{lemma}

\subsection{Stability of Base-Arm Selection Probability}

Before providing the bounds on the stability term, we first introduce the following definitions, which are crucial for the analysis.
For any $\lambda\in\mathbb{R}^d$ and matroid $\sM$, define
\begin{equation*}
    I_i(\lambda;\dis,\nec)=\sP_{r\sim \dis}\qty[\sele_i(r-\lambda;\nec)=1,r_i\geq\nu+\lambda_i].
\end{equation*}
Combining this with the definition of $\phi_i(\cdot)$ and \eqref{eq:r_start} we have
\begin{equation}\label{eq:phi_I_relation}
    I_i(\ul\pn{i};\dis,\nec)=\phi_i(\lambda;\dis,\nec)
\end{equation}
To analyze the derivative, we additionally define
\begin{multline}\label{eq:definition_J}
    J_i(\lambda;\dis,\nec)
    \\
    =\E_{r\sim \dis}\qty[\frac{1}{r_i}\ind\qty[\sele_i(r-\lambda;\nec)=1,r_i\geq\nu+\lambda_i]].
\end{multline}
Based on these definitions, we can bound the stability term in the following lemmas.
In Lemmas~\ref{lem:lambda_star} and \ref{lem:sigma_i} and their proofs in Appendix~\ref{subsec:proof_lemma_lambda_star}--\ref{sec:proof_adv_bound_term}, without loss of generality, we assume $\lambda_1\leq\lambda_2\leq\cdots\leq\lambda_d$ (ties are broken arbitrarily) so that $\sigma_i=i$ for notational simplicity.
\begin{lemma}\label{lem:lambda_star}
    For any $i\in[d]$ and $\lambda\in\mathbb{R}^d$, it holds that
    \begin{equation*}
        \frac{J_i(\ul\pn{i};\dis)}{I_i(\ul\pn{i};\dis)}
        \leq
        \max_{\bm{1}_{\armset^+}\in\sI_i}
        \qty{\frac{J_i(\lambda^*;\dis,\neccr{\armset^+}{[i]\setminus\armset^+})}{I_i(\lambda^*;\dis,\neccr{\armset^+}{[i]\setminus\armset^+})}},
    \end{equation*}
    where we denote
    \begin{equation*}
        \lambda_k^* = 
        \begin{cases}
            \ul_i, & \text{if } k\leq i,\\
            \ul_k\pn{i}, & \text{if } k> i.
        \end{cases}
    \end{equation*}
\end{lemma}

Lemma~\ref{lem:lambda_star} bounds the original ratio $J_i(\ul\pn{i};\cdot)/I_i(\ul\pn{i};\cdot)$ by the
maximum of the ratios corresponding to the reduced problems, where the entries of $\lams$ on the base-arm set $[i]\setminus\armset^+$ are identical.
Therefore, it suffices to consider such ratios with simpler structures, leading to the following bounds.

\begin{lemma}\label{lem:sigma_i}
    For any $\alpha>1$ and $i\in[d]$ we have
    \begin{equation*}
        \frac{J_i(\ul\pn{i};\dis)}{I_i(\ul\pn{i};\dis)}
        \leq
        \frac{1}{0\lor\ul_i}
        \land
        \qty(\phi_i(\lams;\dis,\nec|[i]))^{\frac{1}{\alpha}}.
    \end{equation*}
    Moreover,
    \begin{equation}\label{eq:sum_phi_bound}
        \sum_{i\in[d]}\qty(\phi_i(\lams;\dis,\nec|[i]))^{\frac{1}{\alpha}}
        \leq 
        \frac{\alpha+1}{\alpha-1}
        m^{\frac{1}{\alpha}}d^{1-\frac{1}{\alpha}},
    \end{equation}
    where $\lams$ in each $\phi_i(\cdot)$ defined in Lemma~\ref{lem:lambda_star} depends on $i$.
\end{lemma}

The bound by $\phi_i(\lams;\cdot)^{1/\alpha}$, together with the aggregate bound in \eqref{eq:sum_phi_bound}, addresses a central difficulty in the stability analysis for general matroid semi-bandits in adversarial setting.
Notably, the discussion of \citet{pmlr-v201-honda23a} provides useful intuition for the appearance of $\phi_i(\lams;\cdot)^{1/\alpha}$ in this bound.
For MAB with perturbation following $\mathcal{F}_2$, they interpret $J_i(\cdot)$ as informally playing a role analogous to $O(\phi_i^{3/2})$, which motivates us to construct a novel bound on $J_i(\cdot)/I_i(\cdot)$ through fractional powers of $\phi_i(\cdot)$.
Our bound is consistent with this intuition for $\alpha=2$ and generalizes it to both $\Fdis$ and $\Pdis$ with arbitrary $\alpha>1$.

In fact, for the analysis of MAB and $m$-set semi-bandits, the stability analysis does not require a bound in this more general form, since the symmetry among base-arms $[i]$ in the simplified problems with equal losses directly yields bounds of order $O((1\land m/i)^{1/\alpha})$~\citep{pmlr-v201-honda23a,pmlr-v247-lee24a, chen2026further}.
Under general matroid constraints, however, equal losses need not imply equal selection probabilities.
Our bound in terms of $\phi_i(\lams;\cdot)^{1/\alpha}$ flexibly accounts for how
the matroid constraints affect each base-arm, thereby overcoming this obstacle.

Finally, we bound the stability term in the following lemma, and the proof is provided in Appendix~\ref{sec:proof_adv_bound_term}.
\begin{lemma}\label{lem:adv_bound_term}
    For any $\alpha>1$ and $i\in[d]$, it holds that
    \begin{equation*}
        \E\qty[\hat{\ell}_{t,i}\qty(\phi_i(\eta_t \hat{L}_t;\dis)\!-\!\phi_i(\eta_t \hat{L}_{t+1};\dis))\m \hat{L}_t]
        \!\leq
        \frac{2(\alpha+1)}{0\lor\underline{\hat{L}}_{t,i}},
    \end{equation*}
    where we use the convention $1/0=+\infty$.
    Moreover,
    \begin{multline*}
        \E\qty[\bigl\langle\hat\ell_t,
        \phi(\eta_t\hat L_t;\dis)-\phi(\eta_t\hat L_{t+1};\dis)\bigr\rangle\m \hat{L}_t]
        \\
        \leq
        \frac{2(\alpha+1)^2}{\alpha-1}
        \eta_t m^{\frac{1}{\alpha}}d^{1-\frac{1}{\alpha}}.
    \end{multline*}
\end{lemma}

\subsection{Outline for Stochastic Regret Bound}\label{subsec:outline}

In this part, we outline the derivation of the regret bound in the stochastic setting, which in particular becomes logarithmic when $\alpha=2$. 
Although the proof treats $\alpha\geq2$ and $\alpha\in(1,2)$ separately,
both cases follow essentially the same argument. 
Therefore, we present only the case $\alpha\geq 2$ here and defer the case $\alpha\in(1,2)$ to the complete proof.

We first lower-bound the regret as follows.
\begin{equation*}
    \R(T)\geq\E\qty[\sum_{t=1}^T \sum_{i:a^*_i=0}w_{t,i}\Delta_{i,\min}].
\end{equation*}
Extending the techniques used in MAB~\citep{pmlr-v201-honda23a,pmlr-v247-lee24a} and $m$-set semi-bandits~\citep{chen2026further}, we introduce events $F_t$ and $D_t$ for Fr\'{e}chet and Pareto perturbations, respectively, with their formal definitions given in Appendix~\ref{sec:stab_sto}. 
Both events ensure that $\eta_t\underline{\hat{L}}_{t,i}$ is sufficiently large for every suboptimal base-arm $i$. This quantity measures the ``cumulative loss disadvantage '' that the perturbation $r_{t,i}$ must, at a minimum, overcome for $i$ to be selected.
For concreteness, we present the outline for $\Fdis$ using $F_t$, and the analysis for $\Pdis$ follows the same overall argument with $D_t$.
The main difficulty in extending the analysis of \citet{chen2026further} is that key selection properties used to establish both the upper and lower regret bounds for $m$-set semi-bandits no longer hold in general matroids. In particular, whether a base-arm is selected depends on the identities of the preceding base-arms, rather than only on their number. For the upper bound, Lemmas~\ref{lem:optimal_action_add} and \ref{lem:optimal_action_combine} show that, on $F_t$, the stability contribution of the optimal action can be bounded in terms of statistics of the suboptimal base-arms.
Then, we can obtain the upper bound for the regret as
\begin{equation*}
    \R(T)\leq
    \E\qty[\sum_{t=1}^T O\qty(\ind\qty[F_t]\sum_{i:a^*_i=0}\frac{1}{\underline{\hat{L}}_{t,i}} + \ind\qty[F_t^c]\sqrt{\frac{md}{t}})].
\end{equation*}
Next, we derive a regret lower bound by separately analyzing the cases where $F_t$ and $F_t^c$ occur.
The main difficulty is to identify a tractable sufficient condition under which each base-arm $i$ is selected. 
We establish such a condition in Lemmas~\ref{lem:optimal_action_psi_ex} and \ref{lem:optimal_action_sufficient_condition}, which allow us to recover the desired lower bound as in \citet{chen2026further}.
Then, we obtain a regret lower bound expressed as 
\begin{multline*}
    \R(T)\geq
    \E\Bigg[\sum_{t=1}^T \Omega\Bigg(\ind\qty[F_t]\sum_{i:a^*_i=0}\frac{\gap{i} t^\frac{\alpha}{2}}{m^{\frac{\alpha}{2}-1}d^{1-\frac{\alpha}{2}}\underline{\hat{L}}_{t,i}^\alpha}
    \\
    + \ind\qty[F_t^c]\frac{\Delta_{\min}}{m^{\frac{\alpha}{\alpha-1}}}\Bigg)\Bigg],
\end{multline*}
where $\Delta_{\min}=\min_{i:a^*_i=0}\gap{i}$ and $\Omega$ denotes the big-Omega notation.

Combining these results yields the bound
\begin{align*}
    \frac{\R(T)}{2}\leq
    \E\Bigg[\sum_{t=1}^T O\Bigg(\ind\qty[F_t]\sum_{i:a^*_i=0}\qty(\frac{1}{\underline{\hat{L}}_{t,i}}-\frac{\Delta_{i,\min} t^\frac{\alpha}{2}}{2\frac{m}{d}^{\frac{\alpha}{2}-1}\underline{\hat{L}}_{t,i}^\alpha})
    \\
    + \ind\qty[F_t^c]\qty(\sqrt{\frac{md}{t}}-\frac{\Delta_{\min}}{2m^{\frac{\alpha}{\alpha-1}}})\Bigg)\Bigg].
\end{align*}
Applying the self-bounding technique to this gives
\begin{align*}
    \R(T)\leq
    \sum_{t=1}^TO\qty(\sum_{i:a^*_i=0}\frac{1}{
        \gap{i}^{\frac{1}{\alpha-1}}
        m^{\frac{\alpha-2}{2(\alpha-1)}}
        d^{\frac{2-\alpha}{2(\alpha-1)}}
        t^{\frac{\alpha}{2(\alpha-1)}}
        })\\
    +O\qty(\frac{m^{\frac{2\alpha-1}{\alpha-1}}d}{\Delta_{\min}}),
\end{align*}
which reduces to a logarithmic regret bound when $\alpha=2$.

\section{Conclusion}
We studied the optimality of FTPL with geometric resampling and Fr\'{e}chet or Pareto perturbations with shape parameter $\alpha>1$ in general matroid semi-bandits. To the best of our knowledge, this is the first algorithm to achieve minimax-optimal $O(\sqrt{mdT})$ adversarial regret for general matroid semi-bandits using a greedy optimization oracle for action selection. 
Moreover, FTPL also achieves logarithmic stochastic regret with both perturbations with shape parameter $\alpha=2$, thereby establishing the BOBW guarantees.
A natural direction for future work is to reduce the computational cost without sacrificing the regret guarantees, for example, by adapting conditional geometric resampling~\citep{chen2025cgr,chen2026further} to different matroid constraints.

\bibliography{references}
\bibliographystyle{icml2026}

\newpage
\appendix
\onecolumn

\section{Notation}\label{app:notation}

\begingroup
\small
\renewcommand{\arraystretch}{1.15}
\setlength{\tabcolsep}{6pt}
\begin{longtable}{@{}p{0.31\textwidth}|p{0.65\textwidth}@{}}
\caption{Summary of notation.}\label{tab:notation}\\
\toprule
\endfirsthead
\multicolumn{2}{l}{\tablename~\thetable\ (continued)}\\
\toprule
\textbf{Symbol} & \textbf{Meaning}\\
\midrule
\endhead
\midrule
\multicolumn{2}{r}{Continued on next page}\\
\endfoot
\bottomrule
\endlastfoot
\multicolumn{2}{@{}l@{}}{\textbf{Problem setting and matroid structure}}\\*
\midrule
$\sM=(\armset,\sI)$ & Matroid with ground set $\armset$ and independent-set family $\sI$.\\
$\nec\subset\{0,1\}^d$ & Action set (i.e., the family of bases of $\sM$).\\
$d\in \mathbb{N}$ & Dimensionality of action set.\\
$m\in[d]$ & $m = \|a\|_1$ for any $a\in\nec$, i.e., the size of each action.\\
$T\in \mathbb{N}$ & Time horizon.\\
$\armset\subset[d]$ & Base-arm set.\\
$\basecr{\armset_1}{\armset_2}$ & Independent-set family after contracting $\armset_1$ and restricting to $\armset_2$.\\
$\neccr{\armset_1}{\armset_2}$ & Basis family of $\basecr{\armset_1}{\armset_2}$.\\
$\necc{\armset_1},\ \necr{\armset_2}$ & Basis families under contraction and restriction, respectively.\\
$\sI_j,\ \nec_j$ & Abbreviations for $\basec{\{j\}}$ and $\necc{\{j\}}$, respectively.\\
$\sele(\bw;\nec)=(\sele_1(\bw;\nec),\ldots,\sele_d(\bw;\nec))$ & Maximum-weight basis in $\nec$ with weights $\bw$.\\
$\armset\rightsquigarrow_{\nec}i$ & The greedy algorithm on $\nec$ selects $i$ when exactly the base-arms in $\armset$ precede it.\\
$\bm{1}_E=\sum_{i\in E}e_i$ & Indicator vector of a set $E\subset[d]$.\\
$\supp{a},\, \supp{\mathcal V}$ & Support of a vector $a$ and collection $\{\supp{v}:v\in\mathcal V\}$ for a family $\mathcal V\subset\{0,1\}^d$, respectively.\\
$a^*,\,A^*$ & Optimal action and its support, respectively.\\
$\ell_t\in[0,1]^d$ & Loss vector.\\
\midrule
\multicolumn{2}{@{}l@{}}{\textbf{Follow-the-Perturbed-Leader and perturbation distributions}}\\*
\midrule
$\hat\ell_t\in[0,+\infty)^d$ & Estimated loss vector.\\
$\hat L_t=\sum_{s=1}^{t-1}\hat{\ell}_s\in[0,\infty)^d$ & Cumulative estimated loss vector.\\
$\eta_t>0$ & Learning rate.\\
$\nu\in\mathbb{R}$ & Left endpoint of the perturbation distribution.\\
$r_t\in [\nu,\infty)^d$ & Perturbation vector.\\
$f(x)$ & Probability density function of perturbation. \\
$F(x)$ & Cumulative distribution function of perturbation. \\
$\Fdis$ & Fr\'{e}chet perturbation with shape parameter $\alpha>1$ (See~\eqref{eq:frechet_def}).\\
$\Pdis$ & Pareto perturbation with shape parameter $\alpha>1$ (See~\eqref{eq:pareto_def}).\\
\midrule
\multicolumn{2}{@{}l@{}}{\textbf{Regret analysis}}\\*
\midrule
$\phi_i(\lambda;\dis,\nec)$ & Probability that $\sele_i(r-\lambda;\nec)=1$ for fixed $\lambda$ and $\nec$, where $r\sim\dis$.\\
$w_t=\phi(\eta_t\hat L_t;\dis)\in[0,1]^d$ & Base-arm selection probability.\\
$\blk{i}{\lambda},\ \blkomit{t,i}\in[d]$ & Base-arm defined in~\eqref{eq:def_blk}, with $\blkomit{t,i}=\blk{i}{\eta_t\hat L_t}$.\\
$\ul\pn{i},\, \ul_i$ & Shifted vector $\lambda-\bm1_{[d]}\lambda_{\blk{i}{\lambda}}$ and its $i$-th component, i.e., $\ul\pn{i}_i$, respectively.\\
$\sigma_i(\bm u,\armset)$ & Position of $i$ in the descending order of $\{u_k\}_{k\in\armset}$, with ties broken arbitrarily.
$\bm{u}$ (resp. $\armset$) is omitted when $\bm{u}=-\eta_t\hat{L}_t$ (resp. $\armset=[d]$).\\
\end{longtable}
\endgroup

\section{Structural Properties of Base-Arm Selection under Matroid Constraints}
\label{sec:structural_properties}

We first establish several structural properties of the base-arm selection rule under matroid constraints in Lemmas~\ref{lem:greedy_num}--\ref{lem:greedy_remove}.

\begin{lemma}\label{lem:greedy_num}
    Let $\bw,\bw'\in\mathbb{R}^d$ be two weight vectors such that,
    for some $h\in[d]$, the first $h$ base-arms in the orderings specified by
    $\sigma(\bw)$ and $\sigma(\bw')$ form the same set, although their orders
    within this set may differ. Denote this common set by $\armset_h$.
    Then, for any matroid action set $\nec$ corresponding to the matroid $\sM=(\armset,\sI)$,
    we have
    \begin{equation*}
        \left\lvert \supp{\sele(\bw;\nec)}\cap \armset_h\right\rvert
        =
        \left\lvert \supp{\sele(\bw';\nec)}\cap \armset_h\right\rvert,
    \end{equation*}
    that is, the number of base-arms selected by the greedy method from the first $h$ base-arms is the same under both weight vectors.
\end{lemma}

\begin{proof}
    Suppose for contradiction that the two cardinalities are different. Without loss of generality, assume that
    \begin{equation*}
        \left\lvert \supp{\sele(\bw;\nec)}\cap \armset_h\right\rvert
        >
        \left\lvert \supp{\sele(\bw';\nec)}\cap \armset_h\right\rvert.
    \end{equation*}
    By the matroid augmentation property, there exists a base-arm $j\in (\supp{\sele(\bw;\nec)}\cap\armset_h)\setminus (\supp{\sele(\bw';\nec)}\cap\armset_h)$ such that
    \begin{equation}\label{eq:greedy_num_1}
        (\supp{\sele(\bw';\nec)}\cap\armset_h)\cup\{j\}\in\supp{\sI},
    \end{equation}
    which implies that $j\in\armset_h$ and $\sele_j(\bw';\nec)=0$.

    Now consider the greedy algorithm under $\bw'$.
    Note that every base-arm processed before $j$ lies in $\armset_h$.
    Therefore, the set of base-arms selected just before $j$ is processed,
    which we denote by $\sele_{\prec j}$,
    satisfies $\sele_{\prec j}\subset \supp{\sele(\bw';\nec)}\cap \armset_h$, and thus
    \begin{equation}\label{eq:greedy_num_2}
        \sele_{\prec j}\cup\{j\}\subset (\supp{\sele(\bw';\nec)}\cap\armset_h)\cup\{j\}.
    \end{equation}
    Combining \eqref{eq:greedy_num_1} and \eqref{eq:greedy_num_2},  
    by the hereditary property we have
    $\sele_{\prec j}\cup\{j\}\in\supp{\sI}$.
    This implies that the greedy algorithm selects $j$,
    contradicting $\sele_j(\bw';\nec)=0$.
\end{proof}

\begin{lemma}\label{lem:greedy_equiv}
    For $i\in[d]$, let $\bw,\bw'\in\mathbb{R}^d$ be two weight vectors such that
    $\{j : \sigma_j(\bw)<\sigma_i(\bw)\} = \{j : \sigma_j(\bw')<\sigma_i(\bw')\}$,
    i.e., the set of base-arms that precede $i$ in descending order of weights
    is the same under $\bw$ and $\bw'$, possibly in different orders.
    Then, $\sele_i(\bw;\nec)=1$ if and only if $\sele_i(\bw';\nec)=1$.
\end{lemma}

\begin{proof}
    Let
    \begin{equation*}
        \armset_{\prec i} := \{j : \sigma_j(\bw)<\sigma_i(\bw)\}
        = \{j : \sigma_j(\bw')<\sigma_i(\bw')\},\quad
        h := \lvert \armset_{\prec i} \rvert.
    \end{equation*}
    Under both $\bw$ and $\bw'$, the set $\armset_{\prec i}$ consists of
    the first $h$ base-arms in descending order of weights, and
    $\armset_{\prec i}\cup\{i\}$ consists of the first $h+1$ base-arms.
    Then, by Lemma~\ref{lem:greedy_num} we have
    \begin{align*}
        \bigl\lvert \supp{\sele(\bw;\nec)}\cap\armset_{\prec i} \bigr\rvert
        &= \bigl\lvert \supp{\sele(\bw';\nec)}\cap\armset_{\prec i} \bigr\rvert,\\
        \bigl\lvert \supp{\sele(\bw;\nec)}\cap(\armset_{\prec i}\cup\{i\}) \bigr\rvert
        &= \bigl\lvert \supp{\sele(\bw';\nec)}\cap(\armset_{\prec i}\cup\{i\}) \bigr\rvert,
    \end{align*}
    where the first equality holds trivially when $h=0$.
    Since $i\notin\armset_{\prec i}$, we have $\sele_i(\bw;\nec)=1$ if and only if
    \begin{equation*}
        \bigl\lvert \supp{\sele(\bw;\nec)}\cap(\armset_{\prec i}\cup\{i\}) \bigr\rvert
        - \bigl\lvert \supp{\sele(\bw;\nec)}\cap\armset_{\prec i} \bigr\rvert = 1,
    \end{equation*}
    and the same holds for $\bw'$.
    By the two equalities above, this difference is the same under
    $\bw$ and $\bw'$, which completes the proof.
\end{proof}

\begin{lemma}\label{lem:greedy_remove}
    For any $i\in[d]$, let $\bw,\bw'\in\mathbb{R}^d$ satisfy
    $\{j : \sigma_j(\bw')<\sigma_i(\bw')\}\subset\{j : \sigma_j(\bw)<\sigma_i(\bw)\}$.
    That is, the set of base-arms that precede $i$ in descending order of weights under $\bw'$ is a subset of those under $\bw$.
    Then the following hold:
    \begin{enumerate}[label=(\roman*)]
        \item\label{item:greedy_mono_sel}
            if $\sele_i(\bw;\nec)=1$, then $\sele_i(\bw';\nec)=1$;
        \item\label{item:greedy_mono_notsel}
            if $\sele_i(\bw';\nec)=0$, then $\sele_i(\bw;\nec)=0$.
    \end{enumerate}
\end{lemma}

\begin{proof}
    By letting $\armset_{\prec i} := \{j : \sigma_j(\bw)<\sigma_i(\bw)\}$ and $\armset_{\prec i}' := \{j : \sigma_j(\bw')<\sigma_i(\bw')\}$,
    we have $\armset_{\prec i}'\subset\armset_{\prec i}$.

    We first prove~\ref{item:greedy_mono_sel}.
    Let ${\bf u}\in\mathbb{R}^d$ be a weight vector with distinct entries such that,
    in descending order of ${\bf u}$, the base-arms in $\armset_{\prec i}'$ come first,
    followed by those in $\armset_{\prec i}\setminus\armset_{\prec i}'$, then $i$, and then the remaining
    base-arms. 
    Since $\{j : \sigma_j({\bf u})<\sigma_i({\bf u})\} = \armset_{\prec i}$,
    Lemma~\ref{lem:greedy_equiv} gives $\sele_i({\bf u};\nec)=1$.
    As the base-arms processed before $i$ under ${\bf u}$ are exactly those in $\armset_{\prec i}$,
    this means $(\supp{\sele({\bf u};\nec)}\cap\armset_{\prec i})\cup\{i\}\in\supp{\sI}$.
    Since
    \begin{equation*}
        \supp{\sele({\bf u};\nec)}\cap\armset_{\prec i}'\subset \supp{\sele({\bf u};\nec)}\cap\armset_{\prec i},
    \end{equation*}
    the hereditary property yields
    \begin{equation}\label{eq:greedy_remove_1}
        (\supp{\sele({\bf u};\nec)}\cap\armset_{\prec i}')\cup\{i\}\in\supp{\sI}.
    \end{equation}

    Now let ${\bf u}'$ be obtained from ${\bf u}$ by moving $i$ to the position
    immediately after the base-arms in $\armset_{\prec i}'$, leaving the relative order of
    all other base-arms unchanged. Under ${\bf u}'$, the greedy algorithm processes
    the base-arms in $\armset_{\prec i}'$ in the same order as under ${\bf u}$, so the set of selected base-arms
    just before $i$ is processed is $\supp{\sele({\bf u};\nec)}\cap\armset_{\prec i}'$. 
    Combining this fact with \eqref{eq:greedy_remove_1},
    we obtain $\sele_i({\bf u}';\nec)=1$.

    Finally, since
    \begin{equation*}
        \armset_{\prec i}'=\{j : \sigma_j({\bf u}')<\sigma_i({\bf u}')\} = \{j : \sigma_j(\bw')<\sigma_i(\bw')\},
    \end{equation*}
    Lemma~\ref{lem:greedy_equiv} gives $\sele_i(\bw';\nec)=1$.

    The statement~\ref{item:greedy_mono_notsel} is the contrapositive of \ref{item:greedy_mono_sel} and thus follows immediately.
\end{proof}

\section{Analysis of the Penalty Term}\label{sec:penalty_term}
The penalty term is bounded in the following lemma.

\textbf{Lemma~\ref{lem:penalty_adv_bound} (Explicit version)}
\textit{For any $\hat{L}_t\in\mathbb{R}^d$, $\alpha>1$ and $\dis\in\qty{\Fdis,\Pdis}$, we have
    \begin{equation}\label{eq:penalty_adv_bound}
        \E_{r_t\sim\dis}\qty[\left\langle r_t, a_t - a^*\right\rangle\m\hat{L}_t]
        \leq
        \begin{cases}
            \qty(\frac{\alpha}{\alpha-1}(m-1)^{1-\frac{1}{\alpha}}+\Gamma\qty(1-\frac{1}{\alpha}))\qty(d+1)^{\frac{1}{\alpha}}+m, & \text{if } \dis = \Fdis,\\
            \qty(\frac{\alpha}{\alpha-1}(m-1)^{1-\frac{1}{\alpha}}+\Gamma\qty(1-\frac{1}{\alpha}))\qty(d+1)^{\frac{1}{\alpha}}, & \text{if } \dis = \Pdis.
        \end{cases}
    \end{equation}
    In addition, if $\underline{\hat{L}}_{t,i}>0$ for any $i$ with $a^*_i=0$, we also have
    \begin{equation*}
        \E_{r_t\sim\dis}\qty[\left\langle r_t, a_t - a^*\right\rangle\m\hat{L}_t]
        \leq
        \frac{\alpha}{\alpha-1}\sum_{i:a^*_i=0}\frac{1}{(\eta_t\underline{\hat{L}}_{t,i})^{\alpha-1}}.
    \end{equation*} 
}

\begin{proof}
    The bound~\eqref{eq:penalty_adv_bound} follows from Lemma~12 of \citet{chen2026further}, whose proof applies to any action set with exactly $m$ base-arms in each action.
    
    Next we show the latter part of the lemma, which considers the case $\underline{\hat{L}}_{t,i}>0$ for any $i$ with $a^*_i=0$.
    By~\eqref{eq:r_start}, $a_{t,i}=1$ implies $r_{t,i}\geq\nu+\eta_t\underline{\hat{L}}_{t,i}$.
    When $\dis\in\qty{\Fdis,\Pdis}$, since $f(x)\leq \alpha/x^{\alpha+1}$ for $x\geq \nu$, we have
    \begin{align*}
        \E_{r_t\sim\dis}\qty[\left\langle r_t, a_t - a^*\right\rangle\m\hat{L}_t]
        &\leq
        \sum_{i:a^*_i=0}\E_{r_t\sim\dis}\qty[a_{t,i} r_{t,i}\m\hat{L}_t]\\
        &=
        \sum_{i:a^*_i=0}\E_{r_t\sim\dis}\qty[\ind\qty[a_{t,i}=1] r_{t,i}\m\hat{L}_t]\\
        &\leq
        \sum_{i:a^*_i=0}\E_{r_t\sim\dis}\qty[\ind\qty[r_{t,i}\geq\nu+\eta_t\underline{\hat{L}}_{t,i}] r_{t,i}\m\hat{L}_t]\\
        &=
        \sum_{i:a^*_i=0} \int_{\nu}^\infty (z+\eta_t\underline{\hat{L}}_{t,i})f\qty(z+\eta_t\underline{\hat{L}}_{t,i})\dd z\\
        &\leq
        \sum_{i:a^*_i=0} \int_\nu^\infty \frac{\alpha}{(z+\eta_t\underline{\hat{L}}_{t,i})^\alpha}\dd z 
        \tag{by $f(x)\leq \alpha/x^{\alpha+1}$ for $x\geq \nu$}\\
        &\leq
        \frac{\alpha}{\alpha-1}\sum_{i:a^*_i=0}\frac{1}{(\eta_t\underline{\hat{L}}_{t,i})^{\alpha-1}},
    \end{align*}
    which concludes the proof.
\end{proof}

\section{Analysis of the Stability Term}

\subsection{Supporting Properties of Analysis}

A key supporting property in our analysis is the decomposition in the following lemma, which extends
Lemma~6 of \citet{chen2026further} for the $m$-set semi-bandit setting to the matroid semi-bandit setting. 
Although the result is similar, the proof of this extension is
not straightforward. 
In the $m$-set semi-bandit setting, whether $i$ is selected depends
only on the number of base-arms preceding it.
Under a general matroid constraint, however, the identities of the preceding base-arms also matter. 
Lemma~\ref{lem:greedy_equiv} shows that the selection of $i$ nevertheless depends only on the set of base-arms preceding it, and not on their internal ordering. This weaker invariance turns out to be sufficient to derive an analogous decomposition.

To state the result, we define
\begin{align*}
    \phi_{i\prec j}(&\lambda;\nec)
    \coloneqq
    \sP_{r\sim\dis}\qty[\sele_i(r-\lambda;\nec)=1,\, \lim_{\lambda_j\to -\infty} \sele_i(r-\lambda;\nec)=0]\\
    &=
    \sP_{r\sim\dis}\qty[\sele_i(r-\lambda;\nec)=1,\, \sele_i(r-\lambda;\nec_j)=0]
    \numberthis\label{eq:phi_ij_explain1}\\
    &=
    \sP_{r\sim\dis}\qty[\sele_i(r-\lambda;\nec)=1,\,\sigma_i(r-\lambda)<\sigma_j(r-\lambda)]
    \numberthis\label{eq:phi_ij_explain2}\\
    &=
    \int_\nu^\infty
        f(z+\ul_i)\sP_{\{r_k\}_{k\in[d]\setminus\{i,j\}} \sim \dis,\,r_i=z+\ul_i}
        \qty[\sele_i\qty(r-\ul\pn{i};\nec)=1,\, \sele_i\qty(r-\ul\pn{i};\nec_j)=0]
        F(z+\ul\pn{i}_j)
        \dd z.
        \numberthis\label{eq:def_phi_ij}
\end{align*}
Here, \eqref{eq:phi_ij_explain1} holds because, as $\lambda_j\to-\infty$, base-arm $j$ is eventually ranked first, after which the greedy algorithm on $\nec$ proceeds identically to that on $\nec_j$. Moreover, \eqref{eq:phi_ij_explain2} follows from Lemma~\ref{lem:greedy_equiv}. More specifically, if $j$ precedes $i$, moving $j$ to the front does not change whether $i$ is selected. Therefore, on the event that $i$ is selected under $\nec$ but not under $\nec_j$, $j$ must be ranked after $i$, i.e., $r_j-\lambda_j\leq r_i-\lambda_i$.
Similarly, we define
\begin{align*}
    I_{i\prec j}&(\lambda;\nec)
    \coloneqq
    \sP_{r\sim\dis}\qty[\sele_i(r-\lambda;\nec)=1,\, \lim_{\lambda_j\to -\infty} \sele_i(r-\lambda;\nec)=0,\, r_i\geq\nu+\lambda_i]\\
    &=
    \int_\nu^\infty
        f(z+\lambda_i)\sP_{\{r_k\}_{k\in[d]\setminus\{i,j\}} \sim \dis,\,r_i=z+\lambda_i}
        \qty[\sele_i\qty(r-\lambda;\nec)=1,\, \sele_i\qty(r-\lambda;\nec_j)=0]
        F(z+\lambda_j)
        \dd z,
        \numberthis\label{eq:def_I_ij}\\
    J_{i\prec j}&(\lambda;\nec)
    \coloneqq
    \E_{r\sim\dis}\qty[\frac{1}{r_i}\ind\qty[\sele_i(r-\lambda;\nec)=1,\, \lim_{\lambda_j\to -\infty} \sele_i(r-\lambda;\nec)=0,\, r_i\geq\nu+\lambda_i]],\\
    &=
    \int_\nu^\infty
        \frac{f(z+\lambda_i)}{z+\lambda_i}
        \sP_{\{r_k\}_{k\in[d]\setminus\{i,j\}} \sim \dis,\,r_i=z+\lambda_i}
        \qty[\sele_i\qty(r-\lambda;\nec)=1,\, \sele_i\qty(r-\lambda;\nec_j)=0]
        F(z+\lambda_j)
        \dd z.
        \numberthis\label{eq:def_J_ij}\\
\end{align*}

Now we present the decomposition result in the following lemma.

\begin{lemma}\label{lem:tool_relation}
    Let $\nec$ be a matroid action set corresponding to a matroid $\sM=(\armset,\sI)$.
    Then, for any $i,j\in\armset$ with $\{i,j\}\in\supp{\sI}$, we have
    \begin{align*}
        \phi_i(\lambda;\nec)&=
        \phi_i(\lambda;\nec_j)+\phi_{i\prec j}(\lambda;\nec),\\
        I_i(\lambda;\nec)&=
        I_i(\lambda;\nec_j)+I_{i\prec j}(\lambda;\nec),\\
        J_i(\lambda;\nec)&=
        J_i(\lambda;\nec_j)+J_{i\prec j}(\lambda;\nec).
    \end{align*}
\end{lemma}

\begin{proof}
    Given $\lambda\in\mathbb{R}^d$, for any $j\in[d]\setminus\{i\}$ we have the decomposition as
    \begin{multline*}
        \qty{r_t:\sele_i(r_t-\lambda;\nec)=1}
        =
        \qty{r_t:\sele_i(r_t-\lambda;\nec)=1,\, \lim_{\lambda_j\to -\infty} \sele_i(r_t-\lambda;\nec)=0}\\
        \mathbin{\dot\cup}
        \qty{r_t:\sele_i(r_t-\lambda;\nec)=1,\, \lim_{\lambda_j\to -\infty} \sele_i(r_t-\lambda;\nec)=1},
    \end{multline*}
    where the sets on the RHS are disjoint.
    Note that letting $\lambda_j\to -\infty$ moves the base-arm $j$ to the top of the ranking while preserving the relative order of all other arms.
    Then, Lemma~\ref{lem:greedy_remove} implies that
    \begin{equation*}
        \qty{r_t:\sele_i(r_t-\lambda;\nec)=1,\, \lim_{\lambda_j\to -\infty} \sele_i(r_t-\lambda;\nec)=1}
        =
        \qty{r_t:\lim_{\lambda_j\to -\infty} \sele_i(r_t-\lambda;\nec)=1}.
    \end{equation*}
    By the greedy method, since the base-arm $j$ at the first position must be selected, and the remaining base-arms are selected according to the action set $\nec_j$, we have
    \begin{equation*}
        \sP\qty[r_t:\lim_{\lambda_j\to -\infty} \sele_i(r_t-\lambda;\nec)=1]
        =
        \phi_i(\lambda;\nec_j).
    \end{equation*}
    On the other hand, consider $r_t$ such that $\sele_i(r_t-\lambda;\nec)=1$ and $\lim_{\lambda_j\to -\infty} \sele_i(r_t-\lambda;\nec)=0$, where the latter means that $i$ is not selected when $j$ is moved to the top of the ranking.
    By Lemma~\ref{lem:greedy_equiv}, $i$ is not selected when $j$ precedes $i$ in the ranking and the relative order of all other base-arms is preserved.
    Therefore, $i$ can only be selected when $j$ follows $i$ in the ranking, which implies that
    \begin{equation*}
        \sP\qty[r_t:\sele_i(r_t-\lambda;\nec)=1,\, \lim_{\lambda_j\to -\infty} \sele_i(r_t-\lambda;\nec)=0]
        =
        \phi_{i\prec j}(\lambda;\nec).
    \end{equation*}
    Combining the above results, we have
    \begin{align*}
        \phi_i(\lambda;\nec)
        =
        \phi_i(\lambda;\nec_j)
        +
        \phi_{i\prec j}(\lambda;\nec).
    \end{align*}
    Similarly, we have
    \begin{equation*}
        I_i(\lambda;\nec)=
        I_i(\lambda;\nec_j)+I_{i\prec j}(\lambda;\nec),
        \quad
        J_i(\lambda;\nec)=
        J_i(\lambda;\nec_j)+J_{i\prec j}(\lambda;\nec),
    \end{equation*}
    which completes the proof.
\end{proof}

Next, we present Lemmas~\ref{lem:both_increasing} and \ref{lem:bgaf}, which will be used in the proof of Lemma~\ref{lem:important_inequality}.
Both lemmas are established by \citet{chen2026further}.
Since their proofs do not rely on the structure of the action set, they apply directly to the matroid semi-bandit setting, and we omit the proofs here.

\begin{lemma}\citep[Lemma~13]{chen2026further}\label{lem:both_increasing}
    Let $\psi (x):[\nu,\infty)\to\mathbb{R}$ denote a non-negative function, independent of $\lambda_j$, and satisfying $\psi(x)=0$ for all $x\in[\nu, \nu-\lambda_i]$ whenever $\lambda_i<0$. 
    If $j\neq i$ and $F(x)$ is the cumulative distribution function of Fr\'{e}chet or Pareto distributions, then
    \begin{equation*}
        \frac{\int_{\nu}^\infty \psi(z)F(z+\lambda_j)/(z+\lambda_i) \dd z}{\int_{\nu}^\infty \psi(z)F(z+\lambda_j) \dd z}
    \end{equation*}
    is monotonically increasing in $\lambda_j$.
\end{lemma}

\begin{lemma}\citep[Lemma~14]{chen2026further}\label{lem:bgaf}
    Let $a,b>0$, $f(x),g(x)>0$, where $x\in\mathbb{R}$. If both $f(x)$ and $g(x)/f(x)$ are monotonically increasing in $x$, then for any $x_1<x_2$, we have 
    \begin{equation*}
        \frac{b+g(x_1)}{a+f(x_1)}\leq\frac{b}{a}\lor\frac{b+g(x_2)}{a+f(x_2)}.
    \end{equation*}
    Provided that $\lim_{x\to\infty}\qty(b+g(x))/\qty(a+f(x))$ exists, for any $x_0\in\mathbb{R}$ we have
    \begin{equation*}
        \frac{b+g(x_0)}{a+f(x_0)}\leq\frac{b}{a}\lor\lim_{x\to\infty}\frac{b+g(x)}{a+f(x)}.
    \end{equation*}
\end{lemma}

Based on Lemmas~\ref{lem:tool_relation}--\ref{lem:bgaf}, we establish the following lemma.
This lemma describes how the ratio $J_i(\cdot)/I_i(\cdot)$
changes when a single coordinate $\lambda_j$ is increased. 
To be specific, if
$\{i,j\}\notin\supp{\sI}$, the ratio $J_i(\cdot)/I_i(\cdot)$ is nondecreasing in $\lambda_j$.
Otherwise, the ratio is bounded by the maximum of the ratio after the increase and the ratio under the modified action set in which $j$ is forced to be selected.

\begin{lemma}\label{lem:important_inequality}
Let $\nec\subset\{0,1\}^d$ be a matroid action set associated with a matroid $(\armset,\sI)$, and let $i,j\in\armset$ be two distinct base-arms.
For any $\lambda'\in\mathbb{R}^d$ satisfying $\lambda'_j\geq \lambda_j$ and $\lambda_k'=\lambda_k$ for all $k\neq j$, we have
    \begin{equation*}
        \frac{J_i(\lambda;\dis,\nec)}{I_i(\lambda;\dis,\nec)}\leq\frac{J_i(\lambda';\dis,\necone{j})}{I_i(\lambda';\dis,\necone{j})}\lor\frac{J_i(\lambda';\dis,\nec)}{I_i(\lambda';\dis,\nec)}
    \end{equation*}
if $\{i,j\}\in\supp{\sI}$ (i.e., there exists $a\in\nec$ such that $a_i=a_j=1$), and
    \begin{equation*}
        \frac{J_i(\lambda;\dis,\nec)}{I_i(\lambda;\dis,\nec)}\leq\frac{J_i(\lambda';\dis,\nec)}{I_i(\lambda';\dis,\nec)}
    \end{equation*}
if $\{i,j\}\notin\supp{\sI}$ (i.e., there does not exist $a\in\nec$ such that $a_i=a_j=1$).
\end{lemma}

\begin{proof}
    We first consider the case $\{i,j\}\in\supp{\sI}$. 
    Then, there exists an action $a\in\necone{j}$ with $a_i=1$, meaning that $i$ remains feasible when $j$ is required to be selected.
    By Lemma~\ref{lem:tool_relation}, we have
    \begin{equation}\label{eq:J_phi_decomose}
        \frac{J_i(\lambda;\dis,\nec)}{I_i(\lambda;\dis,\nec)}
        =
        \frac{J_i(\lambda;\dis,\necone{j})+J_{i\prec j}(\lambda;\dis,\nec)}{I_i(\lambda;\dis,\necone{j})+I_{i\prec j}(\lambda;\dis,\nec)}.
    \end{equation}

    Note that
    $J_{i\prec j}(\lambda;\dis,\nec)/I_{i\prec j}(\lambda;\dis,\nec)$ is monotonically increasing in $\lambda_j$ because we can apply Lemma~\ref{lem:both_increasing} with
    \begin{equation}\label{eq:psi_def}
        \psi(z)\coloneqq f(z+\lambda_i)\sP_{\{r_k\}_{k\in[d]\setminus\{i,j\}} \sim \dis,\,r_i=z+\lambda_i}
        \qty[\sele_i\qty(r-\lambda;\nec)=1,\, \sele_i\qty(r-\lambda;\nec_j)=0]
    \end{equation}
    by the representations given in \eqref{eq:def_I_ij} and \eqref{eq:def_J_ij}.
    Then, by noting that $J_i(\lambda;\dis,\necone{j})$ and $I_i(\lambda;\dis,\necone{j})$ do not depend on $\lambda_j$ and applying Lemma~\ref{lem:bgaf} to \eqref{eq:J_phi_decomose} we obtain
    \begin{align*}
        \frac{J_i(\lambda;\dis,\nec)}{I_i(\lambda;\dis,\nec)}
        &\leq \frac{J_i(\lambda;\dis,\necone{j})}{I_i(\lambda;\dis,\necone{j})}
        \lor
        \frac{J_i(\lambda';\dis,\necone{j})+J_{i\prec j}(\lambda';\dis,\nec)}{I_i(\lambda';\dis,\necone{j})+I_{i\prec j}(\lambda';\dis,\nec)}\\
        &=
        \frac{J_i(\lambda;\dis,\necone{j})}{I_i(\lambda;\dis,\necone{j})}
        \lor
        \frac{J_i(\lambda';\dis,\nec)}{I_i(\lambda';\dis,\nec)}\numberthis\label{eq:monotonic_intermediate}\\
        &=
        \frac{J_i(\lambda';\dis,\necone{j})}{I_i(\lambda';\dis,\necone{j})}
        \lor
        \frac{J_i(\lambda';\dis,\nec)}{I_i(\lambda';\dis,\nec)},
    \end{align*}
    where we again used the fact that $J_i(\lambda;\dis,\necone{j})$ and $I_i(\lambda;\dis,\necone{j})$
do not depend on $\lambda_j$ in the last equality.

Next, we consider the case $\{i,j\}\notin\supp{\sI}$.
In particular, if $j$ is placed before $i$, then once $j$ is selected, $i$ cannot be added while preserving independence. 
Combining this fact with Lemma~\ref{lem:greedy_equiv}, we see that $i$ is rejected whenever $j$ precedes $i$, and thus $i$ can be selected only if $j$ appears after $i$.
Therefore, in this case $I_i(\lambda;\dis,\nec)$ can be expressed as
    \begin{align*}
        I_i(\lambda;\dis,\nec)
        &=
        I_{i\prec j}(\lambda;\dis,\nec)\\
        &=
        \int_\nu^\infty
        f(z+\lambda_i)\sP_{\{r_k\}_{k\in[d]\setminus\{i,j\}} \sim \dis,\,r_i=z+\lambda_i}
        \qty[\sele_i\qty(r-\lambda;\necr{[d]\setminus\{j\}})=1]
        F(z+\lambda_j)
        \dd z,
    \end{align*}
and similarly
    \begin{equation*}
        J_i(\lambda;\dis,\nec)
        =
        J_{i\prec j}(\lambda;\dis,\nec)
        =
        \int_\nu^\infty
        \frac{f(z+\lambda_i)}{z+\lambda_i}\sP_{\{r_k\}_{k\in[d]\setminus\{i,j\}} \sim \dis,\,r_i=z+\lambda_i}
        \qty[\sele_i\qty(r-\lambda;\necr{[d]\setminus\{j\}})=1]
        F(z+\lambda_j)
        \dd z.
    \end{equation*}
    By applying Lemma~\ref{lem:both_increasing} with
    $\psi(z)$ defined in \eqref{eq:psi_def}, one can see that $J_i(\lambda;\dis,\nec)/I_i(\lambda;\dis,\nec)$ is monotonically increasing in $\lambda_j$.
    Therefore, we have
    \begin{equation}\label{eq:monotonic_intermediate2}
        \frac{J_i(\lambda;\dis,\nec)}{I_i(\lambda;\dis,\nec)}
        \leq
        \frac{J_i(\lambda';\dis,\nec)}{I_i(\lambda';\dis,\nec)},
    \end{equation}
    which completes the proof.
\end{proof}

The following lemma gives a limiting case of Lemma~\ref{lem:important_inequality}, obtained by letting $\lambda'_j\to\infty$.

\begin{lemma}\label{lem:infty}
Consider a matroid action set $\nec\subset\{0,1\}^d$ associated with a matroid $\sM=(\armset,\sI)$, and two distinct base-arms $i,j\in\armset$.
If $\{i,j\}\in\supp{\sI}$, i.e., there exists $a\in\nec$ such that $a_i=a_j=1$, then it holds that
    \begin{equation*}
        \frac{J_i(\lambda;\dis,\nec)}{I_i(\lambda;\dis,\nec)}\leq\frac{J_i(\lambda;\dis,\necone{j})}{I_i(\lambda;\dis,\necone{j})}\lor\frac{J_i(\lambda;\dis,\necr{[d]\setminus\{j\}})}{I_i(\lambda;\dis,\necr{[d]\setminus\{j\}})}.
    \end{equation*}
Otherwise, we have
    \begin{equation*}
        \frac{J_i(\lambda;\dis,\nec)}{I_i(\lambda;\dis,\nec)}\leq\frac{J_i(\lambda;\dis,\necr{[d]\setminus\{j\}})}{I_i(\lambda;\dis,\necr{[d]\setminus\{j\}})}.
    \end{equation*}
\end{lemma}

\begin{proof}
    We first consider the case $\{i,j\}\in\supp{\sI}$.
    By \eqref{eq:monotonic_intermediate} we have
    \begin{align}
        \frac{J_i(\lambda;\dis,\nec)}{I_i(\lambda;\dis,\nec)}
        &\leq \frac{J_i(\lambda;\dis,\necone{j})}{I_i(\lambda;\dis,\necone{j})}\lor\frac{J_i(\lambda';\dis,\nec)}{I_i(\lambda';\dis,\nec)}
        \n
    \end{align}
    for any
    $\lambda'\in\mathbb{R}^d$ such that $\lambda'_j\geq \lambda_j$ and $\lambda_k'=\lambda_k$ for all $k\neq j$.
    Then we have
    \begin{align}
        \frac{J_i(\lambda;\dis,\nec)}{I_i(\lambda;\dis,\nec)}
        &\leq \frac{J_i(\lambda;\dis,\necone{j})}{I_i(\lambda;\dis,\necone{j})}\lor
    \limsup_{\lambda_j'\to \infty}
    \frac{J_i(\lambda';\dis,\nec)}{I_i(\lambda';\dis,\nec)}
    \label{eq:ratio_limit}
    \end{align}
    by taking the limit.

    For each fixed perturbation vector $r$, base-arm $j$ is eventually processed last as $\lambda'_j\to\infty$.
    Hence, the selection of $i$ agrees with that under $\necr{[d]\setminus\{j\}}$. We take the limit inside the expectation to obtain
    \begin{align*}
        \lim_{\lambda_j'\to\infty}I_i(\lambda';\dis,\nec)
        &=
        \sP_{r\sim\dis}\qty[\sele_i(r-\lambda;\necr{[d]\setminus\{j\}})=1,r_i\geq\nu+\lambda_i]\\
        &=I_i(\lambda;\dis,\necr{[d]\setminus\{j\}}).\numberthis\label{eq:limit_phi}
    \end{align*}
    The same argument applies to $J_i$, since its integrand is bounded by $1/r_i$, which has finite expectation under both perturbation distributions. Therefore, we have
    \begin{align}
    \lim_{\lambda_j'\to\infty}
    J_i(\lambda';\dis,\nec)
    &=
    J_i(\lambda;\dis,\necr{[d]\setminus\{j\}}).\label{eq:limit_J}
    \end{align}
    We obtain the first statement of the lemma 
    by combining \eqref{eq:ratio_limit} with \eqref{eq:limit_phi} and \eqref{eq:limit_J}.

    Next, we consider the case $\{i,j\}\notin\supp{\sI}$.
    By \eqref{eq:monotonic_intermediate2} and taking the limit $\lambda_j'\to\infty$, we have
    \begin{equation}\label{eq:monotonic_intermediate2_limit}
        \frac{J_i(\lambda;\dis,\nec)}{I_i(\lambda;\dis,\nec)}
        \leq
        \limsup_{\lambda_j'\to \infty}
        \frac{J_i(\lambda';\dis,\nec)}{I_i(\lambda';\dis,\nec)}.
    \end{equation}
    Combining~\eqref{eq:monotonic_intermediate2_limit} with the same limits~\eqref{eq:limit_phi} and~\eqref{eq:limit_J} gives the second statement.
\end{proof}

\subsection{Proof of Lemma~\ref{lem:lambda_star}}
\label{subsec:proof_lemma_lambda_star}

\textbf{Lemma~\ref{lem:lambda_star} (Restated)}
\textit{For any $i\in[d]$ and $\lambda\in\mathbb{R}^d$, it holds that
    \begin{equation*}
        \frac{J_i(\ul\pn{i};\dis)}{I_i(\ul\pn{i};\dis)}
        \leq
        \max_{\bm{1}_{\armset^+}\in\sI_i}
        \qty{\frac{J_i(\lambda^*;\dis,\neccr{\armset^+}{[i]\setminus\armset^+})}{I_i(\lambda^*;\dis,\neccr{\armset^+}{[i]\setminus\armset^+})}},
    \end{equation*}
    where we denote
    \begin{equation*}
        \lambda_k^* = 
        \begin{cases}
            \ul_i, & \text{if } k\leq i,\\
            \ul_k\pn{i}, & \text{if } k> i.
        \end{cases}
    \end{equation*}
}    

\begin{proof}
    Fix $i\in[d]$ throughout this proof. We define a sequence of $d$-dimensional vectors $\tilde{\lambda}\pn{i,0},\tilde{\lambda}\pn{i,1},\tilde{\lambda}\pn{i,2},\dots,\tilde{\lambda}\pn{i,i-1}$ as follows.
    Define $\tilde{\lambda}\pn{i,0}=\ul\pn{i}$. For $j=1,\dots,i-1$, we define
    \begin{equation*}
        \tilde{\lambda}\pn{i,j}_k = 
        \begin{cases}
            \ul_i, & \text{if } k\in[j]\cup\qty{i},\\
            \ul\pn{i}_k, & \text{otherwise.}
        \end{cases}
    \end{equation*}
    Consequently, we have $\tilde{\lambda}\pn{i,i-1}=\lambda^*.$

First we show
for $k\in \{0,1,\dots,i-1\}$ that % Jongyeong: [0, 1, ...] -> \{ 0, 1, \}.
\begin{align}
\frac{J_i(\ul\pn{i};\dis)}{I_i(\ul\pn{i};\dis)}
&
\leq \max_{\bm{1}_{\armset^+}\in\sI_i:\armset^+\subset[k]}\left\{
\frac{J_i(\tilde{\lambda}\pn{i,k};\dis,\nec(\armset^+))}{I_i(\tilde{\lambda}\pn{i,k};\dis,\nec(\armset^+))}
\right\}
\label{induction0}
\end{align}
by induction.
The statement \eqref{induction0} becomes trivial when $k=0$.
Assume that the statement holds for
$k= k_0<i-1$. 
Then we obtain from Lemma~\ref{lem:important_inequality} that
\begin{align}
\frac{J_i(\ul\pn{i};\dis)}{I_i(\ul\pn{i};\dis)}
&\le
 \max_{\bm{1}_{\armset^+}\in\sI_i:\armset^+\subset[k_0]}\left\{
\frac{J_i(\tilde{\lambda}\pn{i,k_0};\dis,\nec(\armset^+))}{I_i(\tilde{\lambda}\pn{i,k_0};\dis,\nec(\armset^+))}
\right\}
\nn
&\le
 \max_{\bm{1}_{\armset^+}\in\sI_i:\armset^+\subset[k_0],\,\{i,k_0+1\}\in\supp{\sI(\armset^+)}}\left\{
\frac{J_i(\tilde{\lambda}\pn{i,k_0+1};\dis,\nec(\armset^+\cup\{k_0+1\}))}{I_i(\tilde{\lambda}\pn{i,k_0+1};\dis,\nec(\armset^+\cup\{k_0+1\}))}
\right\}
\nn
&\qquad\lor
 \max_{\bm{1}_{\armset^+}\in\sI_i:\armset^+\subset[k_0]}\left\{
\frac{J_i(\tilde{\lambda}\pn{i,k_0+1};\dis,\nec(\armset^+))}{I_i(\tilde{\lambda}\pn{i,k_0+1};\dis,\nec(\armset^+))}
\right\}.
\label{reduction1}
\end{align}
Note that $\{i,k_0+1\}\in\supp{\sI(\armset^+)}$ means that $\{i,k_0+1\}\cup\armset^+\in\supp{\sI}$, which is equivalent to $\{k_0+1\}\cup{\armset^+}\in\supp{\sI_i}$.
Then, we can rewrite the first term in \eqref{reduction1} as
\begin{multline}\label{eq:constraint_reduction}
 \max_{\bm{1}_{\armset^+}\in\sI_i:\armset^+\subset[k_0],\,\{i,k_0+1\}\in\supp{\sI(\armset^+)}}\left\{
\frac{J_i(\tilde{\lambda}\pn{i,k_0+1};\dis,\nec(\armset^+\cup\{k_0+1\}))}{I_i(\tilde{\lambda}\pn{i,k_0+1};\dis,\nec(\armset^+\cup\{k_0+1\}))}
\right\}
\\=
 \max_{\bm{1}_{\armset^+}\in\sI_i:\armset^+\subset[k_0+1],\,k_0+1\in\armset^+}\left\{
\frac{J_i(\tilde{\lambda}\pn{i,k_0+1};\dis,\nec(\armset^+))}{I_i(\tilde{\lambda}\pn{i,k_0+1};\dis,\nec(\armset^+))}
\right\}.
\end{multline}
Then, since $\{\armset\subset[k_0]:\armset\in\supp{\sI_i}\}=\{\armset\subset[k_0+1]:\armset\in\supp{\sI_i},\,k_0+1\notin\armset\}$, by \eqref{reduction1} and \eqref{eq:constraint_reduction}
we have
\begin{align}
\frac{J_i(\ul\pn{i};\dis)}{I_i(\ul\pn{i};\dis)}
&\le
 \max_{\bm{1}_{\armset^+}\in\sI_i:\armset^+\subset[k_0+1],\,k_0+1\in\armset^+}\left\{
\frac{J_i(\tilde{\lambda}\pn{i,k_0+1};\dis,\nec(\armset^+))}{I_i(\tilde{\lambda}\pn{i,k_0+1};\dis,\nec(\armset^+))}
\right\}
\nn
&\qquad\lor
\max_{\bm{1}_{\armset^+}\in\sI_i:\armset^+\subset[k_0+1],\,k_0+1\notin\armset^+}\left\{
\frac{J_i(\tilde{\lambda}\pn{i,k_0+1};\dis,\nec(\armset^+))}{I_i(\tilde{\lambda}\pn{i,k_0+1};\dis,\nec(\armset^+))}
\right\}
\nn
&=
\max_{\bm{1}_{\armset^+}\in\sI_i:\armset^+\subset[k_0+1]}
\left\{
\frac{J_i(\tilde{\lambda}\pn{i,k_0+1};\dis,\nec(\armset^+))}{I_i(\tilde{\lambda}\pn{i,k_0+1};\dis,\nec(\armset^+))}
\right\},
\n
\end{align}
which completes the induction step.

Now, by letting $k=i-1$ in \eqref{induction0} we have
\begin{align}
\frac{J_i(\ul\pn{i};\dis)}{I_i(\ul\pn{i};\dis)}
&
\leq \max_{\bm{1}_{\armset^+}\in\sI_i:\armset^+\subset[i-1]}\left\{
\frac{J_i(\lambda^*;\dis,\nec(\armset^+))}{I_i(\lambda^*;\dis,\nec(\armset^+))}
\right\}. 
\label{induction3}
\end{align}
Next, we show that for any $k\in\{0,1,2,\dots,d-i\}$ we have
\begin{equation}\label{eq:induction_stage2}
    \frac{J_i(\ul\pn{i};\dis)}{I_i(\ul\pn{i};\dis)}
        \leq
        \max_{\bm{1}_{\armset^+}\in\sI_i:\armset^+\subset([i-1]\cup[d-k+1:d])}
        \qty{\frac{J_i(\lambda^*;\dis,\neccr{\armset^+}{[d-k]\setminus\armset^+})}{I_i(\lambda^*;\dis,\neccr{\armset^+}{[d-k]\setminus\armset^+})}}
\end{equation}
by induction.
Since the ground set of the matroid is $[d]$, when $k=0$, we have $\necc{\armset^+}=\neccr{\armset^+}{([d]\setminus\armset^+)}$, so \eqref{eq:induction_stage2} follows directly from \eqref{induction3}.
Assume that the statement holds for $k= k_0<d-i$. 
Then, by applying Lemma~\ref{lem:infty} we obtain
\begin{align*}
\lefteqn{\frac{J_i(\ul\pn{i};\dis)}{I_i(\ul\pn{i};\dis)}}\\
&\le
 \max_{\bm{1}_{\armset^+}\in\sI_i:\armset^+\subset([i-1]\cup[d-k_0+1:d])}
        \qty{\frac{J_i(\lambda^*;\dis,\neccr{\armset^+}{[d-k_0]\setminus\armset^+})}{I_i(\lambda^*;\dis,\neccr{\armset^+}{[d-k_0]\setminus\armset^+})}}
\nn
&\le
 \max_{\bm{1}_{\armset^+}\in\sI_i:\armset^+\subset([i-1]\cup[d-k_0+1:d]),\,\{i,d-k_0\}\in\supp{\basecr{\armset^+}{[d-k_0]\setminus\armset^+}}}\left\{
\frac{J_i(\lambda^*;\dis,\neccr{\armset^+\cup\{d-k_0\}}{[d-(k_0+1)]\setminus\armset^+})}{I_i(\lambda^*;\dis,\neccr{\armset^+\cup\{d-k_0\}}{[d-(k_0+1)]\setminus\armset^+})}
\right\}
\nn
&\qquad\lor
 \max_{\bm{1}_{\armset^+}\in\sI_i:\armset^+\subset([i-1]\cup[d-k_0+1:d])}\left\{
\frac{J_i(\lambda^*;\dis,\neccr{\armset^+}{[d-(k_0+1)]\setminus\armset^+})}{I_i(\lambda^*;\dis,\neccr{\armset^+}{[d-(k_0+1)]\setminus\armset^+})}
\right\}
\nn
&\le
 \max_{\bm{1}_{\armset^+}\in\sI_i:\armset^+\subset([i-1]\cup[d-k_0:d]),\,d-k_0\in\armset^+}\left\{
\frac{J_i(\lambda^*;\dis,\neccr{\armset^+}{[d-(k_0+1)]\setminus\armset^+})}{I_i(\lambda^*;\dis,\neccr{\armset^+}{[d-(k_0+1)]\setminus\armset^+})}
\right\}
\nn
&\qquad\lor
 \max_{\bm{1}_{\armset^+}\in\sI_i:\armset^+\subset([i-1]\cup[d-k_0:d]),\,d-k_0\notin\armset^+}\left\{
\frac{J_i(\lambda^*;\dis,\neccr{\armset^+}{[d-(k_0+1)]\setminus\armset^+})}{I_i(\lambda^*;\dis,\neccr{\armset^+}{[d-(k_0+1)]\setminus\armset^+})}
\right\}
\nn
&=
\max_{\bm{1}_{\armset^+}\in\sI_i:\armset^+\subset([i-1]\cup[d-k_0:d])}\left\{
\frac{J_i(\lambda^*;\dis,\neccr{\armset^+}{[d-(k_0+1)]\setminus\armset^+})}{I_i(\lambda^*;\dis,\neccr{\armset^+}{[d-(k_0+1)]\setminus\armset^+})}
\right\},
\end{align*}
which completes the induction step.

Finally, setting $k=d-i$ in \eqref{eq:induction_stage2}, the constraint
on sets $\armset^+$ satisfying $\bm{1}_{\armset^+}\in\sI_i$ becomes $\armset^+\subset[i-1]\cup[i+1:d]=[d]\setminus\{i\}$, which holds for every
set $\armset^+$ with $\bm{1}_{\armset^+}\in\sI_i$ and can thus be dropped. This gives
\begin{equation*}
    \frac{J_i(\ul\pn{i};\dis)}{I_i(\ul\pn{i};\dis)}
    \leq
    \max_{\bm{1}_{\armset^+}\in\sI_i}\left\{
\frac{J_i(\lambda^*;\dis,\neccr{\armset^+}{[i]\setminus\armset^+})}{I_i(\lambda^*;\dis,\neccr{\armset^+}{[i]\setminus\armset^+})}
\right\},
\end{equation*}
which completes the proof.
\end{proof}

\subsection{Properties of an Auxiliary Function}\label{subsec:auxiliary_function}

To bound the ratio $J_i(\lams)/I_i(\lams)$, we introduce an auxiliary function as follows.

Let $\nec$ be a matroid action set associated with a matroid $\sM=(\armset,\sI)$, and fix a base-arm $i\in\armset$. 
For $u\in[0,1]$, define
\begin{equation*}
    \easypstar{i}{u;\nec} \coloneqq
    \sP_{\bw: \ind[\bw_k>\bw_i]\,\overset{\text{i.i.d.}}{\sim}\,\mathrm{Ber}(u), k\in\armset\setminus\{i\}}
    \qty[\sele_i(\bw;\nec)=1].
\end{equation*}
Intuitively, $\easypstar{i}{u;\nec}$ is the probability that base-arm $i$ is selected when each other base-arm is independently placed before $i$ (i.e., $\bw_k > \bw_i$) with probability $u$. By Lemma~\ref{lem:greedy_equiv}, whether $i$ is selected depends only on the set of base-arms preceding $i$, and not on their internal ordering. 
Therefore, specifying only the relative positions of the other base-arms with respect to $i$ is sufficient to make $\easypstar{i}{u;\nec}$ well defined.

Next, we define $\easyp{i}{u;\nec,\dis}$ separately for Fr\'{e}chet and Pareto perturbations by
\begin{equation*}
    \easyp{i}{u;\nec,\dis} \coloneqq
    \begin{cases}
        e^{-u}\easypstar{i}{1-e^{-u};\nec}, & \dis=\Fdis,\\
        \easypstar{i}{u;\nec}, & \dis=\Pdis,
    \end{cases}
\end{equation*}
where $u\in[0,\infty)$ for $\dis=\Fdis$ and $u\in[0,1]$ for $\dis=\Pdis$.
We show the useful properties of $\easyp{i}{\cdot}$ in the following lemma, which will be used in the proof of Lemma~\ref{lem:lambda_star}.

\begin{lemma}\label{lem:easyp}
    Let $\nec$ be a matroid action set associated with a matroid $\sM=(\armset,\sI)$, and fix a base-arm $i\in\armset$ and a distribution $\dis\in\{\Fdis,\Pdis\}$.
    Then, $\easyp{i}{u;\nec,\dis}$ is nonincreasing in $u$ on its domain. 

    Furthermore, for any $u,v\in[0,\infty)$ when $\dis=\Fdis$, or $u,v\in[0,1]$ with $u+v\leq1$ when $\dis=\Pdis$,
    \begin{equation*}
        \easyp{i}{u+v;\nec,\dis}
        \leq \easyp{i}{u;\nec,\dis}\,\easyp{i}{v;\nec,\dis}.
    \end{equation*}
\end{lemma}

\begin{proof}
    We first show that $\easypstar{i}{u;\nec}$ is nonincreasing in $u$.

    Fix $0\leq u\leq u'\leq 1$. Let $(U_k)_{k\in\armset\setminus\{i\}}$ be
    i.i.d.\ uniform random variables on $[0,1]$, and for $s\in\{u,u'\}$, define the random subset $S_s\coloneqq\{k : U_k\leq s\}$. Since the events $\{U_k\leq s\}$ are independent across $k$ and each occurs with probability $s$, we have $\easypstar{i}{s;\nec}=\sP[S_s\rightsquigarrow_{\nec} i]$.
    Moreover, $S_u\subset S_{u'}$ holds for every realization of $(U_k)_k$, so by Lemma~\ref{lem:greedy_remove}, $S_{u'}\rightsquigarrow_{\nec} i$ implies $S_u\rightsquigarrow_{\nec} i$.
    Therefore, we have
    \begin{equation*}
        \easypstar{i}{u';\nec}
        = \sP[S_{u'}\rightsquigarrow_{\nec} i]
        \leq \sP[S_u\rightsquigarrow_{\nec} i]
        = \easypstar{i}{u;\nec}.
    \end{equation*}
    This immediately gives the monotonicity of $\easyp{i}{u;\nec,\Pdis}$.
    For Fr\'{e}chet perturbations, $1-e^{-u}$ is increasing in $u$, so both $e^{-u}$ and $\easypstar{i}{1-e^{-u};\nec}$ are nonnegative and nonincreasing. 
    Therefore, their product $\easyp{i}{u;\nec,\Fdis}$ is also nonincreasing.

    Next, we establish a bound for $\easypstar{i}{\cdot\,;\nec}$ valid for all $a,b\in[0,1]$.
    Let $\armset_a$ and $\armset_b$ be independent random subsets of
    $\armset\setminus\{i\}$ that contain each base-arm independently
    with probability $a$ and $b$, respectively. Then, for any
    $k\in\armset\setminus\{i\}$,
    \begin{align*}
        \sP[k\in\armset_a\cup\armset_b]
        &= \sP[k\in\armset_a]+\sP[k\in\armset_b]-\sP[k\in\armset_a]\,\sP[k\in\armset_b]\\
        &= a+b-ab,
    \end{align*}
    and these events are independent across $k$.
    By Lemma~\ref{lem:greedy_remove}, for every realization of $\armset_a$ and
    $\armset_b$, $(\armset_a\cup\armset_b)\rightsquigarrow_{\nec} i$ implies both $\armset_a\rightsquigarrow_{\nec} i$ and
    $\armset_b\rightsquigarrow_{\nec} i$.
    Therefore, by the independence of $\armset_a$ and $\armset_b$, we have
    \begin{align*}
        \easypstar{i}{a+b-ab;\nec}
        &= \sP\qty[(\armset_a\cup\armset_b)\rightsquigarrow_{\nec} i]\\
        &\leq \sP\qty[\armset_a\rightsquigarrow_{\nec} i,\armset_b\rightsquigarrow_{\nec} i]\\
        &= \sP\qty[\armset_a\rightsquigarrow_{\nec} i]\,\sP\qty[\armset_b\rightsquigarrow_{\nec} i]\\
        &= \easypstar{i}{a;\nec}\,\easypstar{i}{b;\nec}.\numberthis\label{eq:submultiplicativity_star}
    \end{align*}
    For Pareto perturbations, for any $u,v\in[0,1]$ with $u+v\leq1$, the monotonicity of $\easypstar{i}{\cdot\,;\nec}$ and \eqref{eq:submultiplicativity_star} give
    \begin{align*}
        \easyp{i}{u+v;\nec,\Pdis}
        &= \easypstar{i}{u+v;\nec}\\
        &\leq \easypstar{i}{u+v-uv;\nec}\\
        &\leq \easypstar{i}{u;\nec}\,\easypstar{i}{v;\nec}\\
        &= \easyp{i}{u;\nec,\Pdis}\,\easyp{i}{v;\nec,\Pdis}.
    \end{align*}
    For Fr\'{e}chet perturbations, note that
    \begin{equation*}
        1-e^{-(u+v)}=(1-e^{-u})+(1-e^{-v})-(1-e^{-u})(1-e^{-v}).
    \end{equation*}
    Then, for any $u,v\in[0,\infty)$, we have
    \begin{align*}
        \easyp{i}{u+v;\nec,\Fdis}
        &= e^{-(u+v)}\easypstar{i}{1-e^{-(u+v)};\nec}\\
        &= e^{-(u+v)}\easypstar{i}{(1-e^{-u})+(1-e^{-v})-(1-e^{-u})(1-e^{-v});\nec}\\
        &\leq e^{-u}\easypstar{i}{1-e^{-u};\nec}\,e^{-v}\easypstar{i}{1-e^{-v};\nec}
        \tag{by \eqref{eq:submultiplicativity_star}}\\
        &= \easyp{i}{u;\nec,\Fdis}\,\easyp{i}{v;\nec,\Fdis},
    \end{align*}
    which completes the proof.
\end{proof}

\subsection{Proof of Lemma~\ref{lem:sigma_i}}\label{subsec:proof_sigma_i}

When $\ul_i \leq 0$, the bound by $1/0=\infty$ trivially holds. % Jongyeong :add
Therefore, it suffices to consider the case $\ul_i > 0$. In this case, % Jongyeong :add
the integrand below vanishes for $z<\nu$ by~\eqref{eq:r_start}, and the first bound follows from
\begin{align*}
J_i(\ul\pn{i};\dis,\nec)
&=
    \int_\nu^\infty \frac{1}{z+\ul_i}
    \sP_{\{r_k\}_{k\in[d]\setminus\{i\}} \sim \dis,\,r_i=z+\ul_i}
    \qty[\sele_i\qty(r-\lambda;\nec)=1]
    \dd F(z+\ul_i)
\\
&\le
\max_{w\ge \nu}\frac{1}{w+\ul_i}
    \int_\nu^\infty \sP_{\{r_k\}_{k\in[d]\setminus\{i\}} \sim \dis,\,r_i=z+\ul_i}
    \qty[\sele_i\qty(r-\lambda;\nec)=1] \dd F(z+\ul_i)
\\
&\leq
\frac{1}{\ul_i}
I_i(\ul\pn{i};\dis,\nec).
\end{align*}

On the other hand, by recalling the definition of $\neccr{\armset^+}{\armset^-}$ for disjoint sets $\armset^+,\armset^-\subset[d]$ with $\bm{1}_{\armset^+}\in\sI_i$, we have
\begin{equation*}
    \neccr{\armset^+}{([i]\setminus\armset^+)}=\qty{a\in\{0,1\}^d: 
a+\bm{1}_{\armset^+}\in\necr{[i]\cup\armset^+}},
\end{equation*}
which is a matroid action set associated with the base-arm set $[i]\setminus\armset^+$.
Note that the elements of $\lams$ are the same for the base-arms $k\in[i]\setminus\armset^+$.

In the following, for any $i\in[d]$ and any set $\armset^+$ with $\bm{1}_{\armset^+}\in\sI_i$ we simply write $\neccr{\armset^+}{[i]\setminus\armset^+}$ as $\nec_{\armset^+}$ for brevity.
By recalling the definitions of $I_i(\cdot)$ we have
\begin{align*}
    I_i(\lams;\dis,\nec_{\armset^+})
    &=
    \sP_{r\sim \dis}\qty[\sele_i(r-\lams;\nec_{\armset^+})=1,r_i\geq\nu+\ul_i]\\
    &=
    \int_{\nu+\ul_i}^\infty \sP_{r\sim \dis}\qty[\sele_i(r-\lams;\nec_{\armset^+})=1\m r_i=z] \dd F(z)\\
    &=
    \int_{\nu+\ul_i}^\infty \sP_{r\sim \dis}\qty[\sele_i(r;\nec_{\armset^+})=1\m r_i=z] \dd F(z),\numberthis\label{eq:remove_lams}
\end{align*}
Here, \eqref{eq:remove_lams} holds since $\lams$ has identical entries on
the ground set $[i]\setminus\armset^+$ of $\neccr{\armset^+}{[i]\setminus\armset^+}$.
Therefore, replacing $r$ with $r-\lams$ shifts the weights of all these base-arms by the same constant, which preserves their ordering and thus whether $i$ is selected.

Note that, conditioned on $r_i=z$, each base-arm $k\in[i-1]\setminus\armset^+$
precedes $i$ if and only if $r_k>z$, which occurs independently across $k$
with probability $1-F(z)$. 
Therefore, $\sP_{r\sim\dis}\qty[\sele_i(r;\nec_{\armset^+})=1\m r_i=z]$ is exactly the probability that $i$ is selected when each such base-arm independently precedes $i$ with probability $1-F(z)$.

By recalling the definitions of $\easyp{i}{u;\nec,\dis}$ given in Appendix~\ref{subsec:auxiliary_function}, for $\nec_{\armset^+}=\neccr{\armset^+}{[i]\setminus\armset^+}$ we have
\begin{equation*}
    \easyp{i}{u;\nec_{\armset^+},\dis} \coloneqq
    \begin{cases}
        e^{-u}\easypstar{i}{1-e^{-u};\nec_{\armset^+}}, & \dis=\Fdis,\\
        \easypstar{i}{u;\nec_{\armset^+}}, & \dis=\Pdis.
    \end{cases}
\end{equation*}
Here, $\easypstar{i}{u;\nec_{\armset^+}}$ is given by
\begin{equation*}
    \easypstar{i}{u;\nec_{\armset^+}} \coloneqq
    \sP_{\bw: \ind[\bw_k>\bw_i]\,\overset{\text{i.i.d.}}{\sim}\,\mathrm{Ber}(u),\, k\in[i-1]\setminus\armset^+}
    \qty[\sele_i(\bw;\nec_{\armset^+})=1],
\end{equation*}
which denotes the probability that base-arm $i$ is selected when each other base-arm in $[i-1]\setminus\armset^+$ is independently placed before $i$ (i.e., $\bw_k > \bw_i$) with probability $u$.
By using this definition, we can rewrite \eqref{eq:remove_lams} as
\begin{equation*}
    I_i(\lams;\dis,\nec_{\armset^+})
    =
    \int_{\nu+\ul_i}^\infty \easypstar{i}{1-F(z);\nec_{\armset^+}} f(z)\dd z.
\end{equation*}
Similarly, we have
\begin{equation*}
    J_i(\lams;\dis,\nec_{\armset^+})
    =
    \int_{\nu+\ul_i}^\infty \frac{1}{z}\easypstar{i}{1-F(z);\nec_{\armset^+}} f(z)\dd z.
\end{equation*}
Then, we have
\begin{align*}
    \frac{J_i(\lams;\dis,\nec_{\armset^+})}{I_i(\lams;\dis,\nec_{\armset^+})}
    &=
    \frac{\int_{\nu+\ul_i}^\infty \frac{1}{z}\easypstar{i}{1-F(z);\nec_{\armset^+}} f(z)\dd z}
    {\int_{\nu+\ul_i}^\infty \easypstar{i}{1-F(z);\nec_{\armset^+}} f(z)\dd z}\\
    &\leq
    \frac{\int_\nu^\infty \frac{1}{z}\easypstar{i}{1-F(z);\nec_{\armset^+}} f(z)\dd z}
    {\int_\nu^\infty \easypstar{i}{1-F(z);\nec_{\armset^+}} f(z)\dd z}
    \numberthis\label{eq:jphi1}\\
    &=
    \begin{cases}
        \displaystyle
        \frac{\int_0^{b_{\dis}} u^{1/\alpha}e^{-u}\easypstar{i}{1-e^{-u};\nec_{\armset^+}}\dd u}
        {\int_0^{b_{\dis}} e^{-u}\easypstar{i}{1-e^{-u};\nec_{\armset^+}}\dd u},
        & \dis=\Fdis,\\[3ex]
        \displaystyle
        \frac{\int_0^{b_{\dis}} u^{1/\alpha}\easypstar{i}{u;\nec_{\armset^+}}\dd u}
        {\int_0^{b_{\dis}} \easypstar{i}{u;\nec_{\armset^+}}\dd u},
        & \dis=\Pdis,
    \end{cases}
    \numberthis\label{eq:to_holder1}\\
    &=
    \frac{\int_0^{b_{\dis}} u^{1/\alpha}\easyp{i}{u;\nec_{\armset^+},\dis} \dd u}
    {\int_0^{b_{\dis}} \easyp{i}{u;\nec_{\armset^+},\dis} \dd u},
    \numberthis\label{eq:to_holder2}
\end{align*}
where \eqref{eq:jphi1} follows by Lemma~\ref{lem:mono_B} with the decreasing function $f(x)\coloneqq 1/x$ with $g(x)\coloneqq\easypstar{i}{1-F(x);\nec_{\armset^+}}f(x)$.
Here, \eqref{eq:to_holder1} follows by the change of variable $u=z^{-\alpha}$ for both Fr\'{e}chet and Pareto perturbations, which yields $1/z=u^{1/\alpha}$ and
\begin{equation*}
    f(z)\dd z=
    \begin{cases}
        -e^{-u}\dd u, & \dis=\Fdis,\\
        -\dd u, & \dis=\Pdis,
    \end{cases}
    \quad\mbox{and}\quad
    b_{\dis} =
    \lim_{z\to\nu^+} z^{-\alpha} =
    \begin{cases}
        \infty, & \dis=\Fdis,\\
        1, & \dis=\Pdis.
    \end{cases}
\end{equation*}
Then, we have
\begin{align*}
    \eqref{eq:to_holder2}
    &=
    \frac{\int_0^{b_{\dis}} \qty(u\easyp{i}{u;\nec_{\armset^+},\dis})^{1/\alpha}\easyp{i}{u;\nec_{\armset^+},\dis}^{1-1/\alpha} \dd u}
    {\int_0^{b_{\dis}} \easyp{i}{u;\nec_{\armset^+},\dis} \dd u}\\
    &\leq
    \frac{\qty(\int_0^{b_{\dis}} u\easyp{i}{u;\nec_{\armset^+},\dis}\dd u)^{1/\alpha}\qty(\int_0^{b_{\dis}} \easyp{i}{u;\nec_{\armset^+},\dis}\dd u)^{1-1/\alpha}}
    {\int_0^{b_{\dis}} \easyp{i}{u;\nec_{\armset^+},\dis} \dd u}\numberthis\label{eq:to_holder3}\\
    &=
    \qty(\frac{\int_0^{b_{\dis}} u\easyp{i}{u;\nec_{\armset^+},\dis}\dd u}{\int_0^{b_{\dis}} \easyp{i}{u;\nec_{\armset^+},\dis}\dd u})^{1/\alpha},
\end{align*}
where \eqref{eq:to_holder3} follows by applying Lemma~\ref{lem:holder}, H{\"o}lder's inequality.

Then, we further provide a bound for $\int_0^{b_{\dis}} u\easyp{i}{u;\nec_{\armset^+},\dis}\dd u$ by
\begin{align*}
    \int_0^{b_{\dis}} u\easyp{i}{u;\nec_{\armset^+},\dis}\dd u
    &=
    \int_0^{b_{\dis}}\int_0^u \easyp{i}{u;\nec_{\armset^+},\dis}\dd v\dd u\\
    &=
    \int_0^{b_{\dis}}\int_0^{b_{\dis}-s} \easyp{i}{s+v;\nec_{\armset^+},\dis}
    \dd v\dd s
    \numberthis\label{eq:change_of_vars}\\
    &\leq
    \int_0^{b_{\dis}}\int_0^{b_{\dis}-s} \easyp{i}{v;\nec_{\armset^+},\dis}\easyp{i}{s;\nec_{\armset^+},\dis}\dd v\dd s
    \tag{by Lemma~\ref{lem:easyp}}\\
    &\leq
    \int_0^{b_{\dis}}\int_0^{b_{\dis}} \easyp{i}{v;\nec_{\armset^+},\dis}\easyp{i}{s;\nec_{\armset^+},\dis}\dd v\dd s
    \tag{by $\easyp{i}{\cdot}\geq 0$}\\
    &=
    \qty(\int_0^{b_{\dis}} \easyp{i}{u;\nec_{\armset^+},\dis} \dd u)^2,
\end{align*}
where \eqref{eq:change_of_vars} holds by applying the change of variables $s=u-v$ (equivalently, $u=s+v$) with Jacobian determinant $\frac{\partial(u,v)}{\partial(s,v)}=1$, and the region  
$
\{(u,v):0\leq v\leq u\le b_{\dis}\}
$
is transformed into
$
\{(s,v):0\leq s\le b_{\dis},\ 0\leq v\le b_{\dis}-s\}.
$

By combining the results above, we immediately obtain
\begin{align*}
    \frac{J_i(\lams;\dis,\nec_{\armset^+})}{I_i(\lams;\dis,\nec_{\armset^+})}
    &\leq
    \qty(\int_0^{b_{\dis}} \easyp{i}{u;\nec_{\armset^+},\dis} \dd u)^{1/\alpha}\\
    &=
    \begin{cases}
        \displaystyle
        \int_0^{b_{\dis}} e^{-u}\easypstar{i}{1-e^{-u};\nec_{\armset^+}}\dd u,
        & \dis=\Fdis,\\[1ex]
        \displaystyle
        \int_0^{b_{\dis}} \easypstar{i}{u;\nec_{\armset^+}}\dd u,
        & \dis=\Pdis,
    \end{cases}
    \\
    &=
    \qty(\int_\nu^\infty \easypstar{i}{1-F(z);\nec_{\armset^+}}f(z) \dd z)^{1/\alpha}
    \tag{by the change of variables $u=z^{-\alpha}$}\\
    &=\qty(\phi_i(\lams;\dis,\nec_{\armset^+}))^{1/\alpha}.
    \numberthis\label{eq:JIbound_to_max}
\end{align*}

Then, for any set $\armset^+$ with $\bm{1}_{\armset^+}\in\sI_i$ we have
\begin{align*}
    \phi_i(\lams;\dis,\neccr{\armset^+}{[i]\setminus\armset^+})
    &=
    \lim_{\max_{j\in\armset^+}\lams_j\to -\infty}
    \phi_i(\lams;\dis,\necr{[i]\cup\armset^+})\\
    &\leq
    \phi_i(\lams;\dis,\necr{[i]\cup\armset^+})
    \tag{by Lemma~\ref{lem:phi_monotone}}\\
    &\leq
    \lim_{\min_{j\in\armset^+\setminus[i]}\lams_j\to +\infty}\phi_i(\lams;\dis,\necr{[i]\cup\armset^+})
    \tag{by Lemma~\ref{lem:phi_monotone}}\\
    &=
    \phi_i(\lams;\dis,\necr{[i]}).
    \numberthis\label{eq:JIbound_to_max2}
\end{align*}
Combining Lemma~\ref{lem:lambda_star} with \eqref{eq:JIbound_to_max}
and \eqref{eq:JIbound_to_max2}, we obtain
\begin{equation*}
    \frac{J_i(\ul\pn{i};\dis)}{I_i(\ul\pn{i};\dis)}\leq
    \qty(\phi_i(\lams;\dis,\necr{[i]}))^{1/\alpha},
\end{equation*}
where recall that $\lams$ is defined separately for each $i\in[d]$ by
    \begin{equation*}
        \lambda_k^* = 
        \begin{cases}
            \ul_i, & \text{if } k\leq i,\\
            \ul_k\pn{i}, & \text{if } k> i.
        \end{cases}
    \end{equation*}
We next prove
    \begin{equation*}
        \sum_{i\in[d]}\qty(\phi_i(\lams;\dis,\nec|[i]))^{\frac{1}{\alpha}}
        \leq 
        \frac{\alpha+1}{\alpha-1}
        m^{\frac{1}{\alpha}}d^{1-\frac{1}{\alpha}}.
    \end{equation*}

For any $u\in[0,1]$, let $\armset(u)$ be a random subset of $[d]$ that contains each base-arm independently with probability $u$, and fix a realization $\armset_0$ of $\armset(u)$.
If the greedy algorithm processes the base-arms in
$\armset_0$ in increasing order of their indices, then the base-arms preceding
each $i\in\armset_0$ are exactly those in $\armset_0\cap[i-1]$.
Therefore, the base-arms
$i\in\armset_0$ with $(\armset_0\cap[i-1])\rightsquigarrow_{\nec} i$ are exactly those selected by the
greedy algorithm.
Since the number of selected base-arms is at most the rank $m$, we have
\begin{equation*}
    \sum_{i\in[d]} \ind[i\in\armset_0,\,\qty(\armset_0\cap[i-1])\rightsquigarrow_{\nec} i]\leq m.
\end{equation*}
Then, by taking the expectation over $\armset_0$, we obtain
\begin{align*}
    \sum_{i\in[d]} \sP\qty[i\in\armset(u),\,\qty(\armset(u)\cap[i-1])\rightsquigarrow_{\nec} i]
    &=
    \sum_{i\in[d]} \sP\qty[i\in\armset(u)]\sP\qty[\qty(\armset(u)\cap[i-1])\rightsquigarrow_{\nec} i]\\
    &=
    u\sum_{i\in[d]}\easypstar{i}{u;\necr{[i]}}
    \leq m,
\end{align*}
which yields
\begin{equation}\label{eq:easyp_sum_bound}
    \sum_{i\in[d]}\easypstar{i}{u;\necr{[i]}}\leq \frac{m}{u}\land d.
\end{equation}
Here, the bound $d$ follows from $\easypstar{i}{u;\necr{[i]}}\leq 1$ for every $i\in[d]$.

Now fix an arbitrary arm $i\in[d]$. Since the vector $\lams$ constructed for this $i$
has identical entries on $[i]$, the substitution $u=1-F(z)$ gives, for either perturbation distribution,
\begin{align*}
    \phi_i(\lams;\dis,\necr{[i]})
    &=
    \int_\nu^\infty \easypstar{i}{1-F(z);\necr{[i]}}\dd F(z)\\
    &=
    \int_0^1 \easypstar{i}{u;\necr{[i]}}\dd u.
    \tag{by the change of variable $u=1-F(z)$}
\end{align*}
Then, for any $s\in[0,1]$ we have
\begin{align*}
    \phi_i(\lams;\dis,\necr{[i]})
    &=
    \int_0^s \easypstar{i}{u;\necr{[i]}}\dd u
    +
    \int_s^1 \easypstar{i}{u;\necr{[i]}}\dd u\\
    &\leq
    s+\int_s^1 \easypstar{i}{u;\necr{[i]}}\dd u
    \tag{by $\easypstar{i}{\cdot}\leq 1$}\\
    &\leq
    s+\easypstar{i}{s;\necr{[i]}}\int_0^{1-s} \easypstar{i}{u;\necr{[i]}}\dd u
    \tag{by Lemma~\ref{lem:easyp} for $\dis=\Pdis$}\\
    &\leq
    s+\easypstar{i}{s;\necr{[i]}}\int_0^{1} \easypstar{i}{u;\necr{[i]}}\dd u
    \tag{by $\easypstar{i}{\cdot}\geq 0$}\\
    &=
    s+\easypstar{i}{s;\necr{[i]}}
    \phi_i(\lams;\dis,\necr{[i]}).\numberthis\label{eq:phi_to_rearrange}
\end{align*}
By rearranging the terms in \eqref{eq:phi_to_rearrange}, we obtain the lower bound of $\easypstar{i}{s;\necr{[i]}}$ as
\begin{equation}\label{eq:easyp_lower_bound}
    \easypstar{i}{s;\necr{[i]}}
    \geq
    \qty(1-\frac{s}{\phi_i(\lams;\dis,\necr{[i]})})\lor 0.
\end{equation}
Next, we show that there exist constants $\beta>0$ and $C>0$ such that
\begin{equation*}
    \phi_i^{1/\alpha}
    \leq
    C\int_0^1 u^{-\beta}\easypstar{i}{u}\dd u
\end{equation*}
for all $i\in[d]$.
Here, we simply write $\phi_i(\lams;\dis,\necr{[i]})$ and $\easypstar{i}{u;\necr{[i]}}$ as $\phi_i$ and $\easypstar{i}{u}$, respectively.
Applying \eqref{eq:easyp_lower_bound} with $s$ replaced by $u$ to the integral on the right-hand side of the above inequality, we obtain
\begin{align*}
    \int_0^1 u^{-\beta}\easypstar{i}{u}\dd u
    &\geq
    \int_0^1 u^{-\beta}\qty(\qty(1-\frac{u}{\phi_i})\lor 0)\dd u\\
    &=
    \int_0^{\phi_i} \qty(u^{-\beta}-\frac{u^{1-\beta}}{\phi_i})\dd u
    \tag{since the integrand equals $0$ for $u>\phi_i$}
    \\
    &=\qty(\frac{1}{1-\beta}-\frac{1}{2-\beta})\phi_i^{1-\beta}.
\end{align*}
Therefore, by choosing $\beta=1-1/\alpha$ we obtain
\begin{equation}\label{eq:phi_easyp_bound}
    \qty(\phi_i(\lams;\dis,\necr{[i]}))^{1/\alpha}
    \leq
    \frac{\alpha+1}{\alpha^2}\int_0^1 u^{\frac{1}{\alpha}-1}
    \easypstar{i}{u;\necr{[i]}}\dd u.
\end{equation}
Summing \eqref{eq:phi_easyp_bound} over $i\in[d]$
and using \eqref{eq:easyp_sum_bound}, we obtain
\begin{align*}
    \sum_{i\in[d]}\qty(\phi_i(\lams;\dis,\necr{[i]}))^{1/\alpha}
    &\leq
    \frac{\alpha+1}{\alpha^2}\int_0^1 u^{\frac{1}{\alpha}-1}
    \sum_{i\in[d]}\easypstar{i}{u;\necr{[i]}}\dd u\\
    &\leq
    \frac{\alpha+1}{\alpha^2}\int_0^1 u^{\frac{1}{\alpha}-1}\qty(\frac{m}{u}\land d)\dd u\\
    &=
    \frac{\alpha+1}{\alpha^2}
    \qty(
    d\int_0^{m/d} u^{\frac{1}{\alpha}-1}\dd u
    +
    \int_{m/d}^{1} u^{\frac{1}{\alpha}-1}\frac{m}{u}\dd u
    )\\
    &=
    \frac{\alpha+1}{\alpha(\alpha-1)}
    \qty(
    \alpha m^{1/\alpha}d^{1-1/\alpha}-m),
\end{align*}
which completes the proof.
\hfill$\square$

\subsection{Proof of Lemma~\ref{lem:adv_bound_term}}
\label{sec:proof_adv_bound_term}

\begin{proof}
    Since $(\hat{L}_t+\hat{\ell}_t)_j\geq (\hat{L}_t + (\ell_{t,i}\widehat{w_{t,i}^{-1}}) e_i)_j$ for all $j\neq i$, and $(\hat{L}_t+\hat{\ell}_t)_i=(\hat{L}_t + (\ell_{t,i}\widehat{w_{t,i}^{-1}}) e_i)_i$,
    by Lemma~\ref{lem:phi_monotone} we immediately have
    \begin{equation*}
        \phi_i(\eta_t(\hat{L}_t + (\ell_{t,i}\widehat{w_{t,i}^{-1}}) e_i);\dis)\leq\phi_i(\eta_t(\hat{L}_t+\hat{\ell}_t);\dis),
    \end{equation*}
    with which we have
    \begin{align*}
        \phi_i(\eta_t\hat{L}_t;\dis)-\phi_i(\eta_t(\hat{L}_t+\hat{\ell}_t);\dis)
        &\leq\phi_i(\eta_t\hat{L}_t;\dis) - \phi_i(\eta_t(\hat{L}_t + (\ell_{t,i}\widehat{w_{t,i}^{-1}}) e_i);\dis)\\ 
        &= \int_0^{\eta_t\ell_{t,i} \widehat{w_{t,i}^{-1}}} -\frac{\partial \phi_i}{\partial x}\qty(\eta_t\hat{L}_t + x e_i;\dis) \dd x.\numberthis\label{eq:phi_i_diff}
    \end{align*}
    Recalling \eqref{eq:definition_phi_extension_no_underline} and the definition of $\blk{i}{\lambda}$, we obtain the following expression for any $\lambda\in\mathbb{R}^d$:
    \begin{equation*}
        \phi_i(\lambda;\dis)=
        \int_{\nu-\lambda_{\blk{i}{\lambda}}}^\infty 
        \sP_{\{r_k\}_{k\in[d]\setminus\{i\}} \sim \dis,\,r_i=z+\lambda_i}
        \qty[\sele_i(r-\lambda;\nec)=1]
    \dd F(z+\lambda_i).
    \end{equation*}
    Here, $\sP[\cdot]$ in the integrand above does not depend on $\lambda_i$ because the $i$-th component of $r-\lambda$ is fixed to $z$ under $r_i=z+\lambda_i$. 
    Therefore, 
    one can see that $\phi'_i(\lambda;\dis)=\frac{\partial \phi_i}{\partial \lambda_i}(\lambda;\dis)$ satisfies
    \begin{align*}
        -\phi'_i&(\lambda;\dis)
        =
        \int_{\nu-\lambda_{\blk{i}{\lambda}}}^\infty -f'(z+\lambda_i)\sP_{\{r_k\}_{k\in[d]\setminus\{i\}} \sim \dis,\,r_i=z+\lambda_i}
        \qty[\sele_i(r-\lambda;\nec)=1] \dd z\\
        &\hspace{3cm}-\ind[i=\blk{i}{\lambda}]f(\nu)\sP_{\{r_k\}_{k\in[d]\setminus\{i\}} \sim \dis,\,r_i=\nu}
        \qty[\sele_i(r-\lambda;\nec)=1]\\
        &\leq
        \int_{\nu-\lambda_{\blk{i}{\lambda}}}^\infty -f'(z+\lambda_i)\sP_{\{r_k\}_{k\in[d]\setminus\{i\}} \sim \dis,\,r_i=z+\lambda_i}
        \qty[\sele_i(r-\lambda;\nec)=1] \dd z\\
        &=
        \int_{\nu}^\infty -f'(z+\ul_i)\sP_{\{r_k\}_{k\in[d]\setminus\{i\}} \sim \dis,\,r_i=z+\ul_i}
        \qty[\sele_i(r-\ul\pn{i};\nec)=1] \dd z,\numberthis\label{eq:phi_derivative_expression}
    \end{align*}
    where the last equality holds by $\ul_i=\ul_i\pn{i}=\lambda_i-\lambda_{\blk{i}{\lambda}}$ for any $i\in[d]$.

    Now we divide the proof into two cases.
    When $\dis=\Fdis$, since for $x>0$,
    \begin{equation*}
        f'(x)=\qty(-\frac{\alpha(\alpha+1)}{x^{\alpha+2}}+\frac{\alpha^2}{x^{2(\alpha+1)}})e^{-1/x^\alpha},
    \end{equation*}
    we have
    \begin{equation}\label{eq:f_derivative_relation}
        -f'(x)=\qty(\frac{\alpha(\alpha+1)}{x^{\alpha+2}}-\frac{\alpha^2}{x^{2(\alpha+1)}})e^{-1/x^\alpha}\leq
        \frac{\alpha(\alpha+1)}{x^{\alpha+2}}e^{-1/x^\alpha}
        =\frac{\alpha+1}{x}f(x).
    \end{equation}
    Recall the definition of $J_i(\cdot)$ given in \eqref{eq:definition_J}. 
    By \eqref{eq:phi_i_diff}--\eqref{eq:f_derivative_relation} we have
    \begin{align*}
        \lefteqn{\phi_i(\eta_t\hat{L}_t;\Fdis)-\phi_i(\eta_t(\hat{L}_t+\hat{\ell}_t);\Fdis)}\\
        &\leq
        \int_0^{\eta_t\ell_{t,i} \widehat{w_{t,i}^{-1}}}\int_{\nu}^\infty \frac{\alpha+1}{z+\underline{\eta_t\hat{L}_t+xe_i}_i}\sP_{\substack{\{r_k\}_{k\in[d]\setminus\{i\}} \sim \dis,\\
        r_i=z+\underline{\eta_t\hat{L}_t+xe_i}_i}}
        \qty[\sele_i(r-\underline{\eta_t\hat{L}_t + x e_i}\pn{i};\nec)=1] \dd F(z+\underline{\eta_t\hat{L}_t+xe_i}_i)\dd x\\
        &=
        (\alpha+1)\int_0^{\eta_t\ell_{t,i} \widehat{w_{t,i}^{-1}}} J_i\qty(\underline{\eta_t\hat{L}_t + x e_i}\pn{i};\Fdis) \dd x \\
        &\leq (\alpha+1)\int_0^{\eta_t\ell_{t,i} \widehat{w_{t,i}^{-1}}} J_i\qty(\eta_t\underline{\hat{L}}_t\pn{i};\Fdis) \dd x\numberthis\label{eq:J_mono_frechet}\\
        &=(\alpha+1)\eta_t\ell_{t,i}  J_i\qty(\eta_t\underline{\hat{L}}_t\pn{i};\Fdis)\widehat{w_{t,i}^{-1}},
    \end{align*}
    where \eqref{eq:J_mono_frechet} follows from the monotonicity of $J_i(\eta_t\underline{\hat{L}}\pn{i}_t;\Fdis)$.
    We obtain the same result for Pareto perturbations ($\dis=\Pdis$), since $-f'(x)=(\alpha+1)f(x)/x$ holds as an equality instead of the inequality in \eqref{eq:f_derivative_relation}.

    Since $\widehat{{w_{t,i}^{-1}}}$ follows the geometric distribution with expectation $1/w_{t,i}$ given $\hat{L}_t$ and $a_{t,i}$, we have
    \begin{equation}\label{eq:square_expectation}
        \E\qty[\widehat{{w_{t,i}^{-1}}}^2\m\hat{L}_t,a_{t,i}]=
        \E^2\qty[\widehat{{w_{t,i}^{-1}}}\m\hat{L}_t,a_{t,i}]+
        \mathrm{Var}\qty[\widehat{{w_{t,i}^{-1}}}\m\hat{L}_t,a_{t,i}]\leq \frac{2}{w_{t,i}^2}.
    \end{equation}
    Since $\hat{\ell}_{t,i}=\ind[a_{t,i}=1]\ell_{t,i}\widehat{{w_{t,i}^{-1}}}$, for $\dis\in\qty{\Fdis,\Pdis}$ we obtain
    \begin{align*}
        \lefteqn{\E\qty[\hat{\ell}_{t,i}\qty(\phi_i\qty(\eta_t\hat{L}_t;\dis)-\phi_i\qty(\eta_t\qty(\hat{L}_t+\hat{\ell}_t);\dis))\m \hat{L}_t]}\\
        &\leq\E\qty[\ind\qty[a_{t,i}=1]\ell_{t,i}\widehat{{w_{t,i}^{-1}}}\cdot(\alpha+1)\eta_t\ell_{t,i}  J_i\qty(\eta_t\underline{\hat{L}}_t\pn{i};\dis)\widehat{w_{t,i}^{-1}}\m \hat{L}_t]\\
        &\leq2(\alpha+1)\eta_t\E\qty[w_{t,i}\frac{\ell^2_{t,i}J_i\qty(\eta_t\underline{\hat{L}}_t\pn{i};\dis)}{w^2_{t,i}}\m \hat{L}_t]
        \displaybreak[0]\\
        &\leq2(\alpha+1)\eta_t\E\qty[\frac{J_i\qty(\eta_t\underline{\hat{L}}_t\pn{i};\dis)}{I_i\qty(\eta_t\underline{\hat{L}}_t\pn{i};\dis)}\m \hat{L}_t]
        \tag{by $w_{t,i}=\phi_i(\eta_t\hat{L}_t;\dis)$ and \eqref{eq:phi_I_relation}}
        \displaybreak[0]\\
        &\leq
        \frac{2(\alpha+1)}{0\lor\underline{\hat{L}}_{t,i}}
        \land
        2(\alpha+1)\eta_t
        \qty(\phi_i(\lams;\dis,\necr{[i]}))^{1/\alpha},
    \end{align*}
    where the last inequality follows from Lemma~\ref{lem:sigma_i}.
    Then, by summing over $i\in[d]$ and applying Lemma~\ref{lem:sigma_i} again, we obtain
    \begin{align*}
        \E\qty[\bigl\langle\hat\ell_t,\
        \phi(\eta_t\hat L_t;\dis)-\phi(\eta_t\hat L_{t+1};\dis)\bigr\rangle\m \hat{L}_t]
        &\leq
        2(\alpha+1)\eta_t\sum_{i=1}^d\qty(\phi_i(\lams;\dis,\necr{[i]}))^{1/\alpha}\\
        &\leq
        \frac{2(\alpha+1)^2}{\alpha-1}
        \eta_t m^{\frac{1}{\alpha}}d^{1-\frac{1}{\alpha}},
    \end{align*}
    which completes the proof.
\end{proof}

\subsection{Proof of Theorem~\ref{thm:adv_bound}}\label{sec:proof_adv_bound}

By combining Lemmas~\ref{lem:regret_decomposition}, \ref{lem:penalty_adv_bound}, and \ref{lem:adv_bound_term} 
with $\eta_t=\frac{c}{\sqrt{t}}m^{\frac{1}{2}-\frac{1}{\alpha}}d^{\frac{1}{\alpha}-\frac{1}{2}}$, for $\dis=\Pdis$ we have
\begin{align*}
    \R(T)&\leq
    \frac{2(\alpha+1)^2}{\alpha-1} cm^{\frac{1}{2}-\frac{1}{\alpha}}d^{\frac{1}{\alpha}-\frac{1}{2}}\cdot m^{\frac{1}{\alpha}}d^{1-\frac{1}{\alpha}}\sum_{t=1}^T \frac{1}{\sqrt{t}}\\
    &\hspace{2cm}+
    \frac{\qty(\frac{\alpha}{\alpha-1}(m-1)^{1-\frac{1}{\alpha}}+\Gamma\qty(1-\frac{1}{\alpha}))\qty(d+1)^{\frac{1}{\alpha}}}{cm^{\frac{1}{2}-\frac{1}{\alpha}}d^{\frac{1}{\alpha}-\frac{1}{2}}}\sum_{t=1}^T\qty(\sqrt{t+1}-\sqrt{t})
    \\
    &\hspace{2cm}+
    \frac{\qty(\frac{\alpha}{\alpha-1}(m-1)^{1-\frac{1}{\alpha}}+\Gamma\qty(1-\frac{1}{\alpha}))\qty(d+1)^{\frac{1}{\alpha}}}{cm^{\frac{1}{2}-\frac{1}{\alpha}}d^{\frac{1}{\alpha}-\frac{1}{2}}}\numberthis\label{eq:adv_bound_to_replace}
    \displaybreak[0]\\
    &\leq 
        \frac{4c(\alpha+1)^2}{\alpha-1}\sqrt{mdT}
        +\frac{\qty(\frac{\alpha}{\alpha-1}(1-\frac{1}{m})^{1-\frac{1}{\alpha}}+\Gamma\qty(1-\frac{1}{\alpha})m^{\frac{1}{\alpha}-1})(1+\frac{1}{d})^{\frac{1}{\alpha}}}{c}
    \qty(\sqrt{mdT}+\sqrt{md}),
\end{align*}
which implies that the regret is of order $O(\sqrt{mdT})$.

The analysis for $\dis=\Fdis$ is similar with the difference that $\frac{\qty(\frac{\alpha}{\alpha-1}(m-1)^{1-\frac{1}{\alpha}}+\Gamma\qty(1-\frac{1}{\alpha}))\qty(d+1)^{\frac{1}{\alpha}}}{cm^{\frac{1}{2}-\frac{1}{\alpha}}d^{\frac{1}{\alpha}-\frac{1}{2}}}$ in \eqref{eq:adv_bound_to_replace} is replaced with $\frac{\qty(\frac{\alpha}{\alpha-1}(m-1)^{1-\frac{1}{\alpha}}+\Gamma\qty(1-\frac{1}{\alpha}))\qty(d+1)^{\frac{1}{\alpha}}+m}{cm^{\frac{1}{2}-\frac{1}{\alpha}}d^{\frac{1}{\alpha}-\frac{1}{2}}}$ coming from Lemmas~\ref{lem:regret_decomposition} and \ref{lem:penalty_adv_bound}. 
This leads to the bound of
\begin{equation*}
    \R(T)\leq 
        \frac{4c(\alpha+1)^2}{\alpha-1}
        \sqrt{mdT}
    +\frac{\qty(\frac{\alpha}{\alpha-1}(1-\frac{1}{m})^{1-\frac{1}{\alpha}}+\Gamma\qty(1-\frac{1}{\alpha})m^{\frac{1}{\alpha}-1})(1+\frac{1}{d})^{\frac{1}{\alpha}}+\qty(\frac{m}{d})^{\frac{1}{\alpha}}}{c}
    \qty(\sqrt{mdT}+\sqrt{md}),
\end{equation*}
which also implies that the regret is of order $O(\sqrt{mdT})$.
\hfill$\square$

\section{Regret Bound for Stochastic Bandits}\label{sec:stab_sto}

This section proves Theorems~\ref{thm:sto_bound} and~\ref{thm:sto_bound_alpha_not2},
our regret bounds for the stochastic setting.
The proofs build on the self-bounding technique~\citep{zimmert2021tsallis},
a standard approach for establishing best-of-both-worlds guarantees of FTRL.
This approach requires two key ingredients, namely, a lower bound on the regret and an upper bound on the regret term associated with the optimal action.
We derive these two parts in Sections~\ref{subsec:regret_lower_bounds}
and~\ref{subsec:regret_optimal_action}, respectively, by extending the analysis of the $m$-set semi-bandit problem~\citep{chen2026further} to the matroid semi-bandit problem,
which in turn builds on earlier work on the multi-armed bandit
problem~\citep{pmlr-v201-honda23a,pmlr-v247-lee24a}.
Finally, we complete the proofs in Section~\ref{subsec:proofs_sto_bounds}.

Following the analysis of $m$-set semi-bandits~\citep{chen2026further},
we define the events $F_t$ and $D_t$ for Fr\'{e}chet and Pareto
perturbations, respectively, as follows.
\begin{align*}
    F_t &\coloneqq\qty{\sum_{i:a_i^*=0}\frac{1}{(\eta_t\underline{\hat{L}}_{t,i})^\alpha}\leq\frac{1}{m^{\alpha/(\alpha-1)}}
    \mbox{ and }\forall i:a_i^*=0,\,\eta_t\underline{\hat{L}}_{t,i}>0},\\
    D_t &\coloneqq\qty{\sum_{i:a_i^*=0} \frac{1}{(2^{\frac{1}{\alpha}}+\eta_t\underline{\hat{L}}_{t,i})^\alpha}
    \leq 
    \frac{1}{(2^{\frac{1}{\alpha}}+m^{1/(\alpha-1)})^\alpha}
    \mbox{ and }\forall i:a_i^*=0,\,\eta_t\underline{\hat{L}}_{t,i}>0}.
\end{align*} 
One can see that both events imply the key property
\begin{equation}\label{eq:key_property}
    \eta_t\underline{\hat{L}}_{t,i}\leq 0,\,\forall i:a_i^*=1,\quad
    \text{and}\quad
    \eta_t\underline{\hat{L}}_{t,i}\geq m^{1/(\alpha-1)},\,\forall i:a_i^*=0.
\end{equation}
Here, the property for base-arms $i$ with $a_i^*=0$ follows directly from the definitions
of $F_t$ and $D_t$. For base-arms with $a_i^*=1$, note that by
\eqref{eq:r_start}, every base-arm $i$ with $\sele_i(-\eta_t\hat L_t;\nec)=1$
satisfies $\eta_t\underline{\hat L}_{t,i}\le 0$, so no base-arm with
$a_i^*=0$ is selected. Since exactly $m$ base-arms are selected and
$\left\lVert a^*\right\rVert_1 =m$, it follows that
$\sele(-\eta_t\hat L_t;\nec)=a^*$. Consequently,
$\eta_t\underline{\hat L}_{t,i}\le 0$ for all $i$ with $a_i^*=1$.

\subsection{Regret Lower Bounds}\label{subsec:regret_lower_bounds}

In this part, we provide the regret lower bounds on $\sum_{i:a^*_i=0}\Delta_{i,\min} w_{t,i}$ in Lemma~\ref{lem:lower_bound_Dt}.

For any $i\in[d]$ with $a_i^*=1$, we define
\begin{equation*}
    \Bex{i}\coloneqq\qty{i}\cup\qty{i'\in[d]:\,a^*-e_i+e_{i'}\in\nec},
\end{equation*}
which consists of $i$ itself and the base-arms $i'$ with $a_{i'}^*=0$ that can replace $i$ in $a^*$ while preserving feasibility.

The following lemma shows that $\blkomit{t,i'}=i$ implies $i'\in\Bex{i}$ under the following sign conditions.
\begin{lemma}\label{lem:optimal_action_psi_ex}
    Suppose that $\eta_t\underline{\hat{L}}_{t,j}\leq0$ for any $j$ with $a_j^*=1$ and $\eta_t\underline{\hat{L}}_{t,j}>0$ for any $j$ with $a_j^*=0$.
    Then, for any $i$ with $a_i^*=1$ and any $i'$ with $a_{i'}^*=0$ such that $\blkomit{t,i'}=i$, we have $i'\in\Bex{i}$.
\end{lemma}

\begin{proof}
    Since $a_i^*=1$ and $a_{i'}^*=0$, we know that $\blkomit{t,i'}\neq i'$.
    Define
    \begin{equation*}
        \armset_{\prec i}\coloneqq\qty{j:\sigma_{j}<\sigma_{i}}
        \quad\text{and}\quad
        \armset_{\preceq i}\coloneqq\armset_{\prec i}\cup\qty{i}.
    \end{equation*}
    By the definition of $\blkomit{t,i'}$, $i'$ is selected if the set of base-arms preceding $i'$ is exactly $\armset_{\prec i}$, and is not selected if the set of base-arms preceding $i'$ is $\armset_{\preceq i}$.

    Consider the permutation $\sigma=(\sigma_1,\ldots,\sigma_d)$ induced by $\hat{L}_t$ without perturbation.
    Under the assumed sign conditions, since $\underline{\hat{L}}_{t,j}\leq 0$ for every $j$ with $a_j^*=1$, while $\underline{\hat{L}}_{t,j}>0$ for every $j$ with $a_j^*=0$, the set of selected base-arms is exactly $A^*=\{j:a^*_j=1\}$.

    Now we only move the base-arm $i'$ to the position such that the set of base-arms preceding $i'$ is exactly $\armset_{\prec i}$.
    Then, the base-arms selected while processing $\armset_{\prec i}$ are exactly $\armset_{\prec i}\cap A^*$.
    Combining this with the fact that $i'$ is selected when the set of base-arms preceding $i'$ is $\armset_{\prec i}$, we immediately obtain
    \begin{equation}\label{eq:exchange_feasible1}
        (\armset_{\prec i}\cap A^*)\cup\qty{i'}
        \in\supp{\sI}.
    \end{equation}
    Now consider the case where we move the base-arm $i'$ to the position such that the set of base-arms preceding $i'$ is $\armset_{\preceq i}$, and the permutation of $[d]\setminus\{i'\}$ is still $\sigma$.
    Then, the base-arms selected while processing $\armset_{\preceq i}$ are exactly $\armset_{\preceq i}\cap A^*$, and $i'$ is not selected,
    which implies that
    \begin{equation}\label{eq:exchange_feasible2}
        (\armset_{\prec i}\cap A^*)\cup\qty{i,i'}
        \notin\supp{\sI}.
    \end{equation}
    Let $A^*_{\prec i}=\armset_{\prec i}\cap A^*$ and $A^*_{\succeq i}=A^*\setminus\armset_{\prec i}$.
    By \eqref{eq:exchange_feasible1} and the matroid augmentation property, there must exist a base-arm $j\in A^*_{\succeq i}$ such that
    \begin{equation*}
        A^*\setminus\qty{j}\cup\qty{i'}
        =
        A^*_{\prec i}\cup (A^*_{\succeq i}\setminus\qty{j})\cup\qty{i'}
        \in\supp{\base}.
    \end{equation*}
    Combining this with \eqref{eq:exchange_feasible2}, it is necessary that $i\notin A^*_{\succeq i}\setminus\qty{j}$, which implies that $j=i$.
    Therefore, we have $A^*\setminus\qty{i}\cup\qty{i'}\in\supp{\base}$, which completes the proof that $i'\in\Bex{i}$.
\end{proof}

We next present the following lemma, which provides a sufficient condition for a base-arm to be selected.

\begin{lemma}\label{lem:optimal_action_sufficient_condition}
    For any $i$ with $a_i^*=1$ and any $i'\in\Bex{i}$, possibly $i$ itself, $i'$ is selected if
    \begin{equation*}
        i'=\argmax_{j\in\Bex{i}}\qty(r_{t,j}-\eta_t\hat{L}_{t,j}).
    \end{equation*}
\end{lemma}

\begin{proof}
    We write
    \begin{align*}
        A^*\setminus\qty{i}&=\qty{\alpha_1,\alpha_2,\cdots,\alpha_{s_1}},\\
        [d]\setminus\qty(A^*\cup\Bex{i})
        &=\qty{\beta_1,\beta_2,\cdots,\beta_{s_2}},\\
        \Bex{i}\setminus\qty{i'}&=\qty{\gamma_1,\gamma_2,\cdots,\gamma_{s_3}},
    \end{align*}
    where the three sets above, together with $i'$, form a partition of $[d]$.
    Then, for the permutation of $[d]$ as
    \begin{equation*}
        \alpha_1,\alpha_2,\cdots,\alpha_{s_1},
        \beta_1,\beta_2,\cdots,\beta_{s_2},
        i',
        \gamma_1,\gamma_2,\cdots,\gamma_{s_3},
    \end{equation*}
    we now consider their selection according to the greedy method.
    
    Firstly, since $A^*\setminus\qty{i}$ is feasible, all base-arms $\alpha_1,\alpha_2,\ldots,\alpha_{s_1}$ are selected by the greedy method as they are processed. 
    Then, by the definition of $\Bex{i}$, among the remaining base-arms, only those in $\Bex{i}$ can be selected from this point onward.
    Therefore, since $\beta_1,\beta_2,\ldots,\beta_{s_2}\notin\Bex{i}$, they are not selected.
    Then, the next base-arm to be processed is $i'$.
    Since $i'\in\Bex{i}$, it is selected by the greedy method.

    Note that for any permutation induced by $i'=\argmax_{j\in\Bex{i}}(r_{t,j}-\eta_t\hat{L}_{t,j})$, the set of base-arms preceding $i'$ must be a subset of $\{\alpha_1,\alpha_2,\ldots,\alpha_{s_1},\beta_1,\beta_2,\ldots,\beta_{s_2}\}$.
    Therefore, by Lemmas~\ref{lem:greedy_equiv} and \ref{lem:greedy_remove}, $i'$ is selected by the greedy method for any such permutation, which concludes the proof.
\end{proof}

\begin{lemma}\label{lem:lower_bound_Dt}
    Let $\Delta_{\min}\coloneqq\min_{i:a_i^*=0}\Delta_{i,\min}$. 
    Then, there exist some distribution-dependent constants $c_{s,1}(\dis),\allowbreak c_{s,2}(\dis)\in\qty(0,1)$ and
    \begin{equation*}
        \wlow=
        \begin{cases}
            1/e, &\text{if }\dis=\Fdis,\\
            0.14, &\text{if }\dis=\Pdis,
        \end{cases}
    \end{equation*}
    such that
    \begin{enumerate}[label=(\roman*)]
        \item\label{item:lower_bound_Ft} On $F_t$ or $D_t$, $\sum_{i:a^*_i=0}\Delta_{i,\min} w_{t,i}\geq c_{s,1}(\dis)\sum_{i:a^*_i=0}\frac{\Delta_{i,\min}}{(\eta_t\underline{\hat{L}}_{t,i})^\alpha}$ and $w_{t,i}\geq \wlow$ for any $i$ with $a^*_i=1$.
        \item\label{item:lower_bound_Ftc} On $F_t^c$ or $D_t^c$, $\sum_{i:a^*_i=0}\Delta_{i,\min} w_{t,i}\geq c_{s,2}(\dis)\Delta_{\min}/m^{\alpha/(\alpha-1)}$.
    \end{enumerate}
\end{lemma}

\begin{proof}
    We first derive two probability bounds assuming that $\eta_t\underline{\hat{L}}_{t,j}\leq0$ for every $j$ with $a_j^*=1$ and $\eta_t\underline{\hat{L}}_{t,j}>0$ for every $j$ with $a_j^*=0$.
    Under these conditions, Lemma~\ref{lem:optimal_action_psi_ex} implies that $i\in\Bex{\blkomit{t,i}}$ for any $i$ with $a_i^*=0$.
    Then, by Lemma~\ref{lem:optimal_action_sufficient_condition}, we have
    \begin{equation*}
        \qty{i=\argmax_{j\in\Bex{\blkomit{t,i}}}\qty(r_{t,j}-\eta_t\hat{L}_{t,j})}
        \subset
        \qty{a_{t,i}=1}
        \mbox{ and }
        \qty{\blkomit{t,i}=\argmax_{j\in\Bex{\blkomit{t,i}}}\qty(r_{t,j}-\eta_t\hat{L}_{t,j})}
        \subset
        \qty{a_{t,\blkomit{t,i}}=1}.
    \end{equation*}
    Therefore, we have
    \begin{align*}
        w_{t,i}&=\sP\qty[a_{t,i}=1\m \hat{L}_t]\\
        &\geq
        \sP\qty[i=\argmax_{j\in\Bex{\blkomit{t,i}}}\qty(r_{t,j}-\eta_t\hat{L}_{t,j})\m \hat{L}_t]\\
        &=
        \sP\qty[i=\argmax_{j\in\Bex{\blkomit{t,i}}}\qty(r_{t,j}-\eta_t\qty(\hat{L}_{t,j}-\hat{L}_{t,\blkomit{t,i}}))\m \hat{L}_t]\\
        &=
        \sP\qty[i=\argmax_{j\in\Bex{\blkomit{t,i}}\setminus\{\blkomit{t,i}\}}\qty(r_{t,j}-\eta_t\qty(\hat{L}_{t,j}-\hat{L}_{t,\blkomit{t,i}}))
        ,\,r_{t,i}-\eta_t\underline{\hat{L}}_{t,i}\geq r_{t,\blkomit{t,i}}\m \hat{L}_t],
        \numberthis\label{eq:lower_bound_Dt_wt}
    \end{align*}
    and similarly,
    \begin{align*}
        w_{t,\blkomit{t,i}}
        &\geq
        \sP\qty[\blkomit{t,i}=\argmax_{j\in\Bex{\blkomit{t,i}}}\qty(r_{t,j}-\eta_t\qty(\hat{L}_{t,j}-\hat{L}_{t,\blkomit{t,i}}))\m \hat{L}_t]\\
        &=
        \sP\qty[r_{t,\blkomit{t,i}}\geq\max_{j\in\Bex{\blkomit{t,i}}\setminus\{\blkomit{t,i}\}}\qty(r_{t,j}-\eta_t\qty(\hat{L}_{t,j}-\hat{L}_{t,\blkomit{t,i}}))\m \hat{L}_t]
    \end{align*}

    Next, we show that $\hat{L}_{t,j}-\hat{L}_{t,\blkomit{t,i}}\geq\underline{\hat{L}}_{t,j}$ for any $j\in\Bex{\blkomit{t,i}}\setminus\{i,\blkomit{t,i}\}$.
    Define
    \begin{equation*}
        \armset_{\prec \blkomit{t,i}}\coloneqq\qty{j:\sigma_{j}<\sigma_{\blkomit{t,i}}}.
    \end{equation*}
    Under these sign conditions, since $\underline{\hat{L}}_{t,j}\leq 0$ for any $j$ with $a_j^*=1$, and $\underline{\hat{L}}_{t,j}>0$ for any $j$ with $a_j^*=0$, the set of base-arms selected while processing $\armset_{\prec \blkomit{t,i}}$ is exactly $\armset_{\prec \blkomit{t,i}}\cap A^*$. Since $j\in\Bex{\blkomit{t,i}}$, we have $a^*-e_{\blkomit{t,i}}+e_j\in\nec$. Moreover, since $\blkomit{t,i}\notin\armset_{\prec \blkomit{t,i}}\cap A^*$, base-arm $j$ can be selected by the greedy method when it is processed immediately after $\armset_{\prec \blkomit{t,i}}$, i.e., immediately before $\blkomit{t,i}$.
    In addition, by the definition of $\blkomit{t,j}$, $j$ is not selected if it is processed after $\blkomit{t,j}$, which implies that $\blkomit{t,i}$ appears no later than $\blkomit{t,j}$ (including the case $\blkomit{t,i}=\blkomit{t,j}$) in the permutation induced by $\hat{L}_t$, and thus $\hat{L}_{t,\blkomit{t,i}}\leq\hat{L}_{t,\blkomit{t,j}}$.
    Therefore, for any $j\in\Bex{\blkomit{t,i}}\setminus\{i,\blkomit{t,i}\}$, we have
    \begin{equation*}
        \hat{L}_{t,j}-\hat{L}_{t,\blkomit{t,i}}\geq
        \hat{L}_{t,j}-\hat{L}_{t,\blkomit{t,j}}=
        \underline{\hat{L}}_{t,j}.
    \end{equation*}
    Combining this with \eqref{eq:lower_bound_Dt_wt} and the fact that $\hat{L}_{t,i}-\hat{L}_{t,\blkomit{t,i}}=\underline{\hat{L}}_{t,i}$, we have
    \begin{align*}
        w_{t,i}
        &\geq
        \sP\qty[i=\argmax_{j\in\Bex{\blkomit{t,i}}\setminus\{\blkomit{t,i}\}}\qty(r_{t,j}-\eta_t\underline{\hat{L}}_{t,j})
        ,\,r_{t,i}-\eta_t\underline{\hat{L}}_{t,i}\geq r_{t,\blkomit{t,i}}\m \hat{L}_t]\\
        &\geq
        \sP\qty[i=\argmax_{j:a^*_j=0}\qty(r_{t,j}-\eta_t\underline{\hat{L}}_{t,j})
        ,\,r_{t,i}-\eta_t\underline{\hat{L}}_{t,i}\geq r_{t,\blkomit{t,i}}
        \m \hat{L}_t].\numberthis\label{eq:lower_bound_to_MAB1}
    \end{align*}
    Similarly, we have
    \begin{equation}\label{eq:lower_bound_to_MAB2}
        w_{t,\blkomit{t,i}}
        \geq
        \sP\qty[r_{t,\blkomit{t,i}}\geq\max_{j\in\Bex{\blkomit{t,i}}\setminus\{\blkomit{t,i}\}}\qty(r_{t,j}-\eta_t\underline{\hat{L}}_{t,j})\m \hat{L}_t]
        \geq
        \sP\qty[r_{t,\blkomit{t,i}}\geq\max_{j:a^*_j=0}\qty(r_{t,j}-\eta_t\underline{\hat{L}}_{t,j})\m \hat{L}_t].
    \end{equation}
    Note that \eqref{eq:lower_bound_to_MAB1} and \eqref{eq:lower_bound_to_MAB2} are respectively the selection probabilities of arm $i$ with $a_i^*=0$ and arm $\blkomit{t,i}$ in an MAB setting with $d-m+1$ arms, consisting of $\{\blkomit{t,i}\}\cup\{j:a^*_j=0\}$,
    where the corresponding cumulative loss vector assigns $\underline{\hat{L}}_{t,j}$ to each arm $j$ with $a_j^*=0$ and $0$ to arm $\blkomit{t,i}$.
    Lemma~21 of \citet{pmlr-v247-lee24a} shows that, if $\{\sum_{i\in \armset_0\setminus\{j\}}\frac{1}{(\eta_t\underline{\hat{L}}_{t,i})^\alpha}\leq 1\}$ holds then\footnote{%
In Lemma~21 of \citet{pmlr-v247-lee24a}, the arm $j$ is assumed to be optimal, but this assumption is not used in the proof.
Likewise, the proof does not use the specific definition of MAB gaps, so it also applies to the nonnegative matroid gaps $\Delta_{j,\min}$ for $j$ with $a_j^*=0$.}, in the case of $m=1$ with $\nec=\{e_1,\ldots,e_d\}$, the selection probabilities of arm $i$ with $a_i^*=0$ and arm $\blkomit{t,i}$ satisfy the following inequalities:
    \begin{equation}\label{eq:lower_bounds_mab_Ft}
        \sum_{i\in \armset_0\setminus\{j\}}\Delta_{i,\min} w_{t,i}\geq
        c_{s,1}(\Fdis)\sum_{i\in \armset_0\setminus\{j\}}\frac{\Delta_{i,\min}}{(\eta_t\underline{\hat{L}}_{t,i})^\alpha}
        \quad\text{and}\quad
        w_{t,\blkomit{t,i}}\geq \frac{1}{e},
    \end{equation}
where
$c_{s,1}(\Fdis)=\max_{b>0} \frac{\exp\qty( -\qty( 1 + \frac{1}{b^\alpha} ))}{(1+b)^\alpha} \in (0,1)$.

With $\armset_0:=\qty{i:a_i^*=0}\cup\qty{\blkomit{t,i}}$ and $j:=\blkomit{t,i}$, we can apply this result to the above setting because the condition for \eqref{eq:lower_bounds_mab_Ft} indeed holds under $F_t$ since
\begin{multline*}
        F_t=\qty{\sum_{i:a_i^*=0}\frac{1}{(\eta_t\underline{\hat{L}}_{t,i})^\alpha}\leq\frac{1}{m^{\alpha/(\alpha-1)}}\mbox{ and }\forall i:a_i^*=0,\,\eta_t\underline{\hat{L}}_{t,i}>0}
        \\
        \subset 
        \qty{\sum_{i:a_i^*=0}\frac{1}{(\eta_t\underline{\hat{L}}_{t,i})^\alpha}\leq 1}
        =
        \qty{\sum_{i\in \armset_0\setminus \{j\}}\frac{1}{(\eta_t\underline{\hat{L}}_{t,i})^\alpha}\leq 1}.\n
\end{multline*}
Then we obtain the following two results on $F_t$ for matroid semi-bandits.
    First, we have
    \begin{align*}
        \sum_{i:a_i^*=0} \Delta_{i,\min} w_{t,i}
        &\geq
        \sum_{i:a_i^*=0} \Delta_{i,\min} \sP\qty[i=\argmax_{j:a^*_j=0}\qty(r_{t,j}-\eta_t\underline{\hat{L}}_{t,j})
        ,\,r_{t,i}-\eta_t\underline{\hat{L}}_{t,i}\geq r_{t,\blkomit{t,i}}
        \m \hat{L}_t]\tag{by \eqref{eq:lower_bound_to_MAB1}}\\
        &\geq c_{s,1}(\Fdis)\sum_{i:a_i^*=0}\frac{\Delta_{i,\min}}{(\eta_t\underline{\hat{L}}_{t,i})^\alpha}.\tag{by the first inequality in \eqref{eq:lower_bounds_mab_Ft}}
    \end{align*}
    Second, for any $i$ with $a^*_i=1$, by considering the set $\Bex{i}$ in the analysis above we obtain
    \begin{equation*}
        w_{t,i}\geq \frac{1}{e},
    \end{equation*}
    which follows from the second inequality in \eqref{eq:lower_bounds_mab_Ft}.
    Combining these two results proves the results of $F_t$ and $\Fdis$ of part~\ref{item:lower_bound_Ft}.

    For the results of $D_t$ and $\Pdis$ of the same part, we use the same argument with Lemma~22 of \citet{pmlr-v247-lee24a} in place of Lemma~21.

    Note that since $F(2^{\frac{1}{\alpha}}+m^{1/(\alpha-1)})\geq F(2^{\frac{1}{\alpha}}+1)$, the event $D_t$ defined here satisfies
    \begin{multline*}
        D_t
        =
        \qty{\sum_{i:a_i^*=0} \frac{1}{(2^{\frac{1}{\alpha}}+\eta_t\underline{\hat{L}}_{t,i})^\alpha}
        \leq 
        \frac{1}{(2^{\frac{1}{\alpha}}+m^{1/(\alpha-1)})^\alpha}
        \mbox{ and }\forall i:a_i^*=0,\,\eta_t\underline{\hat{L}}_{t,i}>0}
        \\
        \subset\qty{\sum_{i:a_i^*=0} \frac{1}{(2^{\frac{1}{\alpha}}+\eta_t\underline{\hat{L}}_{t,i})^\alpha}
        \leq 
        \frac{1}{(2^{\frac{1}{\alpha}}+1)^\alpha}}.
    \end{multline*}
    Here, the RHS is actually the event considered in Lemma~22 of \citet{pmlr-v247-lee24a} for the MAB setting.
    Then, on this event, 
    we apply the lower bounds established in their Lemma~22, which we restate in our notation as follows.

    Lemma~22 of \citet{pmlr-v247-lee24a} shows that, in the case of $m=1$ where $\|a^*\|_1=1$, by letting base-arm $\blkomit{t,i}$ be such that
    $a^*_{\blkomit{t,i}}=1$, we have
    \begin{equation}\label{eq:lower_bounds_mab_Dt}
        \sum_{i:a^*_i=0}\Delta_{i,\min} w_{t,i}\geq c_{s,1}(\Pdis)\sum_{i:a_i^*=0}\frac{\Delta_{i,\min}}{(\eta_t\underline{\hat{L}}_{t,i})^\alpha}
        \quad\text{and}\quad
        w_{t,\blkomit{t,i}}\geq 0.14,
    \end{equation}
    where $c_{s,1}(\Pdis)$ is given as $\frac{1}{e(2^{1/\alpha}+1)^\alpha}$.
    
    As in the Fr\'{e}chet case above, we obtain the following two results for matroid semi-bandits.
    First, we have
    \begin{align*}
        \sum_{i:a_i^*=0} \Delta_{i,\min} w_{t,i}
        &\geq
        \sum_{i:a_i^*=0} \Delta_{i,\min} \sP\qty[i=\argmax_{j:a^*_j=0}\qty(r_{t,j}-\eta_t\underline{\hat{L}}_{t,j})
        ,\,r_{t,i}-\eta_t\underline{\hat{L}}_{t,i}\geq r_{t,\blkomit{t,i}}
        \m \hat{L}_t]\tag{by \eqref{eq:lower_bound_to_MAB1}}\\
        &\geq c_{s,1}(\Pdis)\sum_{i:a_i^*=0}\frac{\Delta_{i,\min}}{(\eta_t\underline{\hat{L}}_{t,i})^\alpha}.\tag{by the first inequality in \eqref{eq:lower_bounds_mab_Dt}}
    \end{align*}
    The second is that, for any $i$ with $a^*_i=1$, considering the set $\Bex{i}$ in the analysis above gives
    \begin{equation*}
        w_{t,i}\geq 0.14,
    \end{equation*}
    which follows from the second inequality in \eqref{eq:lower_bounds_mab_Dt}. 
    Combining these two results completes the the results of $D_t$ and $\Pdis$ of part~\ref{item:lower_bound_Ft}.
    
    Next, we consider the lower bound on $F_t^c$ and $D_t^c$. Define 
    \begin{align*}
        B_t^- &\coloneqq \qty{\exists j:a_j^*=0, \eta_t\underline{\hat{L}}_{t,j}\leq 0},\\
        B_{t,F}^+ &\coloneqq \qty{\sum_{i:a_i^*=0}\frac{1}{(\eta_t\underline{\hat{L}}_{t,i})^\alpha}>\frac{1}{m^{\alpha/(\alpha-1)}}\mbox{ and }\forall i:a_i^*=0,\,\eta_t\underline{\hat{L}}_{t,i}>0},\displaybreak[0]\\
        B_{t,D}^+ &\coloneqq \Bigg\{\sum_{i:a_i^*=0} \frac{1}{(2^{\frac{1}{\alpha}}+\eta_t\underline{\hat{L}}_{t,i})^\alpha}
        >
        \frac{1}{(2^{\frac{1}{\alpha}}+m^{1/(\alpha-1)})^\alpha}
        \mbox{ and }\forall i:a_i^*=0,\,\eta_t\underline{\hat{L}}_{t,i}>0\Bigg\}.
    \end{align*}
    Then, we have
    \begin{equation*}
        F_t^c=B_t^-\cup B_{t,F}^+,
        \quad\text{and}\quad
        D_t^c=B_t^-\cup B_{t,D}^+.
    \end{equation*}
    We first consider $B_t^-$ for $\dis\in\{\Fdis,\Pdis\}$.
    On $B_t^-$, choose $j_0$ with $a^*_{j_0}=0$ and
$\underline{\hat{L}}_{t,j_0}\leq 0$, and let
\[
    \armset_{\prec j_0}\coloneqq
    \qty{j\in[d]:\hat{L}_{t,j}<\hat{L}_{t,j_0}}.
\]
By the definition of $\blkomit{t,j_0}$, the base-arm $j_0$ is selected
when it is processed immediately after $\armset_{\prec j_0}$, with all
base-arms tied with $j_0$ processed after $j_0$.
Since $A^*\cap \armset_{\prec j_0}\subset \armset_{\prec j_0}$, Lemma~\ref{lem:greedy_remove}
implies that $j_0$ is also selected immediately after
$A^*\cap \armset_{\prec j_0}$, and thus we have
\begin{equation*}
    (A^*\cap \armset_{\prec j_0})\cup\{j_0\}\in\supp{\sI}.
\end{equation*}
By repeatedly applying the matroid augmentation property
with the basis $A^*$, this independent set can be extended
to a basis contained in $A^*\cup\{j_0\}$.
This implies that there exists $j_1\in A^*\setminus \armset_{\prec j_0}$ such that
\begin{equation*}
        a^*-e_{j_1}+e_{j_0}\in\nec
    \quad\mbox{and}\quad
    \hat{L}_{t,j_1}\geq\hat{L}_{t,j_0}.
\end{equation*}
By combining this with the fact that $A^*\setminus\{j_1\}\in\supp{\sI}$, we have
    \begin{equation}\label{eq:action_set_relation}
        \neccr{A^*\setminus\qty{j_1}}{\{j_0,j_1\}}
        =
        \qty{e_{j_0},e_{j_1}},
    \end{equation}
    since $a^*-e_{j_1}+e_{j_0},a^*\in\nec$.
Therefore, we have
    \begin{align*}
        \sum_{i:a_i^*=0} \Delta_{i,\min} w_{t,i}
        &\geq
        \Delta_{\min}\sum_{i:a_i^*=0} \phi_i(\eta_t\hat{L}_t;\nec)\\
        &\geq 
        \lim_{\min_{i:a_i^*=0,i\neq j_0}\hat{L}_{t,i}\to +\infty}\Delta_{\min}\sum_{i:a_i^*=0} \phi_i(\eta_t\hat{L}_t;\nec)\tag*{(by Lemma~\ref{lem:phi_monotone})}\\
        &=
        \Delta_{\min}\cdot\phi_{j_0}(\eta_t\hat{L}_t;\necr{A^*\cup\{j_0\}})\\
        &\geq 
        \lim_{\max_{i:a_i^*=1,i\neq j_1}\hat{L}_{t,i}\to -\infty} \Delta_{\min}\cdot\phi_{j_0}(\eta_t\hat{L}_t;\necr{A^*\cup\{j_0\}})\tag*{(by Lemma~\ref{lem:phi_monotone})}\\
        &=
        \Delta_{\min}\cdot\phi_{j_0}(\eta_t\hat{L}_t;\neccr{A^*\setminus\qty{j_1}}{\{j_0,j_1\}})\\
        &=
        \Delta_{\min}\cdot\phi_{j_0}(\eta_t\hat{L}_t;\{e_{j_0},e_{j_1}\})
        \tag*{(by \eqref{eq:action_set_relation})}\\
        &\geq 
        \frac{\Delta_{\min}}{2},\numberthis\label{eq:lower_bound_Dtc_1}
    \end{align*}
    where the last inequality follows from $\hat{L}_{t,j_1}\geq\hat{L}_{t,j_0}$
    and the fact that, conditioned on $\hat{L}_t$, the perturbations of $j_0$ and $j_1$
    are independent and identically distributed with a continuous distribution.

    Next, we consider the subevents $B_{t,F}^+$ and $B_{t,D}^+$.
    Both $B_{t,F}^+$ and $B_{t,D}^+$ require $\eta_t\underline{\hat{L}}_{t,i}>0$ for every $i$ with $a_i^*=0$. As above, this implies $\sele(-\eta_t\hat L_t;\nec)=a^*$, which also gives $\eta_t\underline{\hat{L}}_{t,i}\leq0$ for every $i$ with $a_i^*=1$.
    Thus, both subevents satisfy the sign conditions used to derive~\eqref{eq:lower_bound_to_MAB1} and~\eqref{eq:lower_bound_to_MAB2}, so these bounds apply directly.
    On $B_{t,F}^+$ and $B_{t,D}^+$ we obtain
    \begin{equation}\label{eq:lower_bound_Dtc_2_intermediate}
        \sum_{i:a_i^*=0} \Delta_{i,\min} w_{t,i}
        \geq
        \Delta_{\min} \sum_{i:a_i^*=0} \sP\qty[i=\argmax_{j:a^*_j=0}\qty(r_{t,j}-\eta_t\underline{\hat{L}}_{t,j})
        ,\,r_{t,i}-\eta_t\underline{\hat{L}}_{t,i}\geq r_{t,\blkomit{t,i}}
        \m \hat{L}_t],
    \end{equation}

    Here, the sum on the RHS equals the probability of selecting one of
    the $d-m$ suboptimal arms in an MAB setting with $d-m+1$ arms.
    The suboptimal arms $\{j:a_j^*=0\}$ have cumulative losses $\{\underline{\hat{L}}_{t,j}\}_{j:a_j^*=0}$, and the remaining arm has the smallest cumulative loss of $0$.

    The proofs of Lemmas~17 and~18 in \citet{chen2026further}, which respectively consider the Fr\'{e}chet and Pareto perturbations, also reduce the $m$-set semi-bandit problem to the sum of arm-selection probabilities on the RHS of~\eqref{eq:lower_bound_Dtc_2_intermediate}.
    The conditions imposed there on the shifted cumulative losses $\{\underline{\hat{L}}_{t,j}\}_{j:a_j^*=0}$ coincide with those defining $B_{t,F}^+$ and $B_{t,D}^+$, respectively.
    The remaining arm, which corresponds to $\blkomit{t,i}$ here, is assigned cumulative loss $0$ in both analyses.
    Therefore, the resulting lower bounds in \citet{chen2026further} apply directly here.

    Now we first provide the lower bound on $B_{t,F}^+$ for Fr\'{e}chet perturbations. 
    Lemma~17 of \citet{chen2026further} shows that, if $\{\underline{\hat{L}}_{t,j}\}_{j:a^*_j=0}$ satisfy $B_{t,F}^+$, 
    by letting $c_{s,2}(\Fdis)=\frac{m^{\alpha/(\alpha-1)}}{2^\alpha \qty(m^{\alpha/(\alpha-1)}+1)+1}\in\big[\frac{1}{2^{\alpha+1}+1},\frac{1}{2^\alpha}\big)$ we have
    \begin{equation}\label{eq:lower_bound_Ftc_2}
        \sum_{i:a_i^*=0} \Delta_{i,\min} w_{t,i}\geq
        c_{s,2}(\Fdis)\frac{\Delta_{\min}}{m^{\alpha/(\alpha-1)}}.
    \end{equation}

    Then, Lemma~18 of \citet{chen2026further} shows that, if $\{\underline{\hat{L}}_{t,j}\}_{j:a^*_j=0}$ satisfy the same condition as in $B_{t,D}^+$, we have
    \begin{equation}\label{eq:lower_bound_Dtc_2}
        \eqref{eq:lower_bound_Dtc_2_intermediate}
        \geq
        \Delta_{\min}\frac{e^{-1}}{2}\qty(1-\exp\qty(- 2\qty(1 - F\qty(2^{\frac{1}{\alpha}} + m^{1/(\alpha-1)}))))
    \end{equation}
    Then, by combining \eqref{eq:lower_bound_Dtc_1} and \eqref{eq:lower_bound_Dtc_2}, we have
    \begin{align*}
        \sum_{i:a_i^*=0} \Delta_{i,\min} w_{t,i}
        &\geq
        \Delta_{\min}\min\qty{\frac{1}{2},\frac{e^{-1}}{2}\qty(1 - \exp\qty(-2\qty(1 - F\qty(2^{\frac{1}{\alpha}} + m^{1/(\alpha-1)}))))}\\
        &\geq
        c_{s,2}(\Pdis)\frac{\Delta_{\min}}{m^{\alpha/(\alpha-1)}},
    \end{align*} 
    Here, the last inequality also follows from Lemma~18 of \citet{chen2026further}, and $c_{s,2}(\Pdis)$ is given as
    \begin{equation*}
        c_{s,2}(\Pdis)=\frac{m^{\alpha/(\alpha-1)}}{2^{\alpha-1}e(2+m^{\alpha/(\alpha-1)})+2e}\in\left[\frac{1}{(3\cdot 2^{\alpha-1}+2)e},\frac{1}{2^{\alpha-1}e}\right),
    \end{equation*}
    which with \eqref{eq:lower_bound_Dtc_1} and \eqref{eq:lower_bound_Ftc_2}
    concludes the proof.
\end{proof}

\subsection{Regret for the Optimal Action}\label{subsec:regret_optimal_action}

To apply the self-bounding technique, we need to express the regret associated with the
optimal action in terms of the statistics of the suboptimal base-arms, i.e.,
those with $a^*_i=0$. Compared with the single optimal arm in the MAB
setting~\citep{pmlr-v201-honda23a,pmlr-v247-lee24a}, this step is
substantially more challenging, as the optimal action consists of multiple interdependent base-arms. 
This difficulty already arises in the $m$-set setting~\citep{chen2026further}, and becomes more pronounced for general matroids, where the dependencies among base-arms are governed by the
matroid structure rather than a cardinality constraint. 
Our approach is to first bound the stability term associated with the base-arms of the optimal action. 
Then, we relate this bound to the suboptimal base-arms, which completes the argument.

\begin{lemma}\label{lem:optimal_action_add}
    Define $\delta_{\armset}=\eta_t\max_{i:a^*_i=1}\hat{\ell}_{t,i}\bm{1}_{\armset}$ for any set $\armset\subset[d]$. 
    Then, for any $\lambda\in\mathbb{R}^d$,  $\armset\subsetneq \armset^k\subset \qty{i:a^*_i=1}$ where 
    $\armset^k\setminus \armset=\{k\}$, on $F_t$ (resp.~$D_t$) for $\dis=\Fdis$ (resp.~$\dis=\Pdis$), it holds that
    \begin{equation*}
        \sum_{i \in \armset}\phi_i(\eta_t\hat{L}_t)-\sum_{i \in \armset}\phi_i\qty(\eta_t\hat{L}_t+\delta_\armset)
        \leq
        \sum_{i \in \armset^k}\phi_i(\eta_t\hat{L}_t)-\sum_{i \in \armset^k}\phi_i\qty(\eta_t\hat{L}_t+\delta_{\armset^k})+
        \sum_{j:a_j^*=0}\frac{c_{s,3}(\dis)\eta_t}{(\eta_t\underline{\hat{L}}_{t,j})^\alpha}\max_{i:a^*_i=1}\hat{\ell}_{t,i},
    \end{equation*}
    where
    \begin{equation*}
        c_{s,3}(\dis)=
        \begin{cases}
            4\alpha/e^2, & \dis=\Fdis,\\
            \alpha, & \dis=\Pdis.
        \end{cases}
    \end{equation*}
\end{lemma}

\begin{proof}
    Firstly, by rearranging the terms we obtain
\begin{multline*}
    \qty(\sum_{i \in \armset^k}\phi_i(\eta_t\hat{L}_t)-\sum_{i \in \armset^k}\phi_i(\eta_t\hat{L}_t+\delta_{\armset^k}))-\qty(\sum_{i \in \armset}\phi_i(\eta_t\hat{L}_t)-\sum_{i \in \armset}\phi_i(\eta_t\hat{L}_t+\delta_\armset))\\
    =\qty(\phi_k(\eta_t\hat{L}_t)-\phi_k(\eta_t\hat{L}_t+\delta_{\armset^k}))-
    \sum_{i \in \armset}\qty(\phi_i(\eta_t\hat{L}_t+\delta_{\armset^k})-\phi_i(\eta_t\hat{L}_t+\delta_\armset)).
\end{multline*}
Then, we have
\begin{align*}
    &\phi_k(\eta_t\hat{L}_t)-\phi_k(\eta_t\hat{L}_t+\delta_{\armset^k})\\
    &=
    \sP\qty[\sele_k(r-\eta_t\hat{L}_t)=1]-
    \sP\qty[\sele_k(r-(\eta_t\hat{L}_t+\delta_{\armset^k}))=1
    ]\\
    &=
    \sP\qty[\sele_k(r-\eta_t\hat{L}_t)=1]-
    \sP\qty[\sele_k(r-\eta_t\hat{L}_t)=1,\, \sele_k(r-(\eta_t\hat{L}_t+\delta_{\armset^k}))=1
    ]\tag{by Lemma~\ref{lem:greedy_remove}}\\
    &=
    \sP\qty[\sele_k(r-\eta_t\hat{L}_t)=1,\, \sele_k(r-(\eta_t\hat{L}_t+\delta_{\armset^k}))=0
    ].
\end{align*}
Similarly, we have
\begin{equation*}
    \phi_i(\eta_t\hat{L}_t+\delta_{\armset^k})-\phi_i(\eta_t\hat{L}_t+\delta_\armset)
    =
    \sP\qty[
    \sele_i(r-(\eta_t\hat{L}_t+\delta_{\armset^k}))=1,\,
    \sele_i(r-(\eta_t\hat{L}_t+\delta_\armset))=0].
\end{equation*}
Then we have
\begin{align*}
    \lefteqn{\qty(\phi_k(\eta_t\hat{L}_t)-\phi_k(\eta_t\hat{L}_t+\delta_{\armset^k}))-
    \sum_{i \in \armset}\qty(\phi_i(\eta_t\hat{L}_t+\delta_{\armset^k} )-\phi_i(\eta_t\hat{L}_t+\delta_\armset))}\\
    &=
    \E\bigg[\ind\qty[\sele_k(r-\eta_t\hat{L}_t)=1,\, \sele_k(r-(\eta_t\hat{L}_t+\delta_{\armset^k}))=0
    ]\\
    &\hspace{3cm}-
    \sum_{i\in \armset}\ind\qty[\sele_i(r-(\eta_t\hat{L}_t+\delta_{\armset^k}))=1,\,
    \sele_i(r-(\eta_t\hat{L}_t+\delta_\armset))=0]\bigg].\numberthis\label{eq:expect_diff}
\end{align*}
Now we show that the event
\begin{equation}\label{eq:diff_event}
    \qty{\sele_i(r-(\eta_t\hat{L}_t+\delta_{\armset^k}))=1,\,
    \sele_i(r-(\eta_t\hat{L}_t+\delta_\armset))=0}
\end{equation}
can occur for at most one $i\in \armset$.
Note that $r-(\eta_t\hat{L}_t+\delta_{\armset^k})$ and $r-(\eta_t\hat{L}_t+\delta_\armset)$ are different only for one element, i.e., the $k$-th element.
Define
\begin{equation*}
    \armset_{\preceq k}=\qty{j:\sigma_j(r-(\eta_t\hat{L}_t+\delta_{\armset^k}))\leq\sigma_k(r-(\eta_t\hat{L}_t+\delta_{\armset^k}))},
\end{equation*}
which consists of $k$ itself and the base-arms that precede $k$ when the base-arms
are sorted in descending order of $r-(\eta_t\hat L_t+\delta_{\armset^k})$.
Then, one can see that the first $\left\lvert \armset_{\preceq k}\right\rvert$ base-arms under the permutations induced by $r-(\eta_t\hat{L}_t+\delta_{\armset^k})$ and $r-(\eta_t\hat{L}_t+\delta_\armset)$ form the same set, namely $\armset_{\preceq k}$, with only $k$ being repositioned within the permutation.
Therefore, by Lemma~\ref{lem:greedy_num}, the number of base-arms that can be selected from $\armset_{\preceq k}$ is the same under both permutations, that is,
\begin{equation}\label{eq:greedy_head_number_equal}
    \left\lvert \armset_{\preceq k}\cap\supp{\sele(r-(\eta_t\hat{L}_t+\delta_\armset))}\right\rvert
    =
    \left\lvert \armset_{\preceq k}\cap\supp{\sele(r-(\eta_t\hat{L}_t+\delta_{\armset^k}))}\right\rvert.
\end{equation}
Furthermore, by Lemma~\ref{lem:greedy_remove}, the base-arms except $k$ that are selected according to the permutation induced by $r-(\eta_t\hat{L}_t+\delta_\armset)$ remain selected if the permutation is changed to that induced by $r-(\eta_t\hat{L}_t+\delta_{\armset^k})$, since the set of base-arms preceding each of them can only shrink (with $k$ being the only base-arm that may be removed).
Therefore, we have
\begin{equation}\label{eq:greedy_only_k}
    \qty(\armset_{\preceq k}\cap\supp{\sele(r-(\eta_t\hat{L}_t+\delta_\armset))})
    \setminus
    \qty(\armset_{\preceq k}\cap\supp{\sele(r-(\eta_t\hat{L}_t+\delta_{\armset^k}))})
    \subset \{k\},
\end{equation}
which means that the base-arms that are selected from $\armset_{\preceq k}$ under the permutation sorted by $r-(\eta_t\hat{L}_t+\delta_\armset)$ in descending order but not selected under the permutation sorted by $r-(\eta_t\hat{L}_t+\delta_{\armset^k})$ in descending order can only be $k$. 

For every base-arm $j'\in [d]\setminus E_{\preceq k}$, i.e., every base-arm appearing after the first $|\armset_{\preceq k}|$ positions in both permutations sorted by $r-(\eta_t\hat{L}_t+\delta_\armset)$ and $r-(\eta_t\hat{L}_t+\delta_{\armset^k})$, the set of base-arms preceding $j'$ is identical under both permutations.
Therefore, by Lemma~\ref{lem:greedy_equiv}, their selection status remains unchanged under the two permutations, and thus the event \eqref{eq:diff_event} cannot occur for any of them.
From this fact with \eqref{eq:greedy_head_number_equal} and \eqref{eq:greedy_only_k}, the event \eqref{eq:diff_event} can occur for at most one $i\in \armset$.

When $i$ satisfying \eqref{eq:diff_event} exists, which is denoted by $i_k\in \armset$, it must satisfy
\begin{align}
&\sele_{i_k}(r-\eta_t\hat{L}_t-\delta_{\armset^k})=1,\label{opt_sto3}\\
&\sele_{i_k}(r-\eta_t\hat{L}_t-\delta_\armset)=0.\label{opt_sto4}\\
&\sele_k(r-\eta_t\hat{L}_t-\delta_{\armset^k})=0,\label{opt_sto5}\\
&\sele_k(r-\eta_t\hat{L}_t-\delta_\armset)=1.\label{opt_sto6}
\end{align}

Accordingly, we can see that if
\begin{equation}\label{target_one}
\ind\qty[\sele_k(r-\eta_t\hat{L}_t)=1,\,\sele_k(r-\eta_t\hat{L}_t-\delta_{\armset^k})=0]
-\sum_{i\in \armset}\ind\qty[\sele_i(r-\eta_t\hat{L}_t-\delta_{\armset^k})=1,\,\sele_i(r-\eta_t\hat{L}_t-\delta_\armset)=0]
\end{equation}
is negative, it must be $-1$, which can occur only when there exists $i_k\in \armset$ satisfying
\begin{align}
&\sele_k(r-\eta_t\hat{L}_t)=0 \;\mbox{ or }\;\sele_k(r-\eta_t\hat{L}_t-\delta_{\armset^k})=1,\label{opt_sto2}
\end{align}
and \eqref{opt_sto3}--\eqref{opt_sto6}.
Here, \eqref{opt_sto5} and \eqref{opt_sto2} together imply
$\sele_k(r-\eta_t\hat{L}_t)=0$.
Therefore, by $a_k^*=1$ there must exist
$j$ such that $a_j^*=0$ and
\begin{align}
\sele_j(r-\eta_t\hat{L}_t)=1\;\mbox{ and }\;\sigma_j(r-\eta_t\hat{L}_t)<\sigma_k(r-\eta_t\hat{L}_t),
\label{opt_km}
\end{align}
which trivially implies
\begin{align}
r_{t,j}
\geq
\nu+\eta_t(\hat{L}_{t,j}-\hat{L}_{t,\blkomit{t,j}})=\nu+\eta_t\underline{\hat{L}}_{t,j}.
\label{opt_rj}
\end{align}
Since $\sigma_j(r-\eta_t\hat{L}_t)\geq \sigma_j(r-\eta_t\hat{L}_t-\delta_{\armset^k})$ by $j\notin \armset^k$ from $a_j^*=0$,
we also see from \eqref{opt_sto3} and \eqref{opt_km} that
\begin{align}
\sele_j(r-\eta_t\hat{L}_t-\delta_{\armset^k})=1.
\label{opt_km2}
\end{align}
Since the two orderings differ only in the position of $k$, the selection status of $i_k$ can change only if $k$ crosses $i_k$ in the ordering.
Therefore, by \eqref{opt_sto3} and \eqref{opt_sto5}
and
by \eqref{opt_sto4} and \eqref{opt_sto6},
we have
\begin{align}\label{eq:opt_relative_position}
(r-\eta_t\hat{L}_t-\delta_{\armset^k})_{i_k}>(r-\eta_t\hat{L}_t-\delta_{\armset^k})_{k}
\;\mbox{ and }\;
(r-\eta_t\hat{L}_t-\delta_{\armset})_{i_k}<(r-\eta_t\hat{L}_t-\delta_{\armset})_{k},
\end{align}
respectively.
This is equivalent to
$0<(r-\eta_t\hat{L}_t)_{i_k}-(r-\eta_t\hat{L}_t)_{k}< \eta_t\max_{i:a^*_i=1}\hat{\ell}_{t,i}$
or equivalently
\begin{align}
r_k\in
\left[
r_{i_k}-\eta_t\hat{L}_{t,i_k}+\eta_t\hat{L}_{t,k}-\eta_t\max_{i:a^*_i=1}\hat{\ell}_{t,i}
,\;
r_{i_k}-\eta_t\hat{L}_{t,i_k}+\eta_t\hat{L}_{t,k}
\right].
\label{opt_sto7}
\end{align}

Finally, we show that the choice of the unique base-arm $i_k$ satisfying \eqref{opt_sto3}--\eqref{opt_sto6} is actually independent of $r_{t,j}$ and $r_{t,k}$ under the condition \eqref{opt_km2}.
Fix the permutation of $[d]\setminus\{j,k\}$ induced by $r-(\eta_t\hat{L}_t+\delta_{\armset^k})$.
We consider whether $k$ is selected under the action set $\nec_j$ when $k$ is inserted at an arbitrary position in this permutation.
By Lemma~\ref{lem:greedy_remove}, the selection of $k$ is monotone with respect to its insertion position.
To be specific, if $k$ is selected at some position, then it remains selected at any earlier position, whereas if $k$ is not selected at some position, then it remains unselected at any later position.
Therefore, there is at most one base-arm $i'_k\in[d]\setminus\{j,k\}$ such that $k$ is selected when inserted immediately before $i'_k$ but not selected when inserted immediately after $i'_k$.

On the other hand, by \eqref{opt_sto3}, \eqref{opt_sto4} and \eqref{eq:opt_relative_position}, we can see that the selection of $i_k$ changes depending on whether $k$ is positioned before or after $i_k$.
Furthermore, similarly to the analysis around \eqref{eq:greedy_head_number_equal} and \eqref{eq:greedy_only_k}, we can see that
 the selection status of every base-arm in $[d]\setminus\{j,k,i'_k\}$ remains unchanged regardless of whether $k$ is positioned before or after $i'_k$.
Therefore, we have $i_k\notin[d]\setminus\{j,k,i'_k\}$ and thus
\begin{equation}\label{opt_sto9}
    i_k=i'_k.
\end{equation}
This implies that the choice of $i_k$ is determined only by the permutation of $[d]\setminus\{j,k\}$, which is independent of $r_{t,j}$ and $r_{t,k}$.

Note that \eqref{opt_sto9} can hold for at most one $i_k\in \armset\setminus\{j\}$, and it is determined only by
$(r-(\eta_t\hat{L}_t+\delta_{\armset^k}))_{[d]\setminus \{j,k\}}$.
Therefore, from the above discussions we have
\begin{align}
-\eqref{target_one}
&\le
\sum_{j: a_j^*=0}
\ind\qty[\eqref{opt_rj},\,\mbox{$i_k$ s.t. $\eqref{opt_sto9}$ exists}, \eqref{opt_sto7}]
\n
\end{align}
and by taking the expectation with respect to $r_t$ we have
\begin{equation}\label{eq:target_one_as_prob}
\E[-\eqref{target_one}]
\le
\sum_{j: a_j^*=0}
\sP\left[\eqref{opt_rj}\right]\cdot
\E\Big[
\ind\qty[\mbox{$i_k$ s.t. $\eqref{opt_sto9}$ exists}]
\;\sP\qty[\eqref{opt_sto7}|\{r_{t,i}\}_{i\in [d]\setminus\{j,k\}}]
\Big].
\end{equation}
Here, we have
\begin{equation}\label{eq:first_pr}
    \sP\qty[\eqref{opt_rj}]=\sP\qty[r_{t,j}\geq \nu+\eta_t\underline{\hat{L}}_{t,j}]\leq
    \frac{1}{(\eta_t\underline{\hat{L}}_{t,j})^\alpha},
\end{equation}
since
\begin{equation*}
    \sP\qty[r_{t,j}\geq \eta_t\underline{\hat{L}}_{t,j}]=
    1-F\qty(\eta_t\underline{\hat{L}}_{t,j})=1-\exp\qty{-\frac{1}{(\eta_t\underline{\hat{L}}_{t,j})^\alpha}}\leq\frac{1}{(\eta_t\underline{\hat{L}}_{t,j})^\alpha}
\end{equation*}
for $\dis=\Fdis$, and
\begin{equation*}
    \sP\qty[r_{t,j}\geq 1+\eta_t\underline{\hat{L}}_{t,j}]=
    1-F\qty(1+\eta_t\underline{\hat{L}}_{t,j})=\frac{1}{(1+\eta_t\underline{\hat{L}}_{t,j})^\alpha}<\frac{1}{(\eta_t\underline{\hat{L}}_{t,j})^\alpha}
\end{equation*}
for $\dis=\Pdis$.
Then, by letting $f_{k}(x)$ be the pdf of $r_{k}$,
we have
\begin{align*}
\lefteqn{
\sP\qty[\eqref{opt_sto7}|\{r_{t,i}\}_{i\in [d]\setminus\{j,k\}}]
}\\
&=
\sP\left[
\left.r_k\in
\left[
r_{i_k}-\eta_t\hat{L}_{t,i_k}+\eta_t\hat{L}_{t,k}-\eta_t\max_{i:a^*_i=1}\hat{\ell}_{t,i}
,\;
r_{i_k}-\eta_t\hat{L}_{t,i_k}+\eta_t\hat{L}_{t,k}
\right]
\right|
\{r_{t,i}\}_{i\in [d]\setminus\{j,k\}}
\right]
\\
&\le
\eta_t\max_{i:a^*_i=1}\hat{\ell}_{t,i}
\max_{x\geq \nu}f_k(x)
\leq
\begin{cases}
    \frac{4\alpha}{e^2}\eta_t\max_{i:a^*_i=1}\hat{\ell}_{t,i}, & \dis=\Fdis,\\
    \alpha\eta_t\max_{i:a^*_i=1}\hat{\ell}_{t,i}, & \dis=\Pdis.
\end{cases}
\numberthis\label{eq:second_pr}
\end{align*}
where $f_k(x)=f(x)$ since the condition is independent of $r_k$.

The proof is completed by substituting \eqref{eq:first_pr} and \eqref{eq:second_pr} into \eqref{eq:target_one_as_prob} and combining the resulting expression with \eqref{eq:expect_diff}.
\end{proof}

By applying Lemma~\ref{lem:optimal_action_add} repeatedly, we can bound the stability term associated with base-arms of the optimal action $a^*$ as follows.
The proof follows the same structure as that of Lemma~20 of \citet{chen2026further}, and we provide it here for completeness.
\begin{lemma}\label{lem:optimal_action_combine}
    On $F_t$ or $D_t$ we have
    \begin{multline*}
        \E\qty[\sum_{i:a_i^*=1}\hat{\ell}_{t,i}\qty(\phi_i(\eta_t\hat{L}_t)-\phi_i(\eta_t(\hat{L}_t+\hat{\ell}_t)))\m\hat{L}_t]
        \\\leq
        \E\qty[\max_{i:a_i^*=1}\hat{\ell}_{t,i}\sum_{i:a_i^*=1}\qty(\phi_i(\eta_t\hat{L}_t)-\phi_i(\eta_t(\hat{L}_t+\max_{i:a_i^*=1}\hat{\ell}_{t,i}\bm{1}_{A^*})))\m\hat{L}_t]
        +\sum_{j:a_j^*=0}\frac{c_{s,4}(\dis)}{\underline{\hat{L}}_{t,j}},
    \end{multline*}
    where 
    \begin{equation*}
        c_{s,4}(\dis)=
        \begin{cases}
            8\alpha/e, & \dis=\Fdis,\\
            121\alpha/14, & \dis=\Pdis.
        \end{cases}
    \end{equation*}
\end{lemma}

\begin{proof}
    Firstly, by letting $\armset^{+}=\qty{i\in[d]:a_i^*=1,\,\phi_i(\eta_t\hat{L}_t)-\phi_i(\eta_t(\hat{L}_t+\hat{\ell}_t))\geq 0}$, we have
    \begin{align*}
        \E\qty[\sum_{i:a_i^*=1}\hat{\ell}_{t,i}\qty(\phi_i(\eta_t\hat{L}_t)-\phi_i(\eta_t(\hat{L}_t+\hat{\ell}_t)))\m\hat{L}_t]
        &\leq
        \E\qty[\sum_{i\in \armset^{+}}\hat{\ell}_{t,i}\qty(\phi_i(\eta_t\hat{L}_t)-\phi_i(\eta_t(\hat{L}_t+\hat{\ell}_t)))\m\hat{L}_t]\\
        &\hspace{-3.5cm}\leq
        \E\qty[\qty(\max_{i:a_i^*=1}\hat{\ell}_{t,i})\sum_{i\in \armset^{+}}\qty(\phi_i(\eta_t\hat{L}_t)-\phi_i(\eta_t(\hat{L}_t+\hat{\ell}_t)))\m\hat{L}_t]\\
        &\hspace{-3.5cm}\leq
        \E\qty[\qty(\max_{i:a_i^*=1}\hat{\ell}_{t,i})\sum_{i\in \armset^{+}}\qty(\phi_i(\eta_t\hat{L}_t)-\phi_i\qty(\eta_t(\hat{L}_t+\max_{i:a_i^*=1}\hat{\ell}_{t,i}\bm{1}_{\armset^{+}})))\m\hat{L}_t],
        \numberthis\label{eq:optimal_action_intermediate_positive}
    \end{align*}
    where the last inequality follows from Lemma~\ref{lem:phi_monotone}, since 
    $\hat{\ell}_{t,i}\leq\max_{i:a_i^*=1}\hat{\ell}_{t,i}$ for $i\in \armset^{+}$, and 
    $\hat{\ell}_{t,i}\geq 0$ for $i\in[d]\setminus \armset^{+}$.
    Then, by repeatedly applying Lemma~\ref{lem:optimal_action_add} to \eqref{eq:optimal_action_intermediate_positive} for at most $m$ steps, we obtain
    \begin{multline}\label{eq:optimal_action_intermediate_positive_2}
        \eqref{eq:optimal_action_intermediate_positive}
        \leq
        \E\qty[\max_{i:a_i^*=1}\hat{\ell}_{t,i}\sum_{i:a_i^*=1}\qty(\phi_i(\eta_t\hat{L}_t)-\phi_i(\eta_t(\hat{L}_t+\max_{i:a_i^*=1}\hat{\ell}_{t,i}\bm{1}_{A^*})))\m\hat{L}_t]\\
        +
        \begin{cases*}
            \E\qty[\sum_{j:a_j^*=0}\frac{4\alpha m\eta_t}{e^2(\eta_t\underline{\hat{L}}_{t,j})^\alpha}\qty(\max_{i:a^*_i=1}\hat{\ell}_{t,i})^2\m\hat{L}_t], & $\dis=\Fdis$,\\
            \E\qty[\sum_{j:a_j^*=0}\frac{\alpha m\eta_t}{(\eta_t\underline{\hat{L}}_{t,j})^\alpha}\qty(\max_{i:a^*_i=1}\hat{\ell}_{t,i})^2\m\hat{L}_t], & $\dis=\Pdis$.
        \end{cases*}
    \end{multline}

    We first bound the conditional second moment of $\max_{i:a_i^*=1}\hat{\ell}_{t,i}$.
    Since $0\leq\ell_{t,i}\leq1$ and $\widehat{w_{t,i}^{-1}}\geq1$ for every selected base-arm, we have
    \begin{equation*}
        \qty(\max_{i:a_i^*=1}\hat{\ell}_{t,i})^2
        \leq 1+\sum_{i:a_i^*=1,\,a_{t,i}=1}
        \qty(\widehat{w_{t,i}^{-1}}^2-1).
    \end{equation*}
    Conditional on $\hat{L}_t$ and $a_{t,i}=1$, the GR estimate $\widehat{w_{t,i}^{-1}}$ follows a geometric distribution with parameter $w_{t,i}$, whose second moment is $(2-w_{t,i})/w_{t,i}^2$.
    Taking conditional expectations gives
    \begin{align*}
        \E\qty[\qty(\max_{i:a_i^*=1}\hat{\ell}_{t,i})^2\m\hat{L}_t]
        &\leq 1+\sum_{i:a_i^*=1}w_{t,i}
        \qty(\frac{2-w_{t,i}}{w_{t,i}^2}-1)\\
        &=1+\sum_{i:a_i^*=1}(1-w_{t,i})\qty(\frac{2}{w_{t,i}}+1)\\
        &\leq1+\sum_{i:a_i^*=1}(1-w_{t,i})\qty(\frac{2}{\wlow}+1)
        \tag{by~Lemma~\ref{lem:lower_bound_Dt}}\\
        &=
        1+\qty(\frac{2}{\wlow}+1)\sum_{i:a_i^*=0}w_{t,i},
        \numberthis\label{eq:optimal_action_gr_second_moment_W}
    \end{align*}
    where the last equality follows from $\sum_{i=1}^d w_{t,i}=m$.
    It remains to bound $\sum_{i:a_i^*=0}w_{t,i}$ under each perturbation distribution.
    On $F_t$ or $D_t$, the necessary condition~\eqref{eq:r_start} gives
    \begin{equation}\label{eq:neccessary_condition_imply}
        \sum_{i:a_i^*=0}w_{t,i}
        \leq
        \sum_{j:a_j^*=0}\qty(1-F\qty(\nu+\eta_t\underline{\hat{L}}_{t,j})).
    \end{equation}

    For $\dis=\Pdis$, condition~\eqref{eq:key_property} on $D_t$ gives $\eta_t\underline{\hat{L}}_{t,j}\geq m^{1/(\alpha-1)}$ for every $j$ with $a_j^*=0$.
    Since $2^{1/\alpha}>1$, the ratio below decreases as $\eta_t\underline{\hat{L}}_{t,j}$ increases, so
    \begin{equation}\label{eq:optimal_action_gr_second_moment_w}
        \frac{2^{1/\alpha}+\eta_t\underline{\hat{L}}_{t,j}}{1+\eta_t\underline{\hat{L}}_{t,j}}
        \leq\frac{2^{1/\alpha}+m^{1/(\alpha-1)}}{1+m^{1/(\alpha-1)}}.
    \end{equation}
    Combining~\eqref{eq:neccessary_condition_imply}, \eqref{eq:optimal_action_gr_second_moment_w}, and the definition of $D_t$ yields
    \begin{align*}
        \sum_{i:a_i^*=0}w_{t,i}
        &\leq\sum_{j:a_j^*=0}\frac{1}{(1+\eta_t\underline{\hat{L}}_{t,j})^\alpha}
        \tag{by~\eqref{eq:neccessary_condition_imply}}\\
        &\leq\qty(\frac{2^{1/\alpha}+m^{1/(\alpha-1)}}{1+m^{1/(\alpha-1)}})^\alpha
        \sum_{j:a_j^*=0}\frac{1}{(2^{1/\alpha}+\eta_t\underline{\hat{L}}_{t,j})^\alpha}
        \tag{by~\eqref{eq:optimal_action_gr_second_moment_w}}\\
        &\leq
        \qty(\frac{2^{1/\alpha}+m^{1/(\alpha-1)}}{1+m^{1/(\alpha-1)}})^\alpha
        \frac{1}{(2^{\frac{1}{\alpha}}+m^{1/(\alpha-1)})^\alpha}
        \\
        &=\frac{1}{(1+m^{1/(\alpha-1)})^\alpha}<\frac{1}{2}.
    \end{align*}
    Since $\sum_{i:a_i^*=0}w_{t,i}<1/2$, and $w^*(\Pdis)\geq 0.14$, \eqref{eq:optimal_action_gr_second_moment_W} yields $\E[(\max_{i:a_i^*=1}\hat{\ell}_{t,i})^2|\hat{L}_t]\leq\frac{121}{14}$.

    For $\dis=\Fdis$, on $F_t$, \eqref{eq:neccessary_condition_imply} gives
    \begin{align*}
        \sum_{i:a_i^*=0}w_{t,i}
        &\leq\sum_{j:a_j^*=0}\qty(1-e^{-1/(\eta_t\underline{\hat{L}}_{t,j})^\alpha})\\
        &\leq\sum_{j:a_j^*=0}\frac{1}{(\eta_t\underline{\hat{L}}_{t,j})^\alpha}
        \leq m^{-\frac{\alpha}{\alpha-1}},
    \end{align*}
    where the last inequality follows from the definition of $F_t$.
    For $m\geq2$, this gives $\sum_{i:a_i^*=0}w_{t,i}<1/2$.
    Then, by \eqref{eq:optimal_action_gr_second_moment_W} and $w^*(\Fdis)\geq 1/e$, we have
    \begin{equation*}
        \E\qty[\qty(\max_{i:a_i^*=1}\hat{\ell}_{t,i})^2\m\hat{L}_t]\leq \frac{2e+3}{2}.
    \end{equation*}
    When $m=1$, let $j_1$ denote the unique optimal base-arm.
    Applying~\eqref{eq:square_expectation} and the bound $w_{t,j_1}\geq1/e$ from Lemma~\ref{lem:lower_bound_Dt}, we obtain
    \begin{equation*}
        \E\qty[\qty(\max_{j:a_j^*=1}\hat{\ell}_{t,j})^2\m\hat{L}_t]
        =\E\qty[\hat{\ell}_{t,j_1}^2\m\hat{L}_t]
        \leq\frac{2}{w^*(\Fdis)}\leq 2e.
    \end{equation*}
    Combining the bounds for both distributions, we obtain
    \begin{equation}\label{eq:optimal_action_gr_second_moment_bounds}
        \E\qty[\qty(\max_{i:a_i^*=1}\hat{\ell}_{t,i})^2\m\hat{L}_t]
        \leq
        \begin{cases*}
            2e, & if $\dis=\Fdis$ on $F_t$,\\
            121/14, & if $\dis=\Pdis$ on $D_t$.
        \end{cases*}
    \end{equation}
    By using the result in~\eqref{eq:optimal_action_gr_second_moment_bounds}, we bound the second term of~\eqref{eq:optimal_action_intermediate_positive_2} as follows.
    \begin{align*}
        \E\qty[\sum_{j:a_j^*=0}\frac{4\alpha m\eta_t}{e^2(\eta_t\underline{\hat{L}}_{t,j})^\alpha}\qty(\max_{i:a^*_i=1}\hat{\ell}_{t,i})^2\m\hat{L}_t]
        &\leq
        \sum_{j:a_j^*=0}\frac{8\alpha m\eta_t}{e(\eta_t\underline{\hat{L}}_{t,j})^\alpha}\tag{by \eqref{eq:optimal_action_gr_second_moment_bounds} on $F_t$}\\
        &\leq
        \sum_{j:a_j^*=0}\frac{8\alpha}{e\underline{\hat{L}}_{t,j}}.
    \end{align*}
    Here, the last inequality follows from \eqref{eq:key_property}, which implies that $(\eta_t\underline{\hat{L}}_{t,j})^{\alpha-1}\geq m$ for all $j$ such that $a_j^*=0$ on $F_t$.
        Similarly, when $\dis=\Pdis$, on $D_t$ we have
    \begin{equation*}
        \eqref{eq:optimal_action_intermediate_positive}
        \leq
        \E\qty[\max_{i:a_i^*=1}\hat{\ell}_{t,i}\sum_{i:a_i^*=1}\qty(\phi_i(\eta_t\hat{L}_t)-\phi_i(\eta_t(\hat{L}_t+\max_{i:a_i^*=1}\hat{\ell}_{t,i}\bm{1}_{A^*})))\m\hat{L}_t]
        +\E\qty[\sum_{j:a_j^*=0}\frac{\alpha m\eta_t}{(\eta_t\underline{\hat{L}}_{t,j})^\alpha}\qty(\max_{i:a^*_i=1}\hat{\ell}_{t,i})^2\m\hat{L}_t].
    \end{equation*}
    Here, 
    \begin{equation*}
        \E\qty[\sum_{j:a_j^*=0}\frac{\alpha m\eta_t}{(\eta_t\underline{\hat{L}}_{t,j})^\alpha}\qty(\max_{i:a^*_i=1}\hat{\ell}_{t,i})^2\m\hat{L}_t]\leq \sum_{j:a_j^*=0}\frac{121\alpha}{14\underline{\hat{L}}_{t,j}},
    \end{equation*}
    which completes the proof.
\end{proof}

The following lemma extends the findings in \citet{pmlr-v201-honda23a}, providing the bound of $\E[ \ind[\hat{\ell}_{t,i} > \zeta/\eta_t  ]\hat{\ell}_{t,i} | \hat{L}_t ]$ under Fr\'{e}chet and Pareto distributions for $m$-set semi-bandits. 
In addition, we provide a bound for $\E[ \ind[\hat{\ell}_{t,j} > \zeta/\eta_t  ]\hat{\ell}_{t,i} | \hat{L}_t ]$ for $i\neq j$, which follows from an analysis. 
Both results will be used in the proof of Lemma~\ref{lem:optimal_action_bound}.

\begin{lemma}[Partial results of Lemma~21 in \citet{chen2026further}]\label{lem:optimal_action_bound_case2}
    On $F_t$ or $D_t$, for any $\hat{L}_t$, $\zeta \in (0,1)$ and distinct base-arms $i,j\in A^*$, it holds that
    \begin{equation*}
        \E\left[ \ind\qty[\hat{\ell}_{t,i} > \frac{\zeta}{\eta_t}  ]\hat{\ell}_{t,i} \m \hat{L}_t \right]\leq \frac{1}{1 - \wlow} (1 - \wlow)^{\tfrac{\zeta}{\eta_t}} \qty( \frac{\zeta}{\eta_t} + \frac{1}{\wlow} )
    \end{equation*}
    and
    \begin{equation*}
        \E\left[ \ind\qty[\hat{\ell}_{t,j} > \frac{\zeta}{\eta_t}  ]\hat{\ell}_{t,i} \m \hat{L}_t \right]\leq
        (1 - \wlow)^{\frac{\zeta}{\eta_t} -1},
    \end{equation*}
    where $\wlow$ is the lower bound of $\min_{i:a^*_i=1}w_{t,i}$ given in Lemma~\ref{lem:lower_bound_Dt}.
    Furthermore, when $\eta_t = cm^{\frac{1}{2}-\frac{1}{\alpha}}d^{\frac{1}{\alpha}-\frac{1}{2}}/\sqrt{t}$, we have
    \begin{equation*}
        \sum_{t=1}^{\infty}\frac{1}{1 - \wlow} (1 - \wlow)^{\tfrac{\zeta}{\eta_t}} \qty( \frac{\zeta}{\eta_t} + \frac{1}{\wlow} )
        \leq O\qty( c^2m^{1-\frac{2}{\alpha}}d^{\frac{2}{\alpha}-1}  ),
    \end{equation*}
    and
    \begin{equation*}
        \sum_{t=1}^{\infty}(1 - \wlow)^{\frac{\zeta}{\eta_t} -1}
        \leq O\qty( c^2m^{1-\frac{2}{\alpha}}d^{\frac{2}{\alpha}-1}  ).
    \end{equation*}
\end{lemma}

The following lemma provides an upper bound on the regret of the optimal action under Fr\'{e}chet or Pareto distributions with shape $\alpha>1$.

\begin{lemma}\label{lem:optimal_action_bound}
    Let $\zeta\in(0,1)$ and $\alpha>1$. On $F_t$ (resp. $D_t$) for $\dis=\Fdis$ (resp. $\dis=\Pdis$) we have
    \begin{align*}
        \lefteqn{\E\qty[\sum_{i:a_i^*=1}\hat{\ell}_{t,i}\qty(\phi_i(\eta_t\hat{L}_t)-\phi_i(\eta_t(\hat{L}_t+\hat{\ell}_t)))\m\hat{L}_t]\leq\qty(\frac{c_{s,5}(\dis)}{(1-\zeta)^{\alpha+1}}+c_{s,4}(\dis))\sum_{i:a_i^*=0}\frac{1}{\underline{\hat{L}}_{t,i}}}\\
        &\hspace{3.2cm}+\frac{m}{1 - \wlow} (1 - \wlow)^{\tfrac{\zeta}{\eta_t}} \qty( \frac{\zeta}{\eta_t} + \frac{1}{\wlow} )+
        m^2(1 - \wlow)^{\frac{\zeta}{\eta_t} -1},
    \end{align*}
    where
    \begin{equation*}
        c_{s,5}(\dis)=
        \begin{cases}
            2\alpha e, & \dis=\Fdis,\\
            121\alpha/14, & \dis=\Pdis,
        \end{cases}
    \end{equation*}
    and $c_{s,4}(\dis)$ and $\wlow$ are defined in Lemmas~\ref{lem:optimal_action_combine} and \ref{lem:optimal_action_bound_case2}, respectively.
\end{lemma}

\begin{proof}
    Recall \eqref{eq:key_property}, which shows that, on the event $F_t$ or $D_t$, we have 
    $\eta_t\underline{\hat{L}}_{t,i}\leq 0$ 
    for all $i$ with $a_i^*=1$ and 
    $\eta_t\underline{\hat{L}}_{t,i}\geq m^{1/(\alpha-1)}$ 
    for all $i$ with $a_i^*=0$. 
    We consider the cases 
    (a) $\max_{i:a^*_i=1}\widehat{w_{t,i}^{-1}}\leq\zeta/\eta_t$ 
    and 
    (b) $\max_{i:a^*_i=1}\widehat{w_{t,i}^{-1}}>\zeta/\eta_t$ separately.
    
    (a) Let us consider the first case, which implies 
    $\max_{i:a^*_i=1}\hat{\ell}_{t,i}\leq \max_{i:a^*_i=1}\widehat{w_{t,i}^{-1}}\leq\zeta/\eta_t$.
    Let $b_1,b_2,\dots,b_m$ be an arbitrary ordering of the set of the base-arms of the optimal action $a^*$.
    Then, denote 
    $\hat{L}\pn{k}_t=\hat{L}_t+\max_{i:a^*_i=1}\hat{\ell}_{t,i}\cdot\sum_{j=1}^k e_{b_j}$ 
    and 
    $\hat{L}\pn{0}_t=\hat{L}_t$. 
    Note that $\sum_{i=1}^d\phi_{i}(\lambda)=m$ for any $\lambda\in\mathbb{R}^d$. 
    By Lemma~\ref{lem:optimal_action_combine}, we have
    \begin{align*}
        \lefteqn{\E\qty[\ind\qty[\max_{i:a^*_i=1}\hat{\ell}_{t,i}\leq\zeta/\eta_t]\sum_{i:a_i^*=1}\hat{\ell}_{t,i}\qty(\phi_i(\eta_t\hat{L}_t)-\phi_i(\eta_t(\hat{L}_t+\hat{\ell}_t)))\m\hat{L}_t]}\\
        &\leq
        \E\qty[\ind\qty[\max_{i:a^*_i=1}\hat{\ell}_{t,i}\leq\zeta/\eta_t]\max_{i:a_i^*=1}\hat{\ell}_{t,i}\sum_{i:a_i^*=1}\qty(\phi_i(\eta_t\hat{L}_t)-\phi_i(\eta_t(\hat{L}_t+\max_{i:a_i^*=1}\hat{\ell}_{t,i}\sum_{i:a_i^*=1}e_i)))\m\hat{L}_t]+\sum_{i:a_i^*=0}\frac{c_{s,4}(\dis)}{\underline{\hat{L}}_{t,i}}
        \displaybreak[0]\\
        &=
        \E\qty[\ind\qty[\max_{i:a^*_i=1}\hat{\ell}_{t,i}\leq\zeta/\eta_t]\max_{i:a_i^*=1}\hat{\ell}_{t,i}\sum_{i:a_i^*=0}\qty(\phi_i(\eta_t(\hat{L}_t+\max_{i:a_i^*=1}\hat{\ell}_{t,i}\sum_{i:a_i^*=1}e_i))-\phi_i(\eta_t\hat{L}_t))\m\hat{L}_t]
        +
        \sum_{i:a_i^*=0}\frac{c_{s,4}(\dis)}{\underline{\hat{L}}_{t,i}}
        \displaybreak[0]\\
        &=
        \E\qty[\ind\qty[\max_{i:a^*_i=1}\hat{\ell}_{t,i}\leq\zeta/\eta_t]\max_{i:a_i^*=1}\hat{\ell}_{t,i}\sum_{i:a_i^*=0}\sum_{j=1}^m\qty(\phi_{i}(\eta_t\hat{L}\pn{j}_t)-\phi_{i}(\eta_t\hat{L}\pn{j-1}_t))\m\hat{L}_t]
        +\sum_{i:a_i^*=0}\frac{c_{s,4}(\dis)}{\underline{\hat{L}}_{t,i}}.\numberthis\label{eq:optimal_action_frechet_transfer}
    \end{align*}
    On $F_t$ or $D_t$, for any $i'$ with $a^*_{i'}=0$, denote
    \begin{equation*}
        \armset_{\preceq \blkomit{t,i'}}
        =
        \qty{i\in[d]:\sigma_i\leq\sigma_{\blkomit{t,i'}}}.
    \end{equation*}
    Then, by the definition of $\blkomit{t,i'}$, the base-arm $i'$ is not selected if it
    is processed immediately after $\armset_{\preceq \blkomit{t,i'}}$.
    Furthermore, on $F_t$ or $D_t$, since the selected base-arms must be the base-arms of $a^*$, the set of selected base-arms in $\armset_{\preceq \blkomit{t,i'}}$ is $A^*\cap \armset_{\preceq \blkomit{t,i'}}$.
    Since the base-arms that are not selected don't affect the selection of the following base-arms, we obtain that $i'$ is not selected if it is processed immediately after $A^*\cap \armset_{\preceq \blkomit{t,i'}}$.

    Note that for any $i'$ with $a^*_{i'}=0$ and $i\in A^*\cap \armset_{\preceq \blkomit{t,i'}}$, we have
    \begin{equation*}
        \hat{L}_{t,i'}-\hat{L}_{t,i}
        \geq
        \hat{L}_{t,i'}-\hat{L}_{t,\blkomit{t,i'}}
        =\underline{\hat{L}}_{t,i'}
        \geq m^{1/(\alpha-1)}/\eta_t.
    \end{equation*}
    Moreover, since 
    $\max_{i:a^*_i=1}\widehat{w_{t,i}^{-1}}\leq\zeta/\eta_t$, we have 
    $\max_{i:a^*_i=1}\hat{\ell}_{t,i}\leq \zeta/\eta_t$. Therefore, for any 
    $x\leq \max_{i:a^*_i=1}\hat{\ell}_{t,i}$ and $j\in[m]$ we have
    \begin{align*}
        (\hat{L}_t\pn{j-1}+xe_{b_j})_{i'}- (\hat{L}_t\pn{j-1}+xe_{b_j})_{i}
        &=
        \hat{L}_{t,i'}- (\hat{L}_t+\max_{i:a^*_i=1}\hat{\ell}_{t,i}\sum_{i\in[j-1]}e_{b_i}+xe_{b_j})_{i}\\
        &\geq
        \hat{L}_{t,i'}- (\hat{L}_t+\max_{i:a^*_i=1}\hat{\ell}_{t,i}\sum_{i\in[j]}e_{b_i})_{i}\\
        &\geq(\hat{L}_{t,i'}-\hat{L}_{t,i})-\max_{i:a^*_i=1}\hat{\ell}_{t,i}\\
        &\geq \qty(\eta_t\underline{\hat{L}}_{t,i'}-\zeta m^{1/(\alpha-1)})/\eta_t\\
        &\geq (1-\zeta)(\eta_t\underline{\hat{L}}_{t,i'})/\eta_t> 0.\numberthis\label{eq:zeta_bound_frechet}
    \end{align*}
    Here, note that, for any $i'$ with $a_{i'}^*=0$, $i$ can be chosen arbitrarily from $A^*\cap\armset_{\preceq \blkomit{t,i'}}$.
    Therefore, \eqref{eq:zeta_bound_frechet} implies that, for any $j\in[m]$ and any $x\leq \max_{i:a^*_i=1}\hat{\ell}_{t,i}$, every base-arm in $A^* \cap\armset_{\preceq \blkomit{t,i'}}$ precedes $i'$ under the permutation induced by $(\hat{L}_t\pn{j-1}+xe_{b_j})$.

    Let $\blk{\cdot}{\hat{L}_t\pn{j-1}+xe_{b_j}}$ denote the base-arm satisfying the definition of $\blkomit{t,\cdot}$ with respect to the permutation induced by $(\hat{L}_t\pn{j-1}+xe_{b_j})$.
    Then, $i'$ is selected when it is processed immediately after
    \begin{equation*}
        \armset_{\prec \blk{i'}{\hat{L}_t\pn{j-1}+xe_{b_j}}}
        =
        \qty{i\in[d]:\sigma_i(-(\hat{L}_t\pn{j-1}+xe_{b_j}))
        <
        \sigma_{\blk{i'}{\hat{L}_t\pn{j-1}+xe_{b_j}}}(-(\hat{L}_t\pn{j-1}+xe_{b_j}))}.
    \end{equation*}
    Recalling that $i'$ is not selected when it is processed immediately after $A^*\cap\armset_{\preceq \blkomit{t,i'}}$, Lemma~\ref{lem:greedy_remove} implies that $i'$ is also not selected when it is processed after any superset of $A^*\cap\armset_{\preceq \blkomit{t,i'}}$.
    Therefore, we have
    \begin{equation*}
        (\hat{L}_t\pn{j-1}+xe_{b_j})_{\blk{i'}{\hat{L}_t\pn{j-1}+xe_{b_j}}}
        \leq
        \max_{i\in A^*\cap\armset_{\preceq \blkomit{t,i'}}}(\hat{L}_t\pn{j-1}+xe_{b_j})_i,
    \end{equation*}
    which further implies that
    \begin{align*}
        \underline{\hat{L}_t\pn{j-1}+xe_{b_j}}_{i'}
        &=
        (\hat{L}_t\pn{j-1}+xe_{b_j})_{i'}-(\hat{L}_t\pn{j-1}+xe_{b_j})_{\blk{i'}{\hat{L}_t\pn{j-1}+xe_{b_j}}}\\
        &\geq
        (\hat{L}_t\pn{j-1}+xe_{b_j})_{i'}-\max_{i\in A^*\cap\armset_{\preceq \blkomit{t,i'}}}(\hat{L}_t\pn{j-1}+xe_{b_j})_i\\
        &\geq
        \min_{i\in A^*\cap\armset_{\preceq \blkomit{t,i'}}}\qty{(\hat{L}_t\pn{j-1}+xe_{b_j})_{i'}-(\hat{L}_t\pn{j-1}+xe_{b_j})_i}\\
        &\geq (1-\zeta)(\eta_t\underline{\hat{L}}_{t,i'})/\eta_t,
    \end{align*}
    where the last inequality follows from \eqref{eq:zeta_bound_frechet}.

    For either perturbation distribution, for any $x\leq \zeta/\eta_t$, $i$ with $a_i^*=0$, and any $z\geq \nu$ it holds that
    \begin{align*}
        f\qty(z+\eta_t\underline{\hat{L}_t\pn{j-1}+xe_{b_j}}_i)
        &\leq \frac{\alpha}{(z+\eta_t\underline{\hat{L}_t\pn{j-1}+xe_{b_j}}_i)^{\alpha+1}}.\numberthis\label{eq:transform_pdf}
    \end{align*}
    Now we bound $\frac{\mathrm{d}}{\mathrm{d} x}\phi_{i}(\eta_t(\hat{L}_t\pn{j-1}+xe_{b_j}))$ for $x\leq \zeta/\eta_t$. 
    By Lemma~\ref{lem:tool_relation}, for $\{i,b_j\}\in\supp{\sI}$ we have
    \begin{align*}
        \frac{\mathrm{d}}{\mathrm{d} x}\phi_{i}(\eta_t(\hat{L}_t\pn{j-1}+xe_{b_j}))
        &=\frac{\mathrm{d}}{\mathrm{d} x}\phi_{i}\qty(\eta_t(\hat{L}_t\pn{j-1}+xe_{b_j});\nec_{b_j})
        +
        \frac{\mathrm{d}}{\mathrm{d} x}\phi_{i\prec b_j}\qty(\eta_t(\hat{L}_t\pn{j-1}+xe_{b_j});\nec)\\
        &=
        \frac{\mathrm{d}}{\mathrm{d} x}\phi_{i\prec b_j}\qty(\eta_t(\hat{L}_t\pn{j-1}+xe_{b_j});\nec),
    \end{align*}
    where the last equality holds since $x$ only affects the $b_j$-th component of $\hat{L}_t\pn{j-1}+xe_{b_j}$.
    For $\{i,b_j\}\notin\supp{\sI}$, we directly have
    $\frac{\mathrm{d}}{\mathrm{d} x}\phi_{i}(\cdot)=\frac{\mathrm{d}}{\mathrm{d} x}\phi_{i\prec b_j}(\cdot)$.
    Recall the definition of $\phi_{i\prec b_j}(\cdot)$ given in \eqref{eq:def_phi_ij}, and 
    note that $(\hat{L}_t\pn{j-1}+xe_{b_j})_{i'}=\hat{L}_{t,i'}\pn{j-1}+\ind[i'=b_j]x$ holds for any $i'\in[d]$. 
    For any $i$ with $a_i^*=0$, let $b_{j'}$ be such that $a_{b_{j'}}^*=1$ and
    $b_{j'}=\blk{i}{\hat{L}_t\pn{j-1}+xe_{b_j}}$.
    Then we have
    \begin{align*}
        \lefteqn{\frac{\mathrm{d}}{\mathrm{d} x}\phi_{i\prec b_j}\qty(\eta_t(\hat{L}_t\pn{j-1}+xe_{b_j});\dis,\nec)}\\
        &=
        \frac{\mathrm{d}}{\mathrm{d} x}\int_{\nu-\eta_t(\hat{L}_{t,b_{j'}}\pn{j-1}+\ind[b_{j'}=b_j]x)}^\infty F\qty(z+\eta_t(\hat{L}_{t,b_j}\pn{j-1}+x))f\qty(z+\eta_t\hat{L}_{t,i}\pn{j-1})
        \\
        &\hspace{3cm}
        \sP_{\{r_k\}_{k\in[d]\setminus\{i,b_j\}} \sim \mathcal{D},\,r_i=z+\eta_t\hat{L}_{t,i}\pn{j-1}}\qty[\sele_i(r-\eta_t\hat{L}_t\pn{j-1};\nec)=1,\, \sele_i(r-\eta_t\hat{L}_t\pn{j-1};\nec_{b_j})=0]
        \dd z\\
        &=
        \eta_t\int_{\nu}^\infty f\qty(z+\eta_t\underline{\hat{L}_t\pn{j-1}+xe_{b_j}}\pn{i}_{b_j})f\qty(z+\eta_t\underline{\hat{L}_t\pn{j-1}+xe_{b_j}}_i)\\
        &\hspace{3cm}
        \sP_{\{r_k\}_{k\in[d]\setminus\{i,b_j\}} \sim \mathcal{D},\,r_i=z+\eta_t\underline{\hat{L}_t\pn{j-1}+xe_{b_j}}_i}
        \qty[\sele_i(r-\eta_t\hat{L}_t\pn{j-1};\nec)=1,\, \sele_i(r-\eta_t\hat{L}_t\pn{j-1};\nec_{b_j})=0]
        \dd z\\
        &\hspace{0.5cm}+
        \ind\qty[b_{j'}=b_j]\eta_t F(\nu)f\qty(\nu+\eta_t\underline{\hat{L}_t\pn{j-1}+xe_{b_j}}_i)\\
        &\hspace{3cm}
        \sP_{\{r_k\}_{k\in[d]\setminus\{i,b_j\}} \sim \mathcal{D},\,r_i=\nu+\eta_t\underline{\hat{L}_t\pn{j-1}+xe_{b_j}}_i}
        \qty[\sele_i(r-\eta_t\hat{L}_t\pn{j-1};\nec)=1,\, \sele_i(r-\eta_t\hat{L}_t\pn{j-1};\nec_{b_j})=0]
        \tag*{$\qty(\text{by }\underline{\hat{L}_t\pn{j-1}+xe_{b_j}}_i=(\hat{L}_t\pn{j-1}+xe_{b_j})_i-(\hat{L}_t\pn{j-1}+xe_{b_j})_{b_{j'}})$}
        \displaybreak[0]\\
        &\leq
        \eta_t\int_{\nu}^\infty f\qty(z+\eta_t\underline{\hat{L}_t\pn{j-1}+xe_{b_j}}\pn{i}_{b_j})f\qty(z+\eta_t\underline{\hat{L}_t\pn{j-1}+xe_{b_j}}_i)\dd z
        \tag{by $\sP[\cdot]\leq 1$ and $F(\nu)=0$}
        \displaybreak[0]\\
        &\leq
        \eta_t\int_{\nu}^\infty f\qty(z+\eta_t\underline{\hat{L}_t\pn{j-1}+xe_{b_j}}\pn{i}_{b_j})\frac{\alpha}{(z+\eta_t\underline{\hat{L}_t\pn{j-1}+xe_{b_j}}_i)^{\alpha+1}}\dd z
        \tag{by \eqref{eq:transform_pdf}}\\
        &\leq
        \frac{\alpha\eta_t}{(\eta_t\underline{\hat{L}_t\pn{j-1}+xe_{b_j}}_i)^{\alpha+1}}\int_{\nu}^\infty f\qty(z)\dd z\\
        &\leq
        \frac{\alpha\eta_t}{(1-\zeta)^{\alpha+1}(\eta_t\underline{\hat{L}}_{t,i})^{\alpha+1}}\tag*{(by \eqref{eq:zeta_bound_frechet} and $\int_{\nu}^\infty f\qty(z)\dd z=1$)}\\
        &\leq\frac{\alpha}{m(1-\zeta)^{\alpha+1}\underline{\hat{L}}_{t,i}}.
    \end{align*}
    Here, the last inequality follows from the property \eqref{eq:key_property} of $F_t$ or $D_t$, which implies that
    \begin{equation*}
        (\eta_t\underline{\hat{L}}_{t,i})^\alpha
        \geq
        m^{\alpha/(\alpha-1)}\geq m,\,\forall i:a^*_i=0.
    \end{equation*}
    Then, we have
    \begin{align*}
        \phi_{i}(\eta_t\hat{L}\pn{j}_t)-\phi_{i}(\eta_t\hat{L}\pn{j-1}_t)
        &=
        \int_0^{\max_{i:a^*_i=1}\hat{\ell}_{t,i}}\frac{\mathrm{d}}{\mathrm{d} x}\phi_{i}(\eta_t(\hat{L}_t\pn{j-1}+xe_{b_j}))\dd x\\
        &\leq 
        \max_{i:a^*_i=1}\hat{\ell}_{t,i}\frac{\alpha}{m(1-\zeta)^{\alpha+1}\underline{\hat{L}}_{t,i}}.\numberthis\label{eq:optimal_action_frechet_mL}
    \end{align*}
    By combining \eqref{eq:optimal_action_frechet_transfer} and \eqref{eq:optimal_action_frechet_mL}, we obtain
    \begin{align*}
        \lefteqn{\E\qty[\ind\qty[\max_{i:a^*_i=1}\hat{\ell}_{t,i}\leq\zeta/\eta_t]\sum_{i:a_i^*=1}\hat{\ell}_{t,i}\qty(\phi_i(\eta_t\hat{L}_t)-\phi_i(\eta_t(\hat{L}_t+\hat{\ell}_t)))\m\hat{L}_t]}\displaybreak[0]\\
        &\leq
        \E\qty[\ind\qty[\max_{i:a^*_i=1}\hat{\ell}_{t,i}\leq\zeta/\eta_t]\max_{i:a_i^*=1}\hat{\ell}_{t,i}\sum_{i:a_i^*=0}\sum_{j=1}^m\qty(\phi_{i}(\eta_t\hat{L}\pn{j}_t)-\phi_{i}(\eta_t\hat{L}\pn{j-1}_t))\m\hat{L}_t]
        +\sum_{i:a_i^*=0}\frac{c_{s,4}(\dis)}{\underline{\hat{L}}_{t,i}}
        \displaybreak[0]\\
        &\leq
        \E\qty[\ind\qty[\max_{i:a^*_i=1}\hat{\ell}_{t,i}\leq\zeta/\eta_t]\qty(\max_{i:a_i^*=1}\hat{\ell}_{t,i})^2\sum_{i:a_i^*=0}\frac{\alpha}{(1-\zeta)^{\alpha+1}\underline{\hat{L}}_{t,i}}\m\hat{L}_t]
        +\sum_{i:a_i^*=0}\frac{c_{s,4}(\dis)}{\underline{\hat{L}}_{t,i}}
        \displaybreak[0]\\
        &\leq
        \E\qty[\qty(\max_{i:a_i^*=1}\hat{\ell}_{t,i})^2\m\hat{L}_t]\sum_{i:a_i^*=0}\frac{\alpha}{(1-\zeta)^{\alpha+1}\underline{\hat{L}}_{t,i}}
        +\sum_{i:a_i^*=0}\frac{c_{s,4}(\dis)}{\underline{\hat{L}}_{t,i}}
        \displaybreak[0]\\
        &\leq
        \qty(\frac{c_{s,5}(\dis)}{(1-\zeta)^{\alpha+1}}+c_{s,4}(\dis))\sum_{i:a_i^*=0}\frac{1}{\underline{\hat{L}}_{t,i}}.
        \numberthis\label{eq:optimal_action_frechet_bound_case1}
    \end{align*}
    The last inequality follows from~\eqref{eq:optimal_action_gr_second_moment_bounds}.

    (b) Next, we consider the second case where $\max_{i:a^*_i=1}\widehat{w_{t,i}^{-1}}>\zeta/\eta_t$. 
    By Lemma~\ref{lem:optimal_action_bound_case2} we have
    \begin{align*}
        \lefteqn{\E\qty[\ind\qty[\max_{i:a^*_i=1}\hat{\ell}_{t,i}\geq\zeta/\eta_t]\sum_{i:a_i^*=1}\hat{\ell}_{t,i}\qty(\phi_i(\eta_t\hat{L}_t)-\phi_i(\eta_t(\hat{L}_t+\hat{\ell}_t)))\m\hat{L}_t]}\\
        &\leq
        \E\qty[\ind\qty[\max_{i:a^*_i=1}\hat{\ell}_{t,i}\geq\zeta/\eta_t]\sum_{i:a_i^*=1}\hat{\ell}_{t,i}\m\hat{L}_t]\displaybreak[0]\\
        &\leq
        \sum_{i:a_i^*=1}\sum_{j:a_j^*=1}\E\qty[\ind\qty[\hat{\ell}_{t,i}\geq\zeta/\eta_t]\hat{\ell}_{t,j}\m\hat{L}_t]\displaybreak[0]\\
        &=
        \sum_{i:a_i^*=1}\E\qty[\ind\qty[\hat{\ell}_{t,i}\geq\zeta/\eta_t]\hat{\ell}_{t,i}\m\hat{L}_t]+\sum_{i:a_i^*=1}\sum_{j:a_j^*=1,j\neq i}\E\qty[\ind\qty[\hat{\ell}_{t,i}\geq\zeta/\eta_t]\hat{\ell}_{t,j}\m\hat{L}_t]\displaybreak[0]\\
        &\leq
        \frac{m}{1 - \wlow} (1 - \wlow)^{\tfrac{\zeta}{\eta_t}} \qty( \frac{\zeta}{\eta_t} + \frac{1}{\wlow} )+
        m^2(1 - \wlow)^{\frac{\zeta}{\eta_t} -1}.\numberthis\label{eq:optimal_action_frechet_bound_case2}
    \end{align*}
    We obtain the lemma by combining 
    \eqref{eq:optimal_action_frechet_bound_case1} 
    and 
    \eqref{eq:optimal_action_frechet_bound_case2}.
\end{proof}

\subsection{Proofs of Theorems~\ref{thm:sto_bound} and \ref{thm:sto_bound_alpha_not2}}\label{subsec:proofs_sto_bounds}

In this section we provide proofs of the regret bounds in the stochastic setting with general shape parameters $\alpha>1$ in Theorem~\ref{thm:sto_bound} ($\alpha=2$) and Theorem~\ref{thm:sto_bound_alpha_not2} ($\alpha<2$ and $\alpha>2$). 
We divide the proofs into two cases $\alpha\geq 2$ and $\alpha<2$, both of which follow essentially the same argument. The result of Theorem~\ref{thm:sto_bound} ($\alpha=2$) is recovered as a special case of the first case. 

In the following, for simplicity, let 
$\ma=m^{\frac{1}{2}-\frac{1}{\alpha}}$ 
and
$\da=d^{\frac{1}{\alpha}-\frac{1}{2}}$
so that
$\eta_t=c\ma\da/\sqrt{t}$.
\subsubsection{Analysis for Shape \texorpdfstring{$\alpha\geq 2$}{α ≥ 2}}

By aggregating the results obtained so far, it follows that the regret satisfies
\begin{align*}
    \R(T) &\leq 
    \sum_{t=1}^{T} \E\qty[\E \qty[ \left\langle \hat{\ell}_t, \phi\qty(\eta_t \hat{L}_t ; \dis) - \phi\qty(\eta_t (\hat{L}_t + \hat{\ell}_t) ; \dis) \right\rangle
    +
    \qty(\frac{1}{\eta_{t+1}}-\frac{1}{\eta_t})\left\langle r_{t+1}, a_{t+1} - a^*\right\rangle\m\hat{L}_t]] \\
    &\hspace{3cm}+
    \begin{cases}
        \frac{\qty(\frac{\alpha}{\alpha-1}(m-1)^{1-\frac{1}{\alpha}}+\Gamma\qty(1-\frac{1}{\alpha}))\qty(d+1)^{\frac{1}{\alpha}}+m}{c\ma\da}, & \dis=\Fdis,\\
        \frac{\qty(\frac{\alpha}{\alpha-1}(m-1)^{1-\frac{1}{\alpha}}+\Gamma\qty(1-\frac{1}{\alpha}))\qty(d+1)^{\frac{1}{\alpha}}}{c\ma\da}, & \dis=\Pdis,
    \end{cases}
    \tag{by Lemma~\ref{lem:regret_decomposition}}
    \displaybreak[0]\\
    &\leq \sum_{t=1}^{T} \E\qty[\E \qty[ \left\langle \hat{\ell}_t, \phi\qty(\eta_t \hat{L}_t ; \dis) - \phi\qty(\eta_t (\hat{L}_t + \hat{\ell}_t) ; \dis) \right\rangle
    +
    \frac{\left\langle r_t, a_t - a^*\right\rangle}{c\ma\da\sqrt{t}}\m\hat{L}_t]]
    +C_2(\dis)\sqrt{md},
    \numberthis\label{eq:stochastic_regret_decomposition_pareto}
\end{align*}
where the last inequality follows from the fact that
\begin{equation*}
    \frac{1}{\eta_{t+1}}-\frac{1}{\eta_t}
    =\frac{1}{c\ma\da}\qty(\sqrt{t+1}-\sqrt{t})
    =\frac{1}{c\ma\da(\sqrt{t+1}+\sqrt{t})}
    \leq \frac{1}{2c\ma\da\sqrt{t}},
\end{equation*}
and we defined
\begin{equation*}
    C_2(\dis)\coloneqq
    \begin{cases}
        \frac{3}{2c}\qty(\frac{\alpha}{\alpha-1}(1-\frac{1}{m})^{1-\frac{1}{\alpha}}+\Gamma\qty(1-\frac{1}{\alpha})m^{\frac{1}{\alpha}-1})\qty(1+\frac{1}{d})^{\frac{1}{\alpha}}+\frac{3}{2c}\qty(\frac{m}{d})^{\frac{1}{\alpha}}, & \dis=\Fdis,\\
        \frac{3}{2c}\qty(\frac{\alpha}{\alpha-1}(1-\frac{1}{m})^{1-\frac{1}{\alpha}}+\Gamma\qty(1-\frac{1}{\alpha})m^{\frac{1}{\alpha}-1})\qty(1+\frac{1}{d})^{\frac{1}{\alpha}}, & \dis=\Pdis.
    \end{cases}
\end{equation*}

Now we consider the first term of \eqref{eq:stochastic_regret_decomposition_pareto} on $F_t$ (resp. $D_t$) for $\dis=\Fdis$ (resp. $\dis=\Pdis$), where $\eta_t\underline{\hat{L}}_{t,i}\geq m^{1/(\alpha-1)}$ for any $i$ with $a^*_i=0$. 
For $\alpha\geq 2$ we have
\begin{align*}
    \lefteqn{\E \qty[ \left\langle \hat{\ell}_t, \phi\qty(\eta_t \hat{L}_t ; \dis) - \phi\qty(\eta_t (\hat{L}_t + \hat{\ell}_t) ; \dis) \right\rangle\m\hat{L}_t]}\\
    &=
    \sum_{i:a^*_i=0}\E \qty[ \hat{\ell}_{t,i}\qty(\phi_i\qty(\eta_t \hat{L}_t ; \dis) - \phi_i\qty(\eta_t (\hat{L}_t + \hat{\ell}_t) ; \dis)) \m\hat{L}_t]\\
    &\hspace{3.5cm}+
    \sum_{i:a^*_i=1}\E \qty[ \hat{\ell}_{t,i}\qty(\phi_i\qty(\eta_t \hat{L}_t ; \dis) - \phi_i\qty(\eta_t (\hat{L}_t + \hat{\ell}_t) ; \dis)) \m\hat{L}_t]\numberthis\label{eq:to_apply_L_bound}
    \displaybreak[0]\\
    &\leq
    2(\alpha+1)\sum_{i:a^*_i=0} \frac{1}{\underline{\hat{L}}_{t,i}}
    +
    \qty(\frac{c_{s,5}(\dis)}{(1-\zeta)^{\alpha+1}}+c_{s,4}(\dis))\sum_{i:a_i^*=0}\frac{1}{\underline{\hat{L}}_{t,i}}\\
    &\hspace{1cm}+\frac{m}{1 - \wlow} (1 - \wlow)^{\tfrac{\zeta}{\eta_t}} \qty( \frac{\zeta}{\eta_t} + \frac{1}{\wlow} )+
        m^2(1 - \wlow)^{\frac{\zeta}{\eta_t} -1},\numberthis\label{eq:stability_bound_alpha_ge_2}
\end{align*}
where the inequality follows from application of Lemmas~\ref{lem:adv_bound_term} and \ref{lem:optimal_action_bound} to the first and second terms of \eqref{eq:to_apply_L_bound}, respectively. 
Here, $\zeta\in(0,1)$ is an arbitrary parameter introduced in Lemma~\ref{lem:optimal_action_bound}, and we recall
\begin{equation*}
        c_{s,4}(\dis)=
        \begin{cases}
            8\alpha/e, & \dis=\Fdis,\\
            121\alpha/14, & \dis=\Pdis,
        \end{cases}
        \quad
        c_{s,5}(\dis)=
        \begin{cases}
            2\alpha e, & \dis=\Fdis,\\
            121\alpha/14, & \dis=\Pdis,
        \end{cases}
        \quad
        \wlow=
        \begin{cases}
            1/e, & \dis=\Fdis,\\
            0.14, & \dis=\Pdis.
        \end{cases}
\end{equation*}
Then, since $\underline{\hat{L}}_{t,i}>0$ for any $i$ with $a^*_i=0$ on $F_t$ or $D_t$, by applying the second part of Lemma~\ref{lem:penalty_adv_bound} we have
\begin{equation}\label{eq:penalty_bound_alpha_ge2}
    \E\qty[\frac{\left\langle r_t, a_t - a^*\right\rangle}{c\ma\da\sqrt{t}}\m\hat{L}_t]
    \leq
    \frac{1}{c\ma\da\sqrt{t}}\frac{\alpha}{\alpha-1}\sum_{i:a_i^*=0}\frac{1}{(\eta_t\underline{\hat{L}}_{t,i})^{\alpha-1}}.
\end{equation} 
Combining \eqref{eq:stability_bound_alpha_ge_2} and \eqref{eq:penalty_bound_alpha_ge2}, we have

\begin{align*}
    \lefteqn{\E \qty[ \left\langle \hat{\ell}_t, \phi\qty(\eta_t \hat{L}_t ; \dis) - \phi\qty(\eta_t (\hat{L}_t + \hat{\ell}_t) ; \dis) \right\rangle
    +
    \frac{\left\langle r_t, a_t - a^*\right\rangle}{c\ma\da\sqrt{t}}\m\hat{L}_t]}\\
    &\leq
    \sum_{i:a^*_i=0} \qty(\qty(2(\alpha+1)+\frac{c_{s,5}(\dis)}{(1-\zeta)^{\alpha+1}}+c_{s,4}(\dis))\frac{1}{\underline{\hat{L}}_{t,i}}
    +\frac{1}{c\ma\da\sqrt{t}}\frac{\alpha}{\alpha-1}\frac{1}{(\eta_t\underline{\hat{L}}_{t,i})^{\alpha-1}})\\
    &\hspace{2cm}+\frac{m}{1 - \wlow} (1 - \wlow)^{\tfrac{\zeta}{\eta_t}} \qty( \frac{\zeta}{\eta_t} + \frac{1}{\wlow} )+
        m^2(1 - \wlow)^{\frac{\zeta}{\eta_t} -1}
    \numberthis\label{eq:stochastic_bound_on_Dt_to_copy}
    \displaybreak[0]\\
    &\leq
    \sum_{i:a^*_i=0} \qty(\qty(2(\alpha+1)+\frac{c_{s,5}(\dis)}{(1-\zeta)^{\alpha+1}}+c_{s,4}(\dis))\frac{1}{\underline{\hat{L}}_{t,i}}
    +\frac{1}{(c\ma\da)^2}\frac{\alpha}{\alpha-1}\frac{1}{\underline{\hat{L}}_{t,i}})\\
    &\hspace{2cm}+\frac{m}{1 - \wlow} (1 - \wlow)^{\tfrac{\zeta}{\eta_t}} \qty( \frac{\zeta}{\eta_t} + \frac{1}{\wlow} )+
        m^2(1 - \wlow)^{\frac{\zeta}{\eta_t} -1}
    \numberthis\label{eq:Dt_power_reduction}
    \displaybreak[0]\\
    &=
    C_3(\dis)\sum_{i:a^*_i=0} 
    \frac{1}{\underline{\hat{L}}_{t,i}}
    +C_{4,t}(\dis),\numberthis\label{eq:stochastic_bound_on_Dt}
\end{align*}
where we defined
\begin{align*}
    C_3(\dis)&\coloneqq 2(\alpha+1)+\frac{c_{s,5}(\dis)}{(1-\zeta)^{\alpha+1}}+c_{s,4}(\dis)+\frac{\alpha}{(\alpha-1)(c\ma\da)^2},\\
    C_{4,t}(\dis)&\coloneqq 
    \frac{m}{1 - \wlow} (1 - \wlow)^{\tfrac{\zeta}{\eta_t}} \qty( \frac{\zeta}{\eta_t} + \frac{1}{\wlow} )+
        m^2(1 - \wlow)^{\frac{\zeta}{\eta_t} -1}.
\end{align*}
Here, \eqref{eq:Dt_power_reduction} holds since for $\alpha\geq 2$ we have
\begin{equation*}
    \frac{1}{c\ma\da\sqrt{t}}\frac{\alpha}{\alpha-1}\frac{1}{(\eta_t\underline{\hat{L}}_{t,i})^{\alpha-1}}
    =
    \frac{1}{(c\ma\da)^2}\frac{\alpha}{\alpha-1}\frac{1}{\underline{\hat{L}}_{t,i}}
    \frac{1}{(\eta_t\underline{\hat{L}}_{t,i})^{\alpha-2}}
    \leq
    \frac{1}{(c\ma\da)^2}\frac{\alpha}{\alpha-1}\frac{1}{\underline{\hat{L}}_{t,i}},
\end{equation*}
where the inequality follows from $\eta_t\underline{\hat{L}}_{t,i}\geq m^{1/(\alpha-1)}\geq 1$ for any $i$ with $a^*_i=0$ on $F_t$ or $D_t$.

Then, by the second part of Lemma~\ref{lem:optimal_action_bound_case2}, $C_{4,t}(\dis)$ satisfies
\begin{align*}
    \sum_{t=1}^T C_{4,t}(\dis)&< \sum_{t=1}^\infty C_{4,t}(\dis)\\
    &=\sum_{t=1}^\infty \frac{m}{1 - \wlow} (1 - \wlow)^{\tfrac{\zeta}{\eta_t}} \qty( \frac{\zeta}{\eta_t} + \frac{1}{\wlow} )+
        m^2\sum_{t=1}^\infty (1 - \wlow)^{\frac{\zeta}{\eta_t} -1}\\
    &\leq m\cdot O\qty(c^2m^{1-\frac{2}{\alpha}}d^{\frac{2}{\alpha}-1})+m^2\cdot O\qty(c^2m^{1-\frac{2}{\alpha}}d^{\frac{2}{\alpha}-1})\\
    &\leq
    O\qty(c^2m^{3-\frac{2}{\alpha}}d^{\frac{2}{\alpha}-1}).\numberthis\label{eq:C3t_sum_bound}
\end{align*}

Now we consider the first term of \eqref{eq:stochastic_regret_decomposition_pareto} on $F_t^c$ or $D_t^c$. 
By applying Lemmas~\ref{lem:adv_bound_term} and \ref{lem:penalty_adv_bound}, we obtain
\begin{align*}
    \lefteqn{\E \qty[ \left\langle \hat{\ell}_t, \phi\qty(\eta_t \hat{L}_t ; \dis) - \phi\qty(\eta_t (\hat{L}_t + \hat{\ell}_t) ; \dis) \right\rangle
    +
    \frac{\left\langle r_t, a_t - a^*\right\rangle}{c\ma\da\sqrt{t}}\m\hat{L}_t]}
    \displaybreak[0]\\
    &\leq
    \frac{\frac{2(\alpha+1)^2}{\alpha-1} c\ma\da}{\sqrt{t}}m^{\frac{1}{\alpha}}d^{1-\frac{1}{\alpha}}
    +
    \begin{cases}
        \frac{\qty(\frac{\alpha}{\alpha-1}(m-1)^{1-\frac{1}{\alpha}}+\Gamma\qty(1-\frac{1}{\alpha}))\qty(d+1)^{\frac{1}{\alpha}}+m}{c\ma\da\sqrt{t}}, & \dis=\Fdis,\\
        \frac{\qty(\frac{\alpha}{\alpha-1}(m-1)^{1-\frac{1}{\alpha}}+\Gamma\qty(1-\frac{1}{\alpha}))\qty(d+1)^{\frac{1}{\alpha}}}{c\ma\da\sqrt{t}}, & \dis=\Pdis,\\
    \end{cases}
    \displaybreak[0]\\
    &=
    \frac{2c(\alpha+1)^2}{\alpha-1}\sqrt{\frac{md}{t}}
    +
    \begin{cases}
        \frac{\qty(\frac{\alpha}{\alpha-1}(m-1)^{1-\frac{1}{\alpha}}+\Gamma\qty(1-\frac{1}{\alpha}))\qty(d+1)^{\frac{1}{\alpha}}+m}{c\qty(\frac{m}{d})^{\frac{1}{2}-\frac{1}{\alpha}}\sqrt{t}}, & \dis=\Fdis,\\
        \frac{\qty(\frac{\alpha}{\alpha-1}(m-1)^{1-\frac{1}{\alpha}}+\Gamma\qty(1-\frac{1}{\alpha}))\qty(d+1)^{\frac{1}{\alpha}}}{c\qty(\frac{m}{d})^{\frac{1}{2}-\frac{1}{\alpha}}\sqrt{t}}, & \dis=\Pdis,
    \end{cases}\\
    &\leq 
    C_5(\dis)
    \sqrt{\frac{md}{t}},\numberthis\label{eq:stochastic_bound_on_Dtc}
\end{align*}
where we defined
\begin{equation*}
    C_5(\dis)\coloneqq
    \begin{cases}
        \frac{2c(\alpha+1)^2}{\alpha-1}+\frac{\qty(\frac{\alpha}{\alpha-1}(1-\frac{1}{m})^{1-\frac{1}{\alpha}}+\Gamma\qty(1-\frac{1}{\alpha})m^{\frac{1}{\alpha}-1})(1+\frac{1}{d})^{\frac{1}{\alpha}}+\frac{m}{d}^{\frac{1}{\alpha}}}{c}, & \dis=\Fdis,\\
        \frac{2c(\alpha+1)^2}{\alpha-1}+\frac{\qty(\frac{\alpha}{\alpha-1}(1-\frac{1}{m})^{1-\frac{1}{\alpha}}+\Gamma\qty(1-\frac{1}{\alpha})m^{\frac{1}{\alpha}-1})(1+\frac{1}{d})^{\frac{1}{\alpha}}}{c}, & \dis=\Pdis.
    \end{cases}
\end{equation*}

For the remainder of this proof, we write $F_t$ for the good event associated with $\dis$, using its original definition for $\dis=\Fdis$ and the event $D_t$ for $\dis=\Pdis$. Thus $F_t^c$ denotes the corresponding complementary event.

Combining \eqref{eq:stochastic_regret_decomposition_pareto} with the above results, we have 
\begin{align*}
    \R(T) &\leq
    \sum_{t=1}^{T} \E\Bigg[
        \E\Bigg[
        \ind\qty[F_t]
        \sum_{i:a^*_i=0} \frac{C_3(\dis)}{\underline{\hat{L}}_{t,i}}
        +
        \ind\qty[F_t^c]
        C_5(\dis)
        \sqrt{\frac{md}{t}}\Bigg|
        \hat{L}_t\Bigg]
        \Bigg]+
    C_2(\dis)\sqrt{md}
    +
    \sum_{t=1}^T C_{4,t}(\dis)\tag{by \eqref{eq:stochastic_bound_on_Dt} and \eqref{eq:stochastic_bound_on_Dtc}}\\
    &\leq
    \sum_{t=1}^{T} \E\Bigg[
        \E\Bigg[
        \ind\qty[F_t]
        \sum_{i:a^*_i=0} \frac{C_3(\dis)}{\underline{\hat{L}}_{t,i}}
        +
        \ind\qty[F_t^c]
        C_5(\dis)
        \sqrt{\frac{md}{t}}\Bigg|
        \hat{L}_t\Bigg]
        \Bigg]+
    C_2(\dis)\sqrt{md}
    +
    O\qty(c^2m^{3-\frac{2}{\alpha}}d^{\frac{2}{\alpha}-1}),
    \numberthis\label{eq:self-bounding_upper_bound_pareto}
\end{align*}
where the last inequality follows from \eqref{eq:C3t_sum_bound}. 

On the other hand, by Lemma~\ref{lem:lower_bound_Dt}, the regret is bounded from below as
\begin{equation}\label{eq:self-bounding_lower_bound_pareto}
    \R(T)\geq\sum_{t=1}^T\E\qty[\ind\qty[F_t]c_{s,1}(\dis)\sum_{i:a^*_i=0}\frac{\Delta_{i,\min} t^\frac{\alpha}{2}}{(c\ma\da\underline{\hat{L}}_{t,i})^\alpha}
    +\ind\qty[F_t^c]\frac{c_{s,2}(\dis)\Delta_{\min}}{m^{\alpha/(\alpha-1)}}].
\end{equation}
Applying the self-bounding technique and considering
$\eqref{eq:self-bounding_upper_bound_pareto}-\eqref{eq:self-bounding_lower_bound_pareto}/2$, we have
\begin{align*}
    \frac{\R(T)}{2}&\leq
    \sum_{t=1}^{T}\E\qty[
        \ind\qty[F_t]
        \sum_{i:a^*_i=0} 
        \qty(\frac{C_3(\dis)}{\underline{\hat{L}}_{t,i}}
        -
        \frac{c_{s,1}(\dis)\Delta_{i,\min} t^\frac{\alpha}{2}}{2(c\ma\da\underline{\hat{L}}_{t,i})^\alpha})
    ]\\
    &\hspace{2cm}+\sum_{t=1}^{T}\E\qty[
        \ind\qty[F_t^c]
        \qty(C_5(\dis)
        \sqrt{\frac{md}{t}}
        -
        \frac{c_{s,2}(\dis)\Delta_{\min}}{2m^{\alpha/(\alpha-1)}})
    ]+
    C_2(\dis)\sqrt{md}
    +
    O\qty( c^2m^{3-\frac{2}{\alpha}}d^{\frac{2}{\alpha}-1}).\numberthis\label{eq:self-bounding_general_pareto}
\end{align*}

For the first term of \eqref{eq:self-bounding_general_pareto}, we have
\begin{align*}
    \lefteqn{\frac{C_3(\dis)}{\underline{\hat{L}}_{t,i}}
        -
        \frac{c_{s,1}(\dis)\Delta_{i,\min} t^\frac{\alpha}{2}}{2(c\ma\da\underline{\hat{L}}_{t,i})^\alpha}
    \leq
    C_3(\dis)\frac{\alpha}{\alpha-1}\qty(\frac{2C_3(\dis)}{\alpha c_{s,1}(\dis)\Delta_{i,\min}})^{\frac{1}{\alpha-1}}
    \frac{\qty(c\ma\da)^{\frac{\alpha}{\alpha-1}}}{t^{\frac{\alpha}{2(\alpha-1)}}}}\\
    &=
    \begin{cases}
        \qty(2(\alpha+1)+\frac{\alpha}{(\alpha-1)(c\ma\da)^2}+\frac{2\alpha e}{(1-\zeta)^{\alpha+1}}+\frac{8\alpha}{e})
        \frac{\alpha}{\alpha-1}\\
        \hspace{1.5cm}\times\qty(\frac{4(\alpha+1)+\frac{2\alpha}{(\alpha-1)(c\ma\da)^2}+\frac{4\alpha e}{(1-\zeta)^{\alpha+1}}+\frac{16\alpha}{e}}{\alpha c_{s,1}(\Fdis)\Delta_{i,\min}})^{\frac{1}{\alpha-1}}
        \frac{\qty(c\ma\da)^{\frac{\alpha}{\alpha-1}}}{t^{\frac{\alpha}{2(\alpha-1)}}}, & \dis=\Fdis,\\
        \qty(2(\alpha+1)+\frac{\alpha}{(\alpha-1)(c\ma\da)^2}+\frac{121\alpha}{14(1-\zeta)^{\alpha+1}}+\frac{121\alpha}{14})
        \frac{\alpha}{\alpha-1}\\
        \hspace{1.5cm}\times\qty(\frac{4(\alpha+1)+\frac{2\alpha}{(\alpha-1)(c\ma\da)^2}+\frac{121\alpha}{7(1-\zeta)^{\alpha+1}}+\frac{121\alpha}{7}}{\alpha c_{s,1}(\Pdis)\Delta_{i,\min}})^{\frac{1}{\alpha-1}}
        \frac{\qty(c\ma\da)^{\frac{\alpha}{\alpha-1}}}{t^{\frac{\alpha}{2(\alpha-1)}}}, & \dis=\Pdis,
    \end{cases}\\
    &=
    O\qty(\frac{1}{
        \Delta_{i,\min}^{\frac{1}{\alpha-1}}
        m^{\frac{\alpha-2}{2(\alpha-1)}}
        d^{\frac{2-\alpha}{2(\alpha-1)}}
        t^{\frac{\alpha}{2(\alpha-1)}}
        }),\numberthis\label{eq:self-bounding_first_term_pareto}
\end{align*}
where the inequality follows from $Ax-Bx^\alpha\leq A\frac{\alpha}{\alpha-1}(\frac{A}{\alpha B})^{\frac{1}{\alpha-1}}$ for any $A,B>0$ and $\alpha>1$.

For the second term of \eqref{eq:self-bounding_general_pareto}, we have
\begin{align*}
    \lefteqn{\sum_{t=1}^T\E\qty[\ind\qty[F_t^c]\qty(
    C_5(\dis)\sqrt{\frac{md}{t}}
    -
    \frac{\frac{c_{s,2}(\dis)\Delta_{\min}}{m^{\alpha/(\alpha-1)}}}{2})]}\\
    &\leq
    \sum_{t=1}^T\max\Bigg\{
    C_5(\dis)\sqrt{\frac{md}{t}}
    -
    \frac{c_{s,2}(\dis)\Delta_{\min}}{2m^{\alpha/(\alpha-1)}},0
    \Bigg\}
    \displaybreak[0]\\
    &\leq
    C_5^2(\dis)md\frac{2m^{\alpha/(\alpha-1)}}{c_{s,2}(\dis)\Delta_{\min}}
    \displaybreak[0]\\
    &=
    \begin{cases}
        md\qty(\frac{2c(\alpha+1)^2}{\alpha-1}+
        \frac{\qty(\frac{\alpha}{\alpha-1}(1-\frac{1}{m})^{1-\frac{1}{\alpha}}+\Gamma\qty(1-\frac{1}{\alpha})m^{\frac{1}{\alpha}-1})(1+\frac{1}{d})^{\frac{1}{\alpha}}+\frac{m}{d}^{\frac{1}{\alpha}}}{c})^2
        \frac{2^{\alpha+1} \qty(m^\frac{\alpha}{\alpha-1}+1)+2}{\Delta_{\min}}, & \dis=\Fdis,\\
        md\qty(
        \frac{2c(\alpha+1)^2}{\alpha-1}
        +
        \frac{\qty(\frac{\alpha}{\alpha-1}(1-\frac{1}{m})^{1-\frac{1}{\alpha}}+\Gamma\qty(1-\frac{1}{\alpha})m^{\frac{1}{\alpha}-1})(1+\frac{1}{d})^{\frac{1}{\alpha}}}{c}
        )^2
        \frac{2e\qty(2^{\alpha-1}(2+m^{\alpha/(\alpha-1)})+2)}{\Delta_{\min}}, & \dis=\Pdis,
    \end{cases}
    \displaybreak[0]\\
    &=
    O\qty(\frac{m^{\frac{2\alpha-1}{\alpha-1}}d}{\Delta_{\min}}).\numberthis\label{eq:self-bounding_second_term_pareto}
\end{align*}
By substituting the results of 
\eqref{eq:self-bounding_first_term_pareto} and
\eqref{eq:self-bounding_second_term_pareto} into
\eqref{eq:self-bounding_general_pareto}, we obtain the regret bound for Pareto distributions or Fr\'{e}chet distributions with shape $\alpha\geq 2$ as
\begin{align*}
    \R(T) &\leq
    O\qty(\sum_{i:a^*_i=0}\sum_{t=1}^T\frac{1}{
        \Delta_{i,\min}^{\frac{1}{\alpha-1}}
        m^{\frac{\alpha-2}{2(\alpha-1)}}
        d^{\frac{2-\alpha}{2(\alpha-1)}}
        t^{\frac{\alpha}{2(\alpha-1)}}
        }
    )
    +
    O\qty(\frac{m^{\frac{2\alpha-1}{\alpha-1}}d}{\Delta_{\min}})
    +
    O\qty(\frac{\sqrt{md}}{c})
    +
    O\qty( c^2m^{3-\frac{2}{\alpha}}d^{\frac{2}{\alpha}-1})
    \displaybreak[0]\\
    &\leq
    O\qty(\frac{m^{\frac{2\alpha-1}{\alpha-1}}d}{\Delta_{\min}})+
    O\qty(\frac{\sqrt{md}}{c})+
    O\qty( c^2m^{3-\frac{2}{\alpha}}d^{\frac{2}{\alpha}-1})\\
    &\hspace{2.8cm}+
    \begin{cases*}
        O\qty(\sum_{i:a^*_i=0}\frac{\log T}{\Delta_{i,\min}}), & if $\alpha=2$,\\
        O\qty(\sum_{i:a^*_i=0}\frac{1}{(\alpha-2)}\frac{T^{\frac{\alpha-2}{2(\alpha-1)}}}{\Delta_{i,\min}^{\frac{1}{\alpha-1}}m^{\frac{\alpha-2}{2(\alpha-1)}}d^{\frac{2-\alpha}{2(\alpha-1)}}}), & if $\alpha>2$.
    \end{cases*}
\end{align*}

\subsubsection{Analysis for Shape \texorpdfstring{$\alpha\in\qty(1,2)$}{α ∈ (1,2)}}

We retain the convention that $F_t$ denotes the good event corresponding to $\dis$. 
The proof for the case of $\alpha\in(1,2)$ coincides with the previous case up to \eqref{eq:stochastic_bound_on_Dt_to_copy} and differs from \eqref{eq:Dt_power_reduction} onward, where on $F_t$ we obtain
\begin{align*}
    \lefteqn{\E \qty[ \left\langle \hat{\ell}_t, \phi\qty(\eta_t \hat{L}_t ; \dis) - \phi\qty(\eta_t (\hat{L}_t + \hat{\ell}_t) ; \dis) \right\rangle
    +
    \frac{\left\langle r_t, a_t - a^*\right\rangle}{c\ma\da\sqrt{t}}\m\hat{L}_t]}\\
    &\leq
    \sum_{i:a^*_i=0} \qty(\qty(2(\alpha+1)+\frac{c_{s,5}(\dis)}{(1-\zeta)^{\alpha+1}}+c_{s,4}(\dis))\frac{1}{\underline{\hat{L}}_{t,i}}
    +\frac{1}{c\ma\da\sqrt{t}}\frac{\alpha}{\alpha-1}\frac{1}{(\eta_t\underline{\hat{L}}_{t,i})^{\alpha-1}})\\
    &\hspace{0.8cm}+\frac{m}{1 - \wlow} (1 - \wlow)^{\tfrac{\zeta}{\eta_t}} \qty( \frac{\zeta}{\eta_t} + \frac{1}{\wlow} )+
        m^2(1 - \wlow)^{\frac{\zeta}{\eta_t} -1}
    \tag{\eqref{eq:stochastic_bound_on_Dt_to_copy} itself}
    \displaybreak[0]\\
    &\leq
    \sum_{i:a^*_i=0} \frac{2(\alpha+1)+\frac{c_{s,5}(\dis)}{(1-\zeta)^{\alpha+1}}+c_{s,4}(\dis)}{\underline{\hat{L}}_{t,i}}
    +
    \sum_{i:a_i^*=0}\frac{1}{c\ma\da\sqrt{t}}\frac{\alpha}{\alpha-1}\frac{1}{(\eta_t\underline{\hat{L}}_{t,i})^{\alpha-1}}
    +
    C_{4,t}(\dis)
    \displaybreak[0]\\
    &=
    \sum_{i:a^*_i=0} \frac{C_3'(\dis)}{\underline{\hat{L}}_{t,i}}
    +
    \sum_{i:a_i^*=0}\frac{1}{(c\ma\da)^\alpha}\frac{\alpha}{\alpha-1}\frac{1}{t^{1-\frac{\alpha}{2}}\underline{\hat{L}}_{t,i}^{\alpha-1}}
    +
    C_{4,t}(\dis).
\end{align*}
Here, we defined
\begin{equation*}
    C_3'(\dis)\coloneqq
    \begin{cases}
        2(\alpha+1)+\frac{2\alpha e}{(1-\zeta)^{\alpha+1}}+\frac{8\alpha}{e}, & \dis=\Fdis,\\
        2(\alpha+1)+\frac{121\alpha}{14(1-\zeta)^{\alpha+1}}+\frac{121\alpha}{14}, & \dis=\Pdis.
    \end{cases}
\end{equation*}

Note that we obtained \eqref{eq:self-bounding_upper_bound_pareto} by combining \eqref{eq:stochastic_regret_decomposition_pareto} and \eqref{eq:stochastic_bound_on_Dt}--\eqref{eq:stochastic_bound_on_Dtc}. Here \eqref{eq:stochastic_regret_decomposition_pareto}, \eqref{eq:C3t_sum_bound} and \eqref{eq:stochastic_bound_on_Dtc} also hold for this setting. By replacing \eqref{eq:stochastic_bound_on_Dt} with the above result, instead of \eqref{eq:self-bounding_upper_bound_pareto} we obtain
\begin{align*}
    \R(T)&\leq
    \sum_{t=1}^{T}\E\qty[
        \ind\qty[F_t]
        \sum_{i:a^*_i=0} 
        \qty(\frac{C_3'(\dis)}{\underline{\hat{L}}_{t,i}}
        +
        \frac{1}{(c\ma\da)^\alpha}\frac{\alpha}{\alpha-1}\frac{1}{t^{1-\frac{\alpha}{2}}\underline{\hat{L}}_{t,i}^{\alpha-1}})
    ]\\
    &\hspace{1.7cm}
    +\sum_{t=1}^{T}\E\qty[
        \ind\qty[F_t^c]
        C_5(\dis)\sqrt{\frac{md}{t}}
    ]
    +
    C_2(\dis)\sqrt{md}
    +
    O\qty( c^2m^{3-\frac{2}{\alpha}}d^{\frac{2}{\alpha}-1}).\numberthis\label{eq:self-bounding_upper_bound_alpha_less_2}
\end{align*}
The regret lower bound in \eqref{eq:self-bounding_lower_bound_pareto} still holds for this setting. 
Then, by following the same step as \eqref{eq:self-bounding_general_pareto} to apply the self-bounding technique 
and considering $\eqref{eq:self-bounding_upper_bound_alpha_less_2}-\eqref{eq:self-bounding_lower_bound_pareto}/2$, we have

\begin{align*}
    \frac{\R(T)}{2}&\leq
    \sum_{t=1}^{T}\E\qty[
        \ind\qty[F_t]
        \sum_{i:a^*_i=0} 
        \qty(\frac{C_3'(\dis)}{\underline{\hat{L}}_{t,i}}
        +
        \frac{1}{(c\ma\da)^\alpha}\frac{\alpha}{\alpha-1}\frac{1}{t^{1-\frac{\alpha}{2}}\underline{\hat{L}}_{t,i}^{\alpha-1}}-
        \frac{c_{s,1}(\dis)\Delta_{i,\min} t^\frac{\alpha}{2}}{2(c\ma\da\underline{\hat{L}}_{t,i})^\alpha})
    ]
    \numberthis\label{eq:self-bounding_general_pareto_alpha_less_2_part1}\\
    &+\sum_{t=1}^{T}\E\qty[
        \ind\qty[F_t^c]
        \qty(
        C_5(\dis)\sqrt{\frac{md}{t}}
        -
        \frac{c_{s,2}(\dis)\Delta_{\min}}{2m^{\alpha/(\alpha-1)}})
    ]
    +
    C_2(\dis)\sqrt{md}
    +
    O\qty( c^2m^{3-\frac{2}{\alpha}}d^{\frac{2}{\alpha}-1}),
\end{align*}
where the first term \eqref{eq:self-bounding_general_pareto_alpha_less_2_part1} can be written as
\begin{multline}\label{eq:to_bound_two_terms_pareto_alpha_less_2}
    \eqref{eq:self-bounding_general_pareto_alpha_less_2_part1}=
    \sum_{t=1}^{T}\E\Bigg[
        \ind\qty[F_t]
        \sum_{i:a^*_i=0} 
        \qty(\frac{C_3'(\dis)}{\underline{\hat{L}}_{t,i}}
    -
    \frac{c_{s,1}(\dis)\Delta_{i,\min} t^\frac{\alpha}{2}}{4(c\ma\da\underline{\hat{L}}_{t,i})^\alpha})
    \\+
    \ind\qty[F_t]\sum_{i:a^*_i=0} \frac{1}{(c\ma\da)^\alpha}
    \qty(\frac{\alpha}{\alpha-1}\frac{1}{t^{1-\frac{\alpha}{2}}\underline{\hat{L}}_{t,i}^{\alpha-1}}
    -
    \frac{c_{s,1}(\dis)\Delta_{i,\min} t^\frac{\alpha}{2}}{4\underline{\hat{L}}_{t,i}^\alpha}
    )\Bigg].
\end{multline}
The first term of \eqref{eq:to_bound_two_terms_pareto_alpha_less_2} can be bounded in the same way as \eqref{eq:self-bounding_first_term_pareto}, which results in
\begin{align*}
    \lefteqn{\frac{C_3'(\dis)}{\underline{\hat{L}}_{t,i}}
    -
    \frac{c_{s,1}(\dis)\Delta_{i,\min} t^\frac{\alpha}{2}}{4(c\ma\da\underline{\hat{L}}_{t,i})^\alpha}
    \leq
    C_3'(\dis)\frac{\alpha}{\alpha-1}\qty(\frac{4C_3'(\dis)}{\alpha c_{s,1}(\dis)\Delta_{i,\min}})^{\frac{1}{\alpha-1}}
    \frac{\qty(c\ma\da)^{\frac{\alpha}{\alpha-1}}}{t^{\frac{\alpha}{2(\alpha-1)}}}}\\
    &=
    \begin{cases}
        \qty(2(\alpha+1)+\frac{2\alpha e}{(1-\zeta)^{\alpha+1}}+\frac{8\alpha}{e})
        \frac{\alpha}{\alpha-1}
        \qty(\frac{8(\alpha+1)+\frac{8\alpha e}{(1-\zeta)^{\alpha+1}}+\frac{32\alpha}{e}}{\alpha c_{s,1}(\Fdis)\Delta_{i,\min}})^{\frac{1}{\alpha-1}}
        \frac{\qty(c\ma\da)^{\frac{\alpha}{\alpha-1}}}{t^{\frac{\alpha}{2(\alpha-1)}}}, & \dis=\Fdis,\\
        \qty(2(\alpha+1)+\frac{121\alpha}{14(1-\zeta)^{\alpha+1}}+\frac{121\alpha}{14})
        \frac{\alpha}{\alpha-1}
        \qty(\frac{8(\alpha+1)+\frac{242\alpha}{7(1-\zeta)^{\alpha+1}}+\frac{242\alpha}{7}}{\alpha c_{s,1}(\Pdis)\Delta_{i,\min}})^{\frac{1}{\alpha-1}}
        \frac{\qty(c\ma\da)^{\frac{\alpha}{\alpha-1}}}{t^{\frac{\alpha}{2(\alpha-1)}}}, & \dis=\Pdis.
    \end{cases}
\end{align*}

For the second term of \eqref{eq:to_bound_two_terms_pareto_alpha_less_2}, since $Ax^{\alpha-1}-Bx^\alpha\leq\frac{A}{\alpha}\qty(\frac{\alpha-1}{\alpha}\frac{A}{B})^{\alpha-1}$ for any $A,B>0$ and $\alpha>1$, we have
\begin{align*}
    \frac{\alpha}{\alpha-1}\frac{1}{t^{1-\frac{\alpha}{2}}\underline{\hat{L}}_{t,i}^{\alpha-1}}
    -
    \frac{c_{s,1}(\dis)\Delta_{i,\min} t^\frac{\alpha}{2}}{4\underline{\hat{L}}_{t,i}^\alpha}
    &\leq
    \frac{1}{\alpha-1}\frac{1}{t^{1-\frac{\alpha}{2}}}\qty(\frac{4}{\Delta_{i,\min} t c_{s,1}(\dis)})^{\alpha-1}\\
    &=
    \frac{1}{\alpha-1}\qty(\frac{4}{\Delta_{i,\min} c_{s,1}(\dis)})^{\alpha-1}\frac{1}{t^{\frac{\alpha}{2}}}.
\end{align*}
Therefore, \eqref{eq:self-bounding_general_pareto_alpha_less_2_part1} can be bounded as
\begin{equation*}
    \eqref{eq:self-bounding_general_pareto_alpha_less_2_part1}\leq
    \sum_{i:a_i^*=0}
    \begin{cases}
        \qty(2(\alpha+1)+\frac{2\alpha e}{(1-\zeta)^{\alpha+1}}+\frac{8\alpha}{e})
        \frac{\alpha}{\alpha-1}
        \qty(\frac{8(\alpha+1)+\frac{8\alpha e}{(1-\zeta)^{\alpha+1}}+\frac{32\alpha}{e}}{\alpha c_{s,1}(\Fdis)\Delta_{i,\min}})^{\frac{1}{\alpha-1}}\\
        \hspace{1cm}\times\sum_{t=1}^T \frac{\qty(c\ma\da)^{\frac{\alpha}{\alpha-1}}}{t^{\frac{\alpha}{2(\alpha-1)}}}
        +\frac{1}{(c\ma\da)^\alpha}
        \frac{1}{\alpha-1}\qty(\frac{4}{\Delta_{i,\min} c_{s,1}(\dis)})^{\alpha-1}\sum_{t=1}^T\frac{1}{t^{\frac{\alpha}{2}}}, & \dis=\Fdis,\\
        \qty(2(\alpha+1)+\frac{121\alpha}{14(1-\zeta)^{\alpha+1}}+\frac{121\alpha}{14})
        \frac{\alpha}{\alpha-1}
        \qty(\frac{8(\alpha+1)+\frac{242\alpha}{7(1-\zeta)^{\alpha+1}}+\frac{242\alpha}{7}}{\alpha c_{s,1}(\Pdis)\Delta_{i,\min}})^{\frac{1}{\alpha-1}}\\
        \hspace{1cm}\times\sum_{t=1}^T \frac{\qty(c\ma\da)^{\frac{\alpha}{\alpha-1}}}{t^{\frac{\alpha}{2(\alpha-1)}}}
        +\frac{1}{(c\ma\da)^\alpha}
        \frac{1}{\alpha-1}\qty(\frac{4}{\Delta_{i,\min} c_{s,1}(\dis)})^{\alpha-1}\sum_{t=1}^T\frac{1}{t^{\frac{\alpha}{2}}}, & \dis=\Pdis.
    \end{cases}
\end{equation*}
Here, note that the exponent of $t$ in the first term is $\frac{\alpha}{2(\alpha-1)}>1$.
Therefore, the summation over $t$ of the first term converges to a constant. Then, by following the same steps from \eqref{eq:self-bounding_second_term_pareto},
we obtain the regret bound for Fr\'{e}chet and Pareto distributions with shape $\alpha\in(1,2)$ as
\begin{align*}
    \R(T) &\leq
    O\qty(\sum_{i:a^*_i=0}\frac{1}{2-\alpha}\frac{1}{c^\alpha m^{\frac{\alpha}{2}-1} d^{1-\frac{\alpha}{2}}}\frac{T^{1-\frac{\alpha}{2}}}{\Delta_{i,\min}^{\alpha-1}})
    +
    O\qty(\sum_{i:a^*_i=0}\frac{m^{\frac{\alpha-2}{2(\alpha-1)}}d^{\frac{2-\alpha}{2(\alpha-1)}}}{\Delta_{i,\min}^{\frac{1}{\alpha-1}}})\\
    &\hspace{4cm}
    +O\qty(\frac{m^{\frac{2\alpha-1}{\alpha-1}}d}{\Delta_{\min}})+
    O\qty(\frac{\sqrt{md}}{c})+
    O\qty( c^2m^{3-\frac{2}{\alpha}}d^{\frac{2}{\alpha}-1}).
    \tag*{\qedsymbol}
\end{align*}

\section{Technical Lemmas}

\begin{lemma}\label{lem:phi_monotone}
    If the components of the $d$-dimensional perturbation $r$ are i.i.d., then for $\lambda\in\mathbb{R}^d$ and any base-arm set $\armset\subset[d]$, $\sum_{i\in\armset}\phi_i(\lambda)$ is monotonically decreasing in $\lambda_j$ with $j\in\armset$ and monotonically increasing in $\lambda_j$ with $j\in[d]\setminus\armset$.
\end{lemma}

\begin{proof}
Fix $\lambda\in\mathbb{R}^d$, $j\in[d]$, and $x>0$, and let
$\lambda'=\lambda+xe_j$.
For a fixed realization of $r$, let $A=\supp{\sele(r-\lambda;\nec)}$ and
$A'=\supp{\sele(r-\lambda';\nec)}$ denote the sets of selected base-arms.
The greedy algorithm uses the same fixed priority order to break ties in both runs.

We first show that $A\setminus\{j\}\subset A'$.
For each $k\neq j$, let $\armset_{\text{prefix}}(k)$ and $\armset'_{\text{prefix}}(k)$ be the sets of elements
preceding $k$ in the two greedy orderings.
Since only the weight of $j$ decreases, we have $\armset'_{\text{prefix}}(k)\subset \armset_{\text{prefix}}(k)$.
Then, by Lemma~\ref{lem:greedy_remove}, if $k\in A$ and $k\neq j$, it holds that $k\in A'$, and thus we have $A\setminus\{j\}\subset A'$.

If $j\notin\armset$, this inclusion implies
$|A'\cap\armset|\geq |A\cap\armset|$.
If $j\in\armset$, applying the same argument to $[d]\setminus\armset$
and using $|A|=|A'|=m$ gives
\begin{equation*}
    |A'\cap\armset|
    =m-|A'\cap([d]\setminus\armset)|
    \leq m-|A\cap([d]\setminus\armset)|
    =|A\cap\armset|.
\end{equation*}
Taking expectations over $r$ and using
\begin{equation*}
    \sum_{i\in\armset}\phi_i(\lambda)
    =\mathbb{E}\bigl[|A\cap\armset|\bigr]
\end{equation*}
proves both monotonicity claims.
\end{proof}

\begin{lemma}\label{lem:mono_B}
    For any decreasing function $f(x)$ and positive function $g(x)$ with $x\in[a,b]$ where $b\leq\infty$, $\int_x^b f(z)g(z)\dd z/\int_x^b g(z)\dd z$ is monotonically decreasing in $x\in[a,b]$.
\end{lemma}

\begin{proof}
    We have
    \begin{align*}
        \frac{\mathrm{d}}{\mathrm{d} x}\frac{\int_x^b f(z)g(z)\dd z}{\int_x^b g(z)\dd z} 
        &= \frac{1}{\qty(\int_x^b g(x)\dd x)^2} 
        \left(-f(x)g(x) \int_x^b g(z)\dd z 
        + g(x)\int_x^b f(z)g(z)\dd z  \right) \\
        &= \frac{-g(x)}{\qty(\int_x^b g(x)\dd x)^2} 
        \left( \int_x^b f(x)g(z)\dd z 
        - \int_x^b f(z)g(z)\dd z  \right) \leq 0.\qedhere
    \end{align*}
\end{proof}

\begin{lemma}[H\"older's inequality]\label{lem:holder}
Let $p,q>1$ satisfy $\frac{1}{p}+\frac{1}{q}=1$. For measurable functions
$f$ and $g$ such that $|f|^p$ and $|g|^q$ are integrable,
\begin{equation*}
    \int |f(x)g(x)|\,\dd x
    \leq
    \left(\int |f(x)|^p\,\dd x\right)^{1/p}
    \left(\int |g(x)|^q\,\dd x\right)^{1/q}.
\end{equation*}
\end{lemma}

\section{Expression of Base-arm Selection Probability}\label{app:expression}

In this part, we show how to construct an explicit expression of the base-arm selection probability $\phi_i(\eta_t\hat{L}_t)$ for $i\in[d]$.
Although the explicit expression of the arm-selection probability is not essential for our regret analysis, its construction is nontrivial and may be of independent interest. 

Consider the matroid action set $\nec$ corresponding to the matroid $\sM=([d],\sI)$.
In this part, for $\armset\subset[d]$, let $\sele^\circ(\armset;\nec)$ denote the set of base-arms selected by the greedy algorithm that processes the base-arms in $\armset$ in increasing order of their indices.
Thus,
\begin{equation*}
    \sele^\circ(\armset;\nec)\in\supp{\sI|\armset}.
\end{equation*}
To be specific, starting from $S=\varnothing$, the algorithm examines the
base-arms $k\in\armset$ one by one in increasing order of $k$ and adds $k$ to $S$ whenever $S\cup\{k\}\in\supp{\sI}$.

In the following, we first construct a set $\sC_i$ for each base-arm $i\in[d]$, which enables us to express the selection probability $\phi_i(\lambda)$ in closed form. Based on this construction, we derive the explicit expression in \eqref{eq:phi_expression} and establish its correctness in Lemma~\ref{lem:phi_expression}.

\paragraph{Construction of $\sC_i$.}

For a fixed base-arm $i\in[d]$, define
\begin{equation*}
    \mathfrak{E}_i
    \coloneqq
    \{\armset\subset[d]\setminus\{i\}:\armset\rightsquigarrow_{\nec} i\},
\end{equation*}
which consists of all base-arm sets $\armset$ such that $i$ is selected when $\armset$ forms the set of base-arms preceding $i$.
For each $\armset\in\mathfrak{E}_i$, let
$\armset_{\succ i}\coloneqq[d]\setminus(\armset\cup\{i\})$
denote the set of base-arms following $i$, and define
\begin{equation*}
    \armset^{\cap}_{\prec i}
    \coloneqq
    \sele^\circ(\armset;\nec),
\end{equation*}
which denotes the set of base-arms selected by the greedy algorithm when it processes the base-arms in $\armset$ in increasing order of their indices.

Next, we group the sets in $\mathfrak{E}_i$ according to the value of
$\armset^{\cap}_{\prec i}$. More precisely, define an equivalence relation on
$\mathfrak{E}_i$ by
\begin{equation*}
    \armset\sim_i \widetilde{\armset}
    \quad\Longleftrightarrow\quad
    \sele^\circ(\armset;\nec)=\sele^\circ(\widetilde{\armset};\nec).
\end{equation*}
From each equivalence class, choose one representative $\armset$, and let
$\overline{\mathfrak{E}}_i$ denote the resulting set of representatives.
For each $\armset\in\overline{\mathfrak{E}}_i$, define
\begin{equation*}
\armset^{\cap}_{\prec i}
\coloneqq
\sele^\circ(\armset;\nec)
\end{equation*}
and
\begin{equation*}
\armset^{\cap}_{\succ i}
\coloneqq
\qty{
j\in[d]\setminus(\armset\cup\{i\}):
\armset^{\cap}_{\prec i}
\neq
\sele^\circ(\armset\cup\{j\};\nec)
},
\end{equation*}
which consists of the base-arms $j$ whose addition to $\armset$ changes the set selected by the greedy algorithm when the elements are processed in increasing order of their indices.

Finally, define
\begin{equation}\label{eq:Ci_definition}
\sC_i
\coloneqq
\qty{
(\armset^{\cap}_{\prec i},\armset^{\cap}_{\succ i}):
\armset\in\overline{\mathfrak{E}}_i
}.
\end{equation}

\paragraph{Explicit expression of $\phi_i(\cdot)$ and its correctness.}

We first present the following lemma, which shows how our construction leverages the greedy algorithm.

\begin{lemma}\label{lem:greedy_prefix}
    For any $\Bpair\in\sC_i$ and any $\armset\subset[d]$ satisfying $\armset^{\cap}_{\prec i}\subset\armset\subset[d]\setminus(\armset^{\cap}_{\succ i}\cup\{i\})$, we have
    \begin{equation*}
        \sele^\circ(\armset;\nec)=\armset^{\cap}_{\prec i}.
    \end{equation*}
\end{lemma}

\begin{proof}
    By construction of $\sC_i$, every $\Bpair\in\sC_i$ satisfies the properties as follows.
\begin{enumerate}[label=(\roman*)]
    \item\label{item:P1} $\armset^{\cap}_{\prec i}\in\supp{\sI}$;
    \item\label{item:P2}
        $\sele^\circ(\armset^{\cap}_{\prec i}\cup\{j\};\nec)=\armset^{\cap}_{\prec i}$
        for any $j\in[d]\setminus(\armset^{\cap}_{\prec i}\cup\armset^{\cap}_{\succ i}\cup\{i\})$.
\end{enumerate}
    By~\ref{item:P1}, the greedy algorithm selects all base-arms when run
    on $\armset^{\cap}_{\prec i}$. We add the base-arms in
    $\armset\setminus\armset^{\cap}_{\prec i}$ one at a time and show that each
    addition leaves the output unchanged.
    Suppose that the output on the current set is $\armset^{\cap}_{\prec i}$,
    and we add a base-arm $j$.
    Since $\armset\cap(\armset^{\cap}_{\succ i}\cup\{i\})=\varnothing$, \ref{item:P2} applies
    to $j$, and thus $j$ is rejected. The remaining process is then
    unaffected, so the output is still $\armset^{\cap}_{\prec i}$.
    Repeating this until all base-arms in $\armset$ are added completes the proof.
\end{proof}

Finally, we present the explicit expression of $\phi_i(\lambda)$ and establish its correctness in the following lemma.

\begin{lemma}\label{lem:phi_expression}
    The probability of selecting a base-arm $i$ given $\hat{L}_t$ is $w_{t,i}=\phi_i(\eta_t\hat{L}_t)$, where for $\lambda\in[0,\infty)^d$
    \begin{equation}\label{eq:phi_expression}
        \phi_i(\lambda)
        =
        \int_\nu^\infty\sum_{\Bpair\in\sC_i}\qty(\prod_{i'\in\armset^{\cap}_{\prec i}}(1-F(z+\ul\pn{i}_{i'}))\prod_{i'\in\armset^{\cap}_{\succ i}}F(z+\ul\pn{i}_{i'}))\dd F(z+\ul_i).
    \end{equation}
    Here, $\sC_i$ is a set of pairs of subsets of base-arms defined in \eqref{eq:Ci_definition}.
\end{lemma}

\begin{proof}
    For a given $\lambda\in\mathbb{R}^d$, let
    \begin{equation*}
        \armset_{\prec i}(r)\coloneqq\{k\in[d] : \sigma_k(r-\lambda)<\sigma_i(r-\lambda)\}
    \end{equation*}
    denote the set of base-arms that precede $i$ in descending order of
    $r-\lambda$, which is a random set determined by the perturbation $r$.
    By Lemma~\ref{lem:greedy_equiv}, whether $i$ is selected is determined by $\armset_{\prec i}(r)$, regardless of the order of the base-arms in it.
    For any $\Bpair\in\sC_i$, define
    \begin{equation*}
        \mathfrak{E}\Bpair\coloneqq
        \qty{\armset:\armset^{\cap}_{\prec i}\subset\armset\subset[d]\setminus(\armset^{\cap}_{\succ i}\cup\{i\})},
    \end{equation*}
    which consists of the sets of base-arms that contain all base-arms in $\armset^{\cap}_{\prec i}$ and none in $\armset^{\cap}_{\succ i}\cup\{i\}$, i.e., the possible sets of base-arms preceding $i$ in descending order of $r-\lambda$ that are consistent with $\Bpair$.
    One can see that
    \begin{equation*}
        \phi_i(\lambda)
        =
        \sum_{\Bpair\in\sC_i}\sum_{\armset\in\mathfrak{E}\Bpair}\sP\qty[\armset_{\prec i}(r)=\armset\m\lambda].
    \end{equation*}
    Therefore, to show the lemma, i.e., $\phi_i(\eta_t\hat{L}_t)$ is the probability of selecting base-arm $i$ given $\eta_t\hat{L}_t$, it suffices to establish the following three claims:
    \begin{enumerate}[label=(\arabic*)]
        \item For any $\armset_{\prec i}(r)$ such that $i$ is selected, there must exist a pair $\Bpair\in\sC_i$ such that $\armset_{\prec i}(r)\in\mathfrak{E}\Bpair$.
        \item For any $\Bpair\in\sC_i$ and $\armset_{\prec i}(r)\in\mathfrak{E}\Bpair$, the base-arm $i$ is selected given $\armset_{\prec i}(r)$.
        \item The families $\mathfrak{E}\Bpair$ for distinct pairs $\Bpair\in\sC_i$ are pairwise disjoint.
    \end{enumerate}
    We prove them in order.

    \paragraph{Claim~1.}
    By the construction of $\sC_i$, 
    we can see that for any $\armset_{\prec i}\in\mathfrak{E}_i$, there must exist a pair $\Bpair\in\sC_i$ such that 
    \begin{equation*}
        \armset^{\cap}_{\prec i}=\sele^\circ(\armset_{\prec i};\nec),
    \end{equation*}
    which implies that $\armset^{\cap}_{\prec i}\subset\armset_{\prec i}$.
    Then, it suffices to show that $\armset^{\cap}_{\succ i}\cap\armset_{\prec i}=\varnothing$.
    For any $j\in\armset_{\prec i}\setminus\armset^{\cap}_{\prec i}$, the greedy algorithm rejects $j$ after selecting $\armset^{\cap}_{\prec i}\cap[j-1]$.
    When $j$ is added to the representative used to construct $\Bpair$, the selected set immediately before processing $j$ is again $\armset^{\cap}_{\prec i}\cap[j-1]$.
    Thus, $j$ is again rejected and the greedy output remains unchanged, so $j\notin\armset^{\cap}_{\succ i}$ by definition.
    Since $\armset^{\cap}_{\prec i}\cap\armset^{\cap}_{\succ i}=\varnothing$ by construction, we conclude that $\armset_{\prec i}\cap\armset^{\cap}_{\succ i}=\varnothing$, which completes the proof of Claim~1.
    
    \paragraph{Claim~2.}
    By Lemma~\ref{lem:greedy_equiv}, we may assume that
    the base-arms preceding $i$ are processed in increasing order of their indices.
    Then, by Lemma~\ref{lem:greedy_prefix}, for every $\armset_{\prec i}(r)\in\mathfrak{E}\Bpair$, the set selected before $i$ is processed is $\armset^\cap_{\prec i}$. 
    Therefore, it suffices to show that, for any $\Bpair\in\sC_i$, $i$ is selected if the current selected set is $\armset^\cap_{\prec i}$.
    By the construction of $\sC_i$, for any $\Bpair\in\sC_i$, there exists a set of base-arms $\armset_{\prec i}(r)\in\mathfrak{E}_i$ such that
    \begin{equation*}
        \armset^\cap_{\prec i}=\sele^\circ(\armset_{\prec i}(r);\nec).
    \end{equation*}
    Furthermore, since $\armset_{\prec i}(r)\in\mathfrak{E}_i$, $i$ is selected when the set of base-arms preceding it is $\armset_{\prec i}(r)$.
    These two results together imply that $i$ is selected when the current selected set is $\armset^{\cap}_{\prec i}$, which completes the proof of Claim~2.

    \paragraph{Claim 3.}
    For distinct pairs $\Bpair,(\widetilde{\armset}^{\cap}_{\prec i},\widetilde{\armset}^{\cap}_{\succ i})\in\sC_i$, we have $\armset^{\cap}_{\prec i}\neq\widetilde{\armset}^{\cap}_{\prec i}$.
    Then, by Lemma~\ref{lem:greedy_prefix}, for any $\armset_{\prec i}(r)\in\mathfrak{E}\Bpair$ and $\widetilde{\armset}_{\prec i}(r)\in\mathfrak{E}(\widetilde{\armset}^{\cap}_{\prec i},\widetilde{\armset}^{\cap}_{\succ i})$, we have
    \begin{equation*}
        \armset^\cap_{\prec i}=\sele^\circ(\armset_{\prec i}(r);\nec),\quad \widetilde{\armset}^\cap_{\prec i}=\sele^\circ(\widetilde{\armset}_{\prec i}(r);\nec).
    \end{equation*}
    This implies that $\armset_{\prec i}(r)\neq\widetilde{\armset}_{\prec i}(r)$, which completes the proof of Claim~3.
\end{proof}

\end{document}